\documentclass[10pt,journal]{IEEEtran}
\usepackage[dvips]{epsfig}
\usepackage{amsmath, amssymb, amsthm, bm}
\usepackage{lipsum}
\usepackage{cuted}
\usepackage[tight,footnotesize]{subfigure}
\usepackage{multirow}
\usepackage[switch]{lineno}
\usepackage{longtable}
\usepackage{rotating}
\usepackage{dsfont}
\usepackage{float}
\usepackage{algorithm}
\usepackage{color}
\usepackage{hyperref}
\usepackage{algpseudocode}
\usepackage[square, comma, sort&compress, numbers]{natbib}

\newtheorem{theorem}{Theorem}

\begin{document}

\title{ Multiobjective Test Problems with Degenerate Pareto Fronts
\thanks{L. Zhen is with the Institute of High Performance Computing, Agency for Science, Technology and Research (A*STAR), Singapore 138632.}
\thanks{M. Li is with the Centre of Excellence for Research in Computational Intelligence and Applications (CERCIA), School of Computer Science, University of Birmingham, Birmingham B152TT,~UK.}
\thanks{R. Cheng and X. Yao are with the Shenzhen Key Laboratory of Computational Intelligence, University Key Laboratory of Evolving Intelligent Systems of Guangdong Province, Department of Computer Science and Engineering, Southern University of Science and Technology, Shenzhen 518055, China.}
\thanks{D. Peng is with the Machine Intelligence Laboratory, College of Computer Science, Sichuan University,  Chengdu 610065,~China.}
\thanks{X. Yao is also with the CERCIA, School of Computer Science, University of Birmingham,~UK.}
\thanks{Corresponding author:~Xin Yao~(email: x.yao@cs.bham.ac.uk).}}

\author{Liangli~Zhen, Miqing~Li, Ran~Cheng, Dezhong~Peng, and Xin~Yao,~\IEEEmembership{Fellow,~IEEE}}

\maketitle

\begin{abstract}
In multiobjective optimisation, a set of scalable test problems with a variety of features allow researchers to investigate and evaluate the abilities of different optimisation algorithms, and thus can help them to design and develop more effective and efficient approaches. Existing test problem suites mainly focus on situations where all the objectives are fully conflicting with each other. In such cases, an $m$-objective optimisation problem has an $(m-1)$-dimensional Pareto front in the objective space. However, in some optimisation problems, there may be unexpected characteristics among objectives, \textit{e.g.}, redundancy. The redundancy of some objectives can lead to the multiobjective problem having a degenerate Pareto front, i.e., the dimension of the Pareto front of the $m$-objective problem be less than $(m-1)$. In this paper, we systematically study degenerate multiobjective problems. We abstract three general characteristics of degenerate problems, which are not formulated and systematically investigated in the literature. Based on these characteristics, we present a set of test problems to support the investigation of multiobjective optimisation algorithms under situations with redundant objectives. To the best of our knowledge, this work is the first one that explicitly formulates these three characteristics of degenerate problems, thus allowing the resulting test problems to be featured by their generality, in contrast to existing test problems designed for specific purposes (\textit{e.g.}, visualisation). The usefulness of the proposed test problems has been demonstrated by the extensive studies of ten representative multiobjective evolutionary algorithms on these problems. The results indicate that none of the tested algorithms is able to effectively solve these proposed problems, calling for the need for developing new approaches to address degenerate multiobjective problems. The discussion regarding why certain MOEAs struggle to solve the degenerate problems and potentially useful strategies for overcoming the challenges of these degenerate problems are also presented.
\end{abstract}

\begin{IEEEkeywords}
Multiobjective test problems, degenerated Pareto fronts, objective reduction, many-objective optimisation.
\end{IEEEkeywords}

\IEEEpeerreviewmaketitle

\section{Introduction}
\label{Sec:1}

\begin{figure}[thbp]
\centering
\includegraphics[width= 0.34 \textwidth]{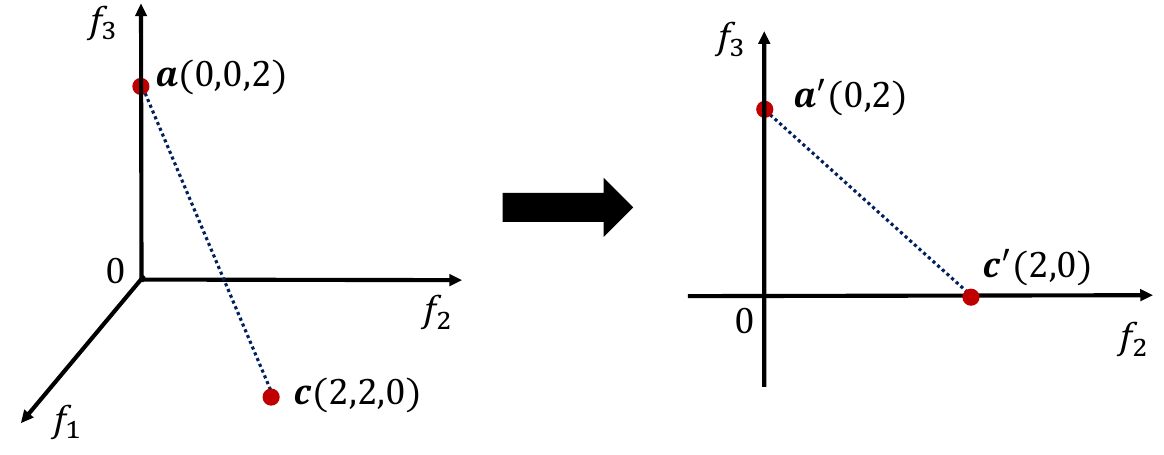}
\caption{A toy example of a degenerate problem. The left side shows the optimal solution set (the points on Line $ac$) in the objective space of the $3$-objective problem, and the right side shows the optimal solution set (the points on Line $a^\prime c^\prime$) in the space spanned by the essential objectives $f_2$ and $f_3$, where $f_1 = f_2 = x$, $f_3 = 2-x$ and $x$ is the decision variable. The objective of $f_1$ is redundant since the existence of $f_1$ does not affect the optimal solution set.}
\label{toyexample}
\end{figure}

\IEEEPARstart{I}{n} multiobjective optimisation, researchers generally assume that the objectives are conflicting with each other. In such case, an $m$-objective optimisation problem has an $(m-1)$-dimensional Pareto front in the objective space. In practice, however, this may not always be true. For example, when dealing with a dynamic optimisation scenario, an engineer may not look carefully into the connection of the existing objectives, but rather add new objectives to accommodate new requirements. The added new objectives may be harmonious with the existing objectives (or their combinations). The redundancy of the objectives can lead to the multiobjective problem has a degenerate Pareto front, i.e., the dimension of the Pareto front of the $m$-objective problem be less than $(m-1)$. Such problems are called degenerate problems~\cite{Ishibuchi2016de}. A toy example is shown in Fig.~\ref{toyexample}. The degenerate problems widely exist in real-world applications, such as multi-speed gearbox design~\cite{jain1990theory}, storm-drainage system planning~\cite{musselman1980tradeoff}, car structure design~\cite{gu2001optimisation, sinha2013using}, and optimal product selection in software engineering~\cite{Hierons2016}.

Degenerate problems appear relatively rare in evolutionary multiobjective optimization research. And more importantly, they are designed to serve some particular purposes or in accordance with specific landscape patterns, thus failing to represent the variety of real-world scenarios. For example, Deb-Thiele-Laumanns-Zitzler~(DTLZ)~5~\cite{Deb2005a}, DTLZ6~\cite{Deb2005a}, and Walking fish group~(WFG)~3~\cite{huband2005scalable, Huband2006} are three degenerate test problems\footnote{DTLZ5, DTLZ6 and WFG3 were originally intended to be degenerate multiobjective test problems. However, their true Pareto fronts have nondegenerate parts when they have four or more objectives~\cite{Ishibuchi2016de}.} whose PFs lie on 1D curves no matter how many objectives the problem has, and all of the objectives except the last one are multiples of the first objective on their PFs. The DTLZ5 problem is further extended into DTLZ5($I$,$M$)~\cite{Deb2005on}, an $M$-objective problem with $I$ specifiable essential objectives\footnote{The Pareto front of DTLZ5($I$,$M$) also has a nondegenerate part. Reference \cite{Saxena2013objective} introduced a constraint to the original formulation to remove the nondegenerate part.}. There are also some other degenerate multiobjective test problems. For instance, to help researchers to easily view and understand the search behaviour of multiobjective optimizers, the multi-point distance minimization problem (MP-DMP)~\cite{karlsson1988scanline, Ishibuchi2013many, Cheung2016} and the multi-line distance minimization problem (ML-DMP)~\cite{Rectangle, Li2017} are developed. MP-DMP is to simultaneously minimize the distances of a point to a pre-specified set of target points, and ML-DMP aims to simultaneously minimize the distances of a point to a set of target lines. Since the Pareto-optimal regions of these two problems in the decision space typically lie on a $2$D manifold (regardless of the number of the objectives and the decision variables)~\cite{Li2017}, their PFs are also on a $2$D manifold, and this makes them be degenerate problems. In \cite{saxena2011framework}, a set of problems whose PFs lie on a $1$D or a $2$D manifold are presented to stress the complexity of the Pareto optimal solutions in the decision space. In contrast, in \cite{liu2017adaptively} to emphasize the effectiveness of the Pareto optimal solutions in the objective space, another set of problems is proposed, where the redundant objectives are all equal to zero and the degenerate PF is determined only by the first part of the objectives.

On the other hand, objective reduction techniques have been receiving increasing attention in the evolutionary multiobjective optimisation area~\cite{Brockhoff2009, Singh2011, Saxena2013objective, jaimes2014objective, Wang2016, Cheung2016, yuan2017objective}. However, the lack of a comprehensive set of degenerate problems may limit systematic investigations of their performance. Algorithms which perform well on existing degenerate test problems with particular properties may not be able to work in real-life degenerate cases, where the correlation between objectives can be of high complexity.

As we mentioned above, most of the existing degenerate test problems only considered the explicit redundancy of objectives, \textit{e.g.}, DTLZ5, DTLZ6, WFG3, and DTLZ5($I$,$M$). There do exist some test problems, such as MP-DMP and ML-DMP, which have implicit redundancy of objectives, but they have fixed essential objectives, and more importantly, they were designed for solutions visualisation. Furthermore, the objective redundancy of degenerate problems is not systematically investigated and their characteristics are not explicitly formulated as well.

Given this, we propose a set of degenerate test problems aiming to reflect the generality of degenerate problems. The main contributions of this paper are summarised as follows:
\begin{itemize}
  \item From the relation among the objectives of problems, we capture three characteristics of degenerate problems: explicit redundancy, implicit redundancy and partial redundancy.
  \item Based on a unified formulation and the captured characteristics, five test problems are proposed. These problems contain a variety of representative characteristics and features, which enable researchers to investigate the working mechanisms of different multiobjective evolutionary algorithms (MOEAs) on degenerate problems, particularly objective reduction-based algorithms.
  \item Ten evolutionary algorithms are evaluated on the proposed degenerate test problems. The results show that the proposed three characteristics have a significant impact on the performance of the existing algorithms. They bring different types of difficulties for objective reduction techniques. In addition, none of the tested algorithms can obtain good performance on the high-dimensional degenerate instances. The discussion regarding why certain MOEAs struggle to solve the degenerate problems and potentially useful strategies for overcoming the challenges of these degenerate problems are also presented.
\end{itemize}

The remainder of this paper is organised as follows. Section \ref{Sec:2} and Section \ref{Sec:3} are devoted to the design principles and the description of three characteristics of test problems with degenerate PFs, respectively. Based on the principles and the analysis in the previous two sections, five test problems are proposed in Section \ref{Sec:4}. Section \ref{Sec:5} presents experimental results of ten state-of-the-art MOEAs on the proposed test problems and also the impact of the presented three characteristics for existing objective-reduction techniques. Section \ref{Sec:6} provides a discussion of potential solutions to degenerate problems based on our own practice. Finally, Section \ref{Sec:7} concludes the paper.

\section{Design Principles}
\label{Sec:2}
In order to extend and generalise the test problems easily, we follow four basic principles to design the test problems as suggested in \cite{Deb2005a}, \cite{Huband2006} and \cite{Cheng2016}:
\begin{enumerate}
  \item They can be constructed with a uniform formulation.
  \item They should be scalable to the number of decision variables.
  \item They should be scalable to the number of objectives.
  \item The resulting PF of the problem should be exactly known, and the corresponding decision variable values should also be easy to find.
\end{enumerate}

In this paper, we use the following uniform formulation for all the proposed test problems:
\begin{equation}
\label{eq3.1}
\begin{split}
&f_1(\mathbf{x})= h_1(\gamma_1(\mathbf{x}), \dots, \gamma_d(\mathbf{x}));\\
&\hspace{26mm} \vdots\\
&f_m(\mathbf{x})= h_m(\gamma_1(\mathbf{x}), \dots, \gamma_d(\mathbf{x}))
\end{split}
\end{equation}
with
\begin{equation}
\label{eq3.2}
\begin{split}
&\gamma_1(\mathbf{x}) = p_1(\mathbf{x}^l)(1+g_1(\mathbf{x}^r));\\
&\hspace{24mm} \vdots\\
&\gamma_d(\mathbf{x}) =  p_d(\mathbf{x}^l)(1+g_d(\mathbf{x}^r)),
\end{split}
\end{equation}
where $\mathbf{x}^l = (x_1, \dots, x_{d-1})$ is the first part of the decision vector and $\mathbf{x}^r = (x_d, \dots, x_n)$ is the other part, the functions $p_1, \dots, p_d$ are used to define the shape of the Pareto front, known as the \textit{shape functions}, $g_1, \dots, g_d$ define the fitness landscape, known as the \textit{landscape functions}~\cite{Cheng2016}, $\gamma_1, \gamma_2, \dots, \gamma_d$ are essential objective functions, $f_1, \dots, f_m$ are problem objectives, and $h_1(.), \dots, h_m(.)$ are transforming functions which define the relation between the problem objectives and the essential objectives.

Denoting $ \mathbf{g}(\mathbf{x}^r) = (g_1(\mathbf{x}^r),  \dots, g_d(\mathbf{x}^r))$, $\mathbf{p}(\mathbf{x}^l) = (p_1(\mathbf{x}^l), \dots, p_d(\mathbf{x}^l))$, $\bm{\gamma}(\mathbf{x}) = (\gamma_1(\mathbf{x}), \dots, \gamma_d(\mathbf{x}))$, $\mathbf{h}(\bm{\gamma}(\mathbf{x})) = (h_1(\bm{\gamma}(\mathbf{x})), \dots, h_m(\bm{\gamma}(\mathbf{x})))$, and $\mathbf{f}(\mathbf{x}) = (f_1(\mathbf{x}), \dots, f_m(\mathbf{x}))$, we can rewrite the formulation as
\begin{equation}
\label{eq3.3}
\begin{split}
&\mathbf{f}(\mathbf{x})= \mathbf{h}(\bm{\gamma}(\mathbf{x}))
\end{split}
\end{equation}
with
\begin{equation}
\label{eq3.4}
\begin{split}
&\bm{\gamma}(\mathbf{x}) = \mathbf{p}(\mathbf{x}^l)\circ(\mathbf{1}+\mathbf{g}(\mathbf{x}^r)),
\end{split}
\end{equation}
where the symbol $\circ$ denotes an entry-wise product operation, and $\mathbf{1} \in \mathbb{R}^d$ is a vector of ones. Please note that (\ref{eq3.4}) defines essential objective functions, and (\ref{eq3.3}) transforms these essential objectives into another high-dimensional space and obtains the objective functions for the test problems.

For a given test problem formulated by (\ref{eq3.4}) and (\ref{eq3.3}), the goal of a multiobjective optimiser is to find the Pareto decision vectors $\mathbf{x} = (\mathbf{x}^l, \mathbf{x}^r)$ such that $\mathbf{g}(\mathbf{x}^r) = \mathbf{0}$ and $\mathbf{f}(\mathbf{x}) = \mathbf{h}(\mathbf{p}(\mathbf{x}^l)$. Following this manner, we can use the designed $\mathbf{g}(\mathbf{x}^r)$ to evaluate the ability of an algorithm to converge to the PF, and use the designed $\mathbf{p}(\mathbf{x}^l)$ to test an algorithm's ability to obtain diverse solutions.

Since this paper aims to present some important characteristics of the problems with degenerate PFs, we focus on the design of the
4
 transformations $\mathbf{h}$ in (\ref{eq3.3}), which controls the relationships between the problem objectives $\mathbf{f}(\mathbf{x})$ and the essential objectives $\bm{\gamma}(\mathbf{x})$. For the definition of the essential objectives in (\ref{eq3.4}), we simply select/design based on the existing definitions in \cite{Deb2005a} and \cite{Huband2006}. In the next section, we will present the detailed characteristics of the proposed test problems with respect to the design of $\mathbf{h}$ in (\ref{eq3.3}).

\section{Problem Characteristics}
\label{Sec:3}

As we mentioned before, the correlation between the objectives has not been systematically considered in the existing test problems. In this section, we present the following three characteristics of degenerate problems.

\subsection{Explicitly Redundant Objectives}

In many-objective optimisation, there exist some problems that explicitly have redundant objectives, \textit{i.e.}, the existence of these objectives does not have any impact on the solutions for the optimisation problem. The redundant objectives are part of the problem's objectives. With regard to this case, we have the following theorem:
\begin{theorem}
\label{pro1}
Supposing that there is an optimisation problem with $d$ conflicting objectives $f_1(\mathbf{x}), f_2(\mathbf{x}), \dots, f_d(\mathbf{x})$, we further add more objective $f_{d+1}(\mathbf{x}) = h_{d+1}(f_1(\mathbf{x}), f_2(\mathbf{x}), \dots, f_d(\mathbf{x})), \dots, f_{m}(\mathbf{x}) = h_{m}(f_1(\mathbf{x}), f_2(\mathbf{x}), \dots, f_d(\mathbf{x}))$, where $h_{d+1}(\cdot), \dots, h_{m}(\cdot)$ are non-decreasing functions with respective to $f_1(\mathbf{x}), f_2(\mathbf{x}), \dots, f_d(\mathbf{x})$, then the new problem with $m$ objectives has the same Pareto solution set as the original $d$-objective problem.
\end{theorem}
\begin{IEEEproof}
See Appendix \ref{ProofT1}.
\end{IEEEproof}

Based on Theorem \ref{pro1}, we can define test problems with an arbitrary number of redundant objectives. If an algorithm can find the essential objectives of a problem, it can ignore all redundant objectives and potentially obtain a good set of solutions for the original problem. Here, we give a general formulation of this type of problems. Let $f_1(\mathbf{x}), \dots, f_d(\mathbf{x})$ be $d$ conflicting objectives, then we add $(m-d)$ redundant objectives to the problem as:
\begin{equation}
\label{eq4.0}
\begin{split}
&f_{d+1}(\mathbf{x}) = h_{d+1}(f_1(\mathbf{x}), f_2(\mathbf{x}), \dots, f_d(\mathbf{x}));\\
&\hspace{26mm} \vdots \\
&f_{m}(\mathbf{x}) = h_{m}(f_1(\mathbf{x}), f_2(\mathbf{x}), \dots, f_d(\mathbf{x})),
\end{split}
\end{equation}
where $h_{d+1}(\cdot), \dots, h_{m}(\cdot)$ are non-decreasing functions with respect to their corresponding inputs.

\subsection{Implicitly Redundant Objectives}

\begin{figure*}[bpt]
\centering
\subfigure[]{
\centering
\includegraphics[width=0.2\textwidth]{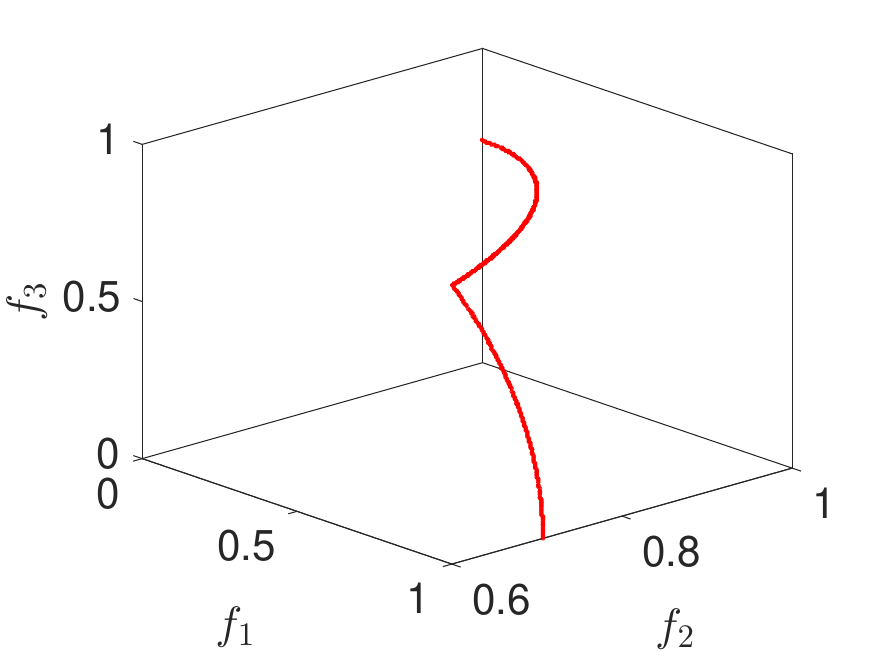}}
\subfigure[]{
\centering
\includegraphics[width=0.2\textwidth]{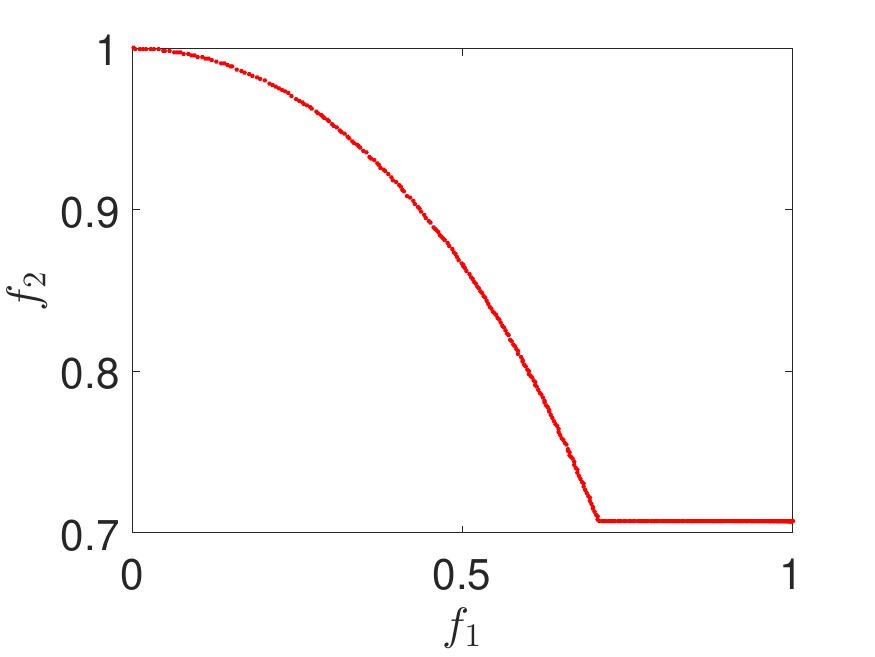}}
\subfigure[]{
\centering
\includegraphics[width=0.2\textwidth]{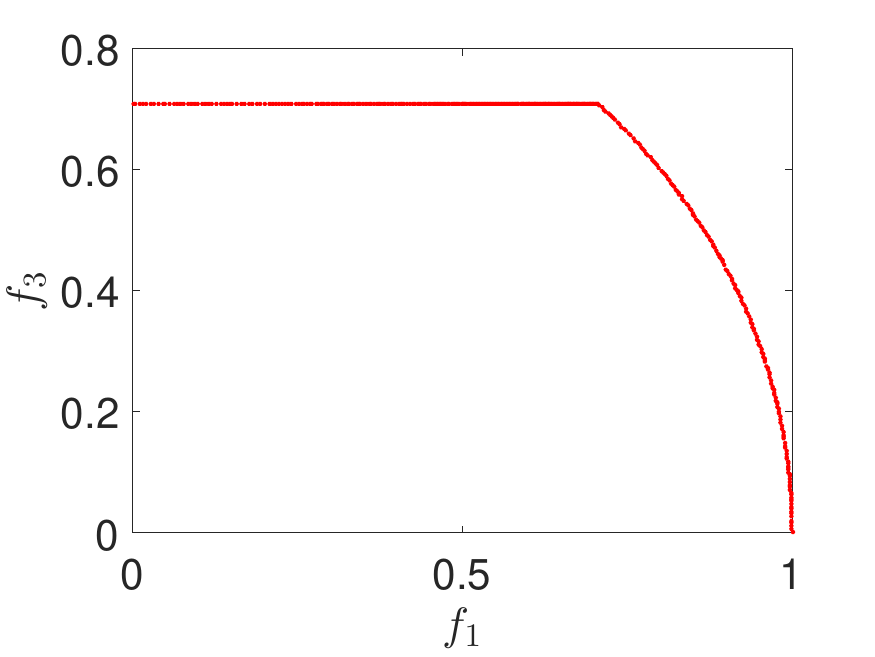}}
\subfigure[]{
\centering
\includegraphics[width=0.2\textwidth]{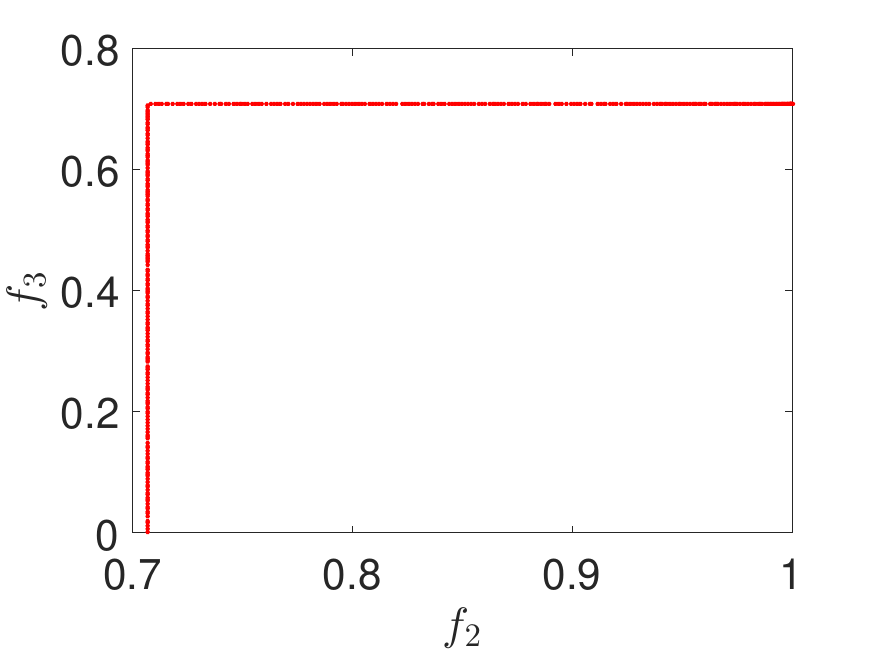}}
\caption{The reference points sampled from the PF of a three-objective problem with two essential objectives defined in (\ref{eq4.1}). (a) The PF in the $3$D objective space. (b) Projection of the PF on the subspace of $f_1$ versus $f_2$. (c) Projection of the PF on the subspace of $f_1$ versus $f_3$. (d) Projection of the PF on the subspace of $f_2$ versus $f_3$.}
\label{fig:4.1}
\end{figure*}

In the real-world applications, the number of essential objectives may be smaller than the number of the objectives of the underlying problem, while the essential objective set is not a subset of the objective set of the problem. For this case, let us first consider a three-objective minimisation problem:

\begin{equation}
\label{eq4.1}
\begin{split}
&f_1(\mathbf{x}) = \gamma_1(\mathbf{x});\\
&f_2(\mathbf{x})  =  \left\{\begin{array}{ll}
\gamma_2(\mathbf{x}), & x_1 < 0.5;\\
\cos(\frac{\pi}{4})(1+g(\mathbf{x}^r)), & \textrm{otherwise},
\end{array}
\right.\\
&f_3(\mathbf{\mathbf{x}})  =  \left\{\begin{array}{ll}
\cos(\frac{\pi}{4})(1+g(\mathbf{x}^r)), & x_1 < 0.5;\\
\gamma_2(\mathbf{x}), & \textrm{otherwise},
\end{array}
\right.\\
&\text{s.t.} \hspace{4mm} 0 \leq x_i \leq 1, \hspace{4mm}\text{for}\hspace{4mm} i = 1, 2, \dots, n
\end{split}
\end{equation}
with
\begin{equation}
\begin{split}
&\gamma_1(\mathbf{x}) = \sin(\frac{\pi}{2}x_1)(1+g(\mathbf{x}^r));\\
&\gamma_2(\mathbf{x}) = \cos(\frac{\pi}{2}x_1)(1+g(\mathbf{x}^r));\\
&g(\mathbf{x}^r) = \sum_{i = 2}^n(x_i - 0.5)^2,\\
\end{split}
\end{equation}
where $n$ is the number of decision variables, and $\mathbf{x}^r = (x_2, \dots, x_n)$. Fig.~\ref{fig:4.1}(a) shows the PF of this problem, from which we can see that the PF lies in a $1$D manifold since the problem has only two essential objectives $\gamma_1$ and $\gamma_2$. Generally, this problem cannot be well solved with the objective selection-based reduction methods. The projections of the original objectives on subspaces $\{f_1, f_2\}$, $\{f_1, f_3\}$, and $\{f_2, f_3\}$ are plotted in Fig.~\ref{fig:4.1}(b)-(d), respectively. From the plots, we can see that only part of the Pareto-optimal solution set (PS) of the original three-objective problem and the PSs of the reduced two-objective problems are overlapping. Nevertheless, the PS of this problem can be obtained by optimising the problem with two essential objectives $\gamma_1$ and $\gamma_2$.

Following the manner in the above example, we assume the essential (conflicting) objectives are $\gamma_1(\mathbf{x}), \gamma_2(\mathbf{x}), \dots, \gamma_d(\mathbf{x})$, and then we construct a minimisation problem with $m (m>d)$ objectives:
\begin{equation}
\label{eq4.2}
\begin{split}
&f_1(\mathbf{x}) = \phi_1(\gamma_1(\mathbf{x}));\\
&\hspace{10mm} \vdots \\
&f_{d-1}(\mathbf{x}) = \phi_{d-1}(\gamma_{d-1}(\mathbf{x}));\\
&f_{d}(\mathbf{x}) = \phi_{d}(\min(\gamma_{d}(\mathbf{x}), \eta_1));\\
&f_{d+1}(\mathbf{x}) = \phi_{d}(\min(\max(\gamma_{d}(\mathbf{x}), \eta_1), \eta_2));\\
&\hspace{10mm} \vdots \\
&f_{m-1}(\mathbf{x}) = \phi_{m-1}(\min(\max(\gamma_{d}(\mathbf{x}), \eta_{m-d-1}), \eta_{m-d}));\\
&f_{m}(\mathbf{x}) = \phi_{m}(\max(\gamma_{d}(\mathbf{x}), \eta_{m-d})),
\end{split}
\end{equation}
where $\phi_1(\cdot), \dots, \phi_m(\cdot)$ are increasing functions, and $\min(\gamma_{d}(\mathbf{x}))< \eta_1 < \eta_2 < \dots < \eta_{m-1} < \max(\gamma_{d}(\mathbf{x}))$. With regard to this case, we have the following theorem.
\begin{theorem}
\label{pro2}
Supposing there exists an optimisation problem with $d$ conflicting objectives $\gamma_1(\mathbf{x}), \dots, \gamma_d(\mathbf{x})$, we construct another problem via (\ref{eq4.2}), then the new problem with the objectives $f_1(\mathbf{x}), \dots, f_m(\mathbf{x})$ has the same PS as the problem with the objectives $\gamma_1(\mathbf{x}), \dots, \gamma_d(\mathbf{x})$.
\end{theorem}
\begin{IEEEproof}
See Appendix \ref{ProofT2}.
\end{IEEEproof}

Since the essential objectives are not all included in the objective set of the problem, this type of problems cannot be well solved with objective selection-based MOEAs. This type of problems is proposed to test the ability of algorithms to extract essential objectives.

\subsection{Partially Redundant Objectives}
\begin{figure}[pbt]
\centering
\subfigure[]{
\centering
\includegraphics[width=0.2\textwidth]{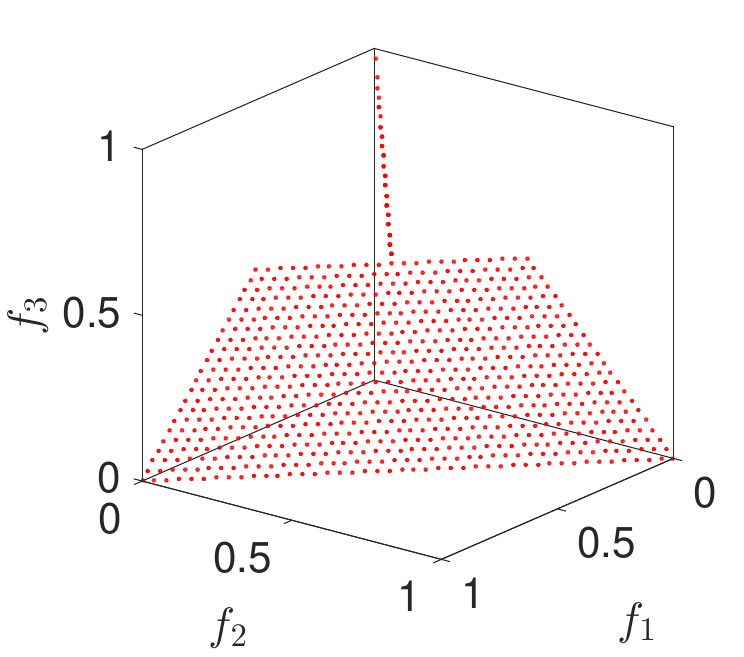}}
\subfigure[]{
\centering
\includegraphics[width=0.18\textwidth]{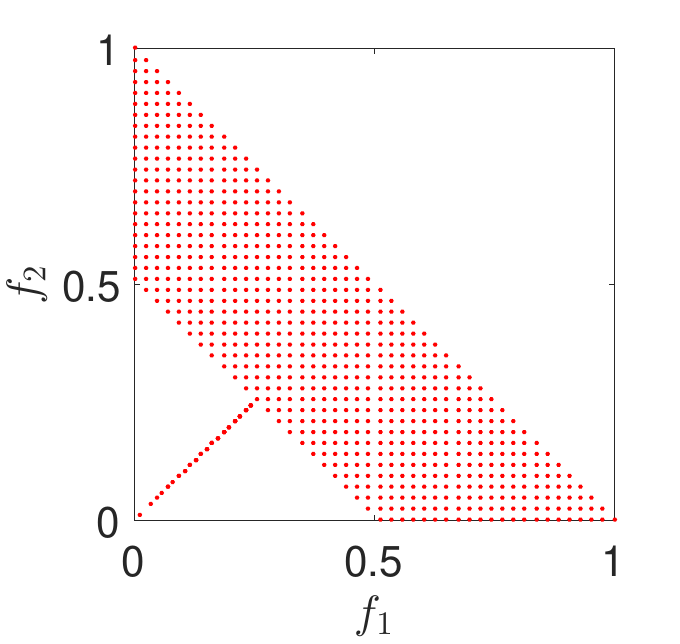}}
\caption{The reference points sampled from the PF of a three-objective problem with partially degenerate PF defined in (\ref{eq4.5}). (a) The PF in the $3$D objective space. (b) Projection of the PF on the subspace of $f_1$ versus $f_2$.}
\label{fig:4.2}
\end{figure}

The correlation between objectives may differ in different regions of the objective space. Two objectives may be harmonious on some parts of the PF, while they are unrelated/conflicting on some other parts of the PF. Let us consider a three-objective minimisation problem:

\begin{equation}
\label{eq4.5}
\begin{split}
&f_1(\mathbf{x})  =  \left\{\begin{array}{ll}
x_1(1-x_2)(1+g(\mathbf{x}^r)), & x_1 > 0.5;\\
\frac{x_1}{2}(1+g(\mathbf{x}^r)), & \textrm{otherwise},
\end{array}
\right.\\
&f_2(\mathbf{x})  =  \left\{\begin{array}{ll}
x_1x_2(1+g(\mathbf{x}^r)), & x_1 > 0.5;\\
\frac{x_1}{2}(1+g(\mathbf{x}^r)), & \textrm{otherwise},
\end{array}
\right.\\
&f_3(\mathbf{x}) = (1-x_1)(1+g(\mathbf{x}^r))\\
&\text{s.t.} \hspace{4mm} 0 \leq x_i \leq 1, \hspace{4mm}\text{for}\hspace{4mm} i = 1, 2, \dots, n
\end{split}
\end{equation}
with
\begin{equation}
g(\mathbf{x}^r) = \sum_{i = 2}^n(x_i - 0.5)^2,
\end{equation}
where $n$ is the number of decision variables, and $\mathbf{x}^r = (x_2, \dots, x_n)^T$. Fig.~\ref{fig:4.2} shows the PF of this problem, from which we can see that $f_1$ and $f_2$ are conflicting at $x_1 < 0.5$, while they are equal at $x_1 \geq 0.5$, and it makes the problem has a degenerate PF at this region, \textit{i.e.}, it leads to a partially degenerate test problem.

This type of test problem is proposed to test the ability of algorithms to discover degenerate segments of PF in the objective space.

Note that there may exist other more complicated types of redundancy and this paper only focuses on these three characteristics.

\section{Test Problems}
\label{Sec:4}
Based on the basic principles and the above three characteristics, we present here a representative set of test problems with degenerate PFs, called as DPF\footnote{The MATLAB code of the proposed test problems is available at https://liangli-zhen.github.io/code/dpf.zip.}. The essential objectives are carefully selected/designed with diverse properties which cover a good representation of various real-world scenarios, such as being multimodal, disconnected, partially separable, biased, and having different shapes of PFs. The characteristics and features of these five test problems are summarised in Table~\ref{tab0}. More test problems can also be constructed by designing different essential objective functions and transforming functions.

\begin{table}[bpt]
\caption{Characteristics and features of the proposed test problems. The correlation denotes the relation between the problem objectives and the essential objectives.}
\label{tab0}
\center
\begin{center}
\begin{tabular}{l|llll}
\hline
Problem & Redundant & Correlation & PF shape & Other features\\
\hline
DPF1 & Explicitly & Linear    & Linear & Multimodal\\
DPF2 & Explicitly & Nonlinear & Mixed & Disconnected\\
DPF3 & Implicitly & Linear & Concave & Bias\\
DPF4 & Implicitly & Nonlinear & Concave & Multimodal\\
DPF5 & Partially  & Linear    & Convex & Partially separable\\
\hline
\end{tabular}
\end{center}
\end{table}

\subsection{Test Problem DPF1}
In the first case, we construct an $m$-objective minimisation problem with $d$ essential objectives. The $m-d$ explicitly redundant objectives are linearly correlated with the $d$ essential objectives. The objective functions of DPF1 are defined as

\begin{equation}
\label{eq5.1}
\begin{split}
&f_1(\mathbf{x}) = \gamma_1(\mathbf{x}) = \frac{1}{2}x_1x_2\dots x_{d-1}(1+g(\mathbf{x}^r));\\
&f_2(\mathbf{x}) = \gamma_2(\mathbf{x})= \frac{1}{2}x_1x_2\dots (1-x_{d-1})(1+g(\mathbf{x}^r));\\
&\hspace{26mm} \vdots\\
&f_d(\mathbf{x}) = \gamma_d(\mathbf{x})= \frac{1}{2}(1-x_1)(1+g(\mathbf{x}^r));\\
&f_{d+1}(\mathbf{x}) = (f_1(\mathbf{x}), f_2(\mathbf{x}), \dots, f_d(\mathbf{x}))\mathbf{u}_1;\\
&\hspace{26mm} \vdots \\
&f_{m}(\mathbf{x}) = (f_1(\mathbf{x}), f_2(\mathbf{x}), \dots, f_d(\mathbf{x}))\mathbf{u}_{m-d},\\
&\text{s.t.} \hspace{4mm} 0 \leq x_i \leq 1, \hspace{4mm}\text{for}\hspace{4mm} i = 1, 2, \dots, n
\end{split}
\end{equation}
with
\begin{equation}
\label{eq5.1a}
g(\mathbf{x}^r) = 100\left[k + \sum_{i = d}^n(x_i - 0.5)^2 - \cos(20\pi(x_i-0.5))\right],
\end{equation}
where $n = d + k - 1$ is the number of decision variables, $k$ (typically set as $10$) denotes the number of elements in $\mathbf{x}^r = (x_d, \dots, x_n)$, and $\mathbf{u}_1, \mathbf{u}_2, \dots, \mathbf{u}_{m-d}$ are column vectors with $d$ non-negative elements.

The essential objectives $f_1(\mathbf{x}), f_2(\mathbf{x}), \dots, f_d(\mathbf{x})$ simply refer to that in DTLZ1~\cite{Deb2005a}. The Pareto optimal solution corresponds to $\mathbf{x}^r = (0.5, \dots, 0.5)$ and the objective vectors lie on the linear hyper-plane: $f_1(\mathbf{x}) + f_2(\mathbf{x})+ \dots + f_d(\mathbf{x}) = \frac{1}{2}$. The difficulty of this problem is to select the essential objectives and converge to the hyper-plane.

\subsection{Test Problem DPF2}
The second constructed $m$-objective minimisation problem has $d$ essential conflicting objectives as well, but the $m-d$ explicitly redundant objectives are non-linearly correlated with the $d$ essential objectives. The problem is to minimise the following objective functions:

\begin{equation}
\label{eq5.2}
\begin{split}
&f_1(\mathbf{x}) = \gamma_1(\mathbf{x}) = x_1;\\
&\hspace{10mm} \vdots\\
&f_{d-1}(\mathbf{x}) = \gamma_{d-1}(\mathbf{x}) = x_{d-1};\\
&f_d(\mathbf{x}) = \gamma_d(\mathbf{x}) \\
&= (d - \sum_{i=1}^{d-1}\frac{f_i(\mathbf{x})}{1+g(\mathbf{x}^r)}(1+\sin(3\pi f_i(\mathbf{x}))))(1+g(\mathbf{x}^r));\\
&f_{d+1}(\mathbf{x}) = \phi_1((f_1(\mathbf{x}), f_2(\mathbf{x}), \dots, f_d(\mathbf{x}))\mathbf{u}_1);\\
&\hspace{10mm} \vdots \\
&f_{m}(\mathbf{x}) = \phi_{m-d}((f_1(\mathbf{x}), f_2(\mathbf{x}), \dots, f_d(\mathbf{x}))\mathbf{u}_{m-d}),\\
&\text{s.t.} \hspace{4mm} 0 \leq x_i \leq 1, \hspace{4mm}\text{for}\hspace{4mm} i = 1, 2, \dots, n
\end{split}
\end{equation}
with
\begin{equation}
\label{eq5.2a}
g(\mathbf{x}^r) = 1+\frac{9}{k}\sum_{i=d}^{n}x_i,
\end{equation}
where $n = d + k - 1$ is the number of decision variables, $k$ (typically set as $20$) denotes the number of elements in $\mathbf{x}^r  = (x_d, \dots, x_n)$, $\mathbf{u}_1, \mathbf{u}_2, \dots, \mathbf{u}_{m-d}$ are column vectors with $d$ non-negative elements, and $\phi_{1}, \dots, \phi_{m-d}$ are nonlinearly and nondecreasingly mapping functions.

\subsection{Test Problem DPF3}
The third problem is an $m$-objective minimisation problem with $d$ essential objectives. The redundant objectives do not explicitly exist in the problem, but implicitly exist as follows
\begin{equation}
\label{eq5.3}
\begin{split}
&f_1(\mathbf{x}) =\gamma_1(\mathbf{x});\\
&\hspace{10mm} \vdots \\
&f_{d-1}(\mathbf{x}) = \gamma_{d-1}(\mathbf{x});\\
&f_{d}(\mathbf{x}) = \min(\gamma_{d}(\mathbf{x}), \eta_1);\\
&f_{d+1}(\mathbf{x}) = \min(\max(\gamma_{d}(\mathbf{x}), \eta_1), \eta_2);\\
&\hspace{10mm} \vdots \\
&f_{m-1}(\mathbf{x}) = \min(\max(\gamma_{d}(\mathbf{x}), \eta_{m-d-1}), \eta_{m-d});\\
&f_{m}(\mathbf{x}) = \max(\gamma_{d}(\mathbf{x}), \eta_{m-d}),\\
&\text{s.t.} \hspace{4mm} 0 \leq x_i \leq 1, \hspace{4mm}\text{for}\hspace{4mm} i = 1, 2, \dots, n
\end{split}
\end{equation}
with
\begin{equation}
\label{eq5.3a}
\begin{split}
&\gamma_1(\mathbf{x}) = (1-\cos(\theta_1)\dots \cos(\theta_{d-1}))(1+g(\mathbf{x}^r));\\
&\gamma_2(\mathbf{x}) = (1-\cos(\theta_1)\dots \cos(\theta_{d-2})\sin(\theta_{d-1}))(1+g(\mathbf{x}^r));\\
&\hspace{26mm} \vdots\\
&\gamma_{d-1}(\mathbf{x}) = (1 - \cos(\theta_1)\sin(\theta_2))(1+g(\mathbf{x}^r));\\
&\gamma_d(\mathbf{x}) =(1 - \sin(\theta_1))(1+g(\mathbf{x}^r));\\
&g(\mathbf{x}^r) = \sum_{i = d}^n(x_i - 0.5)^2;\\
&\theta_j = \frac{\pi}{2} x_j^{100} \hspace{4mm}\text{for}\hspace{4mm} j\in \{1, \dots, d-1\},
\end{split}
\end{equation}
where $n = d + k - 1$ is the number of decision variables, $k$ (typically set as $10$) denotes the number of elements in $\mathbf{x}^r  = (x_d, \dots, x_n)$, and $\min(\gamma_{d}(\mathbf{x}))< \eta_1 < \eta_2 < \dots < \eta_{m-d} < \max(\gamma_{d}(\mathbf{x}))$. Please note that the obtained objective functions $f_{d}, \dots, f_m$ are partially linear with the essential objective functions $\gamma_1, \dots, \gamma_{d}$ in DPF3. From the definition of the essential objective functions, we can see that the search space has a variable density of solutions due to the bias transforming on the decision variables. The essential objective $\gamma_d(\mathbf{x})$ is locally linearly correlated with the last $m-d+1$ objectives of the defined problem.

\subsection{Test Problem DPF4}
In the fourth case, we construct an $m$-objective minimisation problem with $d$ essential objectives. Different from DPF3, the objectives of this problem are nonlinearly correlated with the essential objectives as

\begin{equation}
\label{eq5.4}
\begin{split}
&f_1(\mathbf{x}) = \phi_1(\gamma_1(\mathbf{x}));\\
&\hspace{10mm} \vdots \\
&f_{d-1}(\mathbf{x}) = \phi_{d-1}(\gamma_{d-1}(\mathbf{x}));\\
&f_{d}(\mathbf{x}) = \phi_d(\min(\gamma_{d}(\mathbf{x}), \eta_1));\\
&f_{d+1}(\mathbf{x}) = \phi_{d+1}(\min(\max(\gamma_{d}(\mathbf{x}), \eta_1), \eta_2));\\
&\hspace{10mm} \vdots \\
&f_{m-1}(\mathbf{x}) = \phi_{m-1}(\min(\max(\gamma_{d}(\mathbf{x}), \eta_{m-d-1}), \eta_{m-d}));\\
&f_{m}(\mathbf{x}) = \phi_m(\max(\gamma_{d}(\mathbf{x}), \eta_{m-d}))\\
&\text{s.t.} \hspace{4mm} 0 \leq x_i \leq 1, \hspace{4mm}\text{for}\hspace{4mm} i = 1, 2, \dots, n
\end{split}
\end{equation}
with
\begin{equation}
\label{eq5.4a}
\begin{split}
&\gamma_1(\mathbf{x}) = (1-\cos(\theta_1)\dots \cos(\theta_{d-1}))(1+g(\mathbf{x}^r));\\
&\gamma_2(\mathbf{x}) = (1-\cos(\theta_1)\dots \cos(\theta_{d-2})\sin(\theta_{d-1}))(1+g(\mathbf{x}^r));\\
&\hspace{26mm} \vdots\\
&\gamma_{d-1}(\mathbf{x}) = (1 - \cos(\theta_1)\sin(\theta_2))(1+g(\mathbf{x}^r));\\
&\gamma_d(\mathbf{x}) =(1 - \sin(\theta_1))(1+g(\mathbf{x}^r));\\
&g(\mathbf{x}^r) = 100\left[k + \sum_{i = d}^n(x_i - 0.5)^2 - \cos(20\pi(x_i-0.5))\right];\\
&\theta_j = \frac{\pi}{2} x_j \hspace{4mm}\text{for}\hspace{4mm}  j\in \{1, \dots, d-1\},
\end{split}
\end{equation}
where $n = d + k - 1$ is the number of decision variables, $k$ (typically set as $10$) denotes the number of elements in $\mathbf{x}^r = (x_d, \dots, x_n)$, $\min(\gamma_{d}(\mathbf{x}))< \eta_1 < \eta_2 < \dots < \eta_{m-d} < \max(\gamma_{d}(\mathbf{x}))$, and $\phi_{1}, \dots, \phi_{m}$ are nonlinear and nondecreasing mapping functions. This problem is multimodal as $g(\mathbf{x}^r)$ has $11^k - 1$ local minima. Extracting essential objectives in this problem is more difficult than that in DPF3 because of the nonlinear correlation between the problem objectives and the essential objectives.

\begin{figure}[pbt]
\centering
\subfigure[]{
\centering
\includegraphics[width=0.2\textwidth]{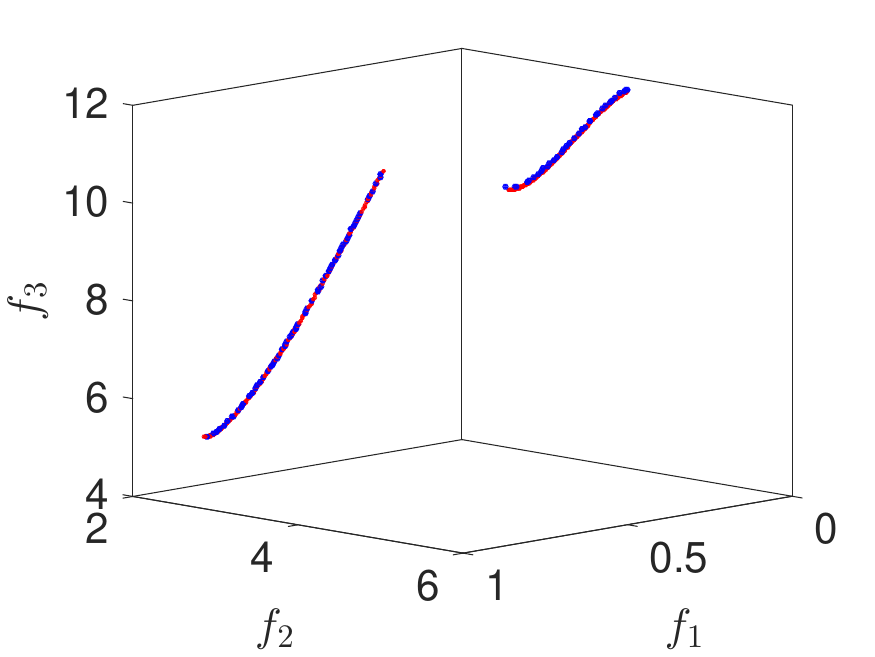}}
\subfigure[]{
\centering
\includegraphics[width=0.2\textwidth]{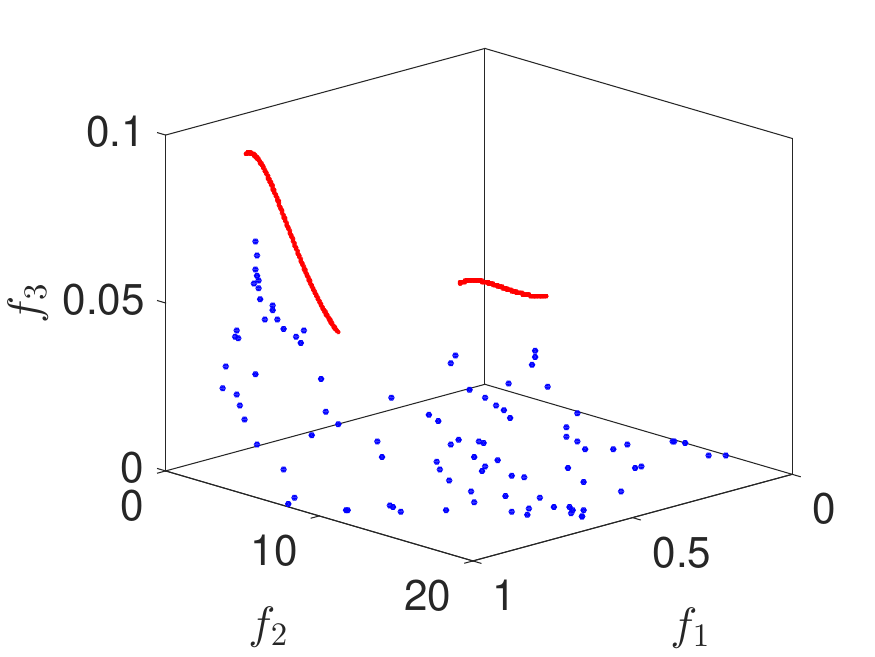}}
\caption{The solution set obtained by NSGA-II after $100$ generations optimisation on DPF2 with quadratic and sigmoid transforming functions in (a) and (b), respectively, where the red points are sampled from the PF and the blue dots denote the NSGA-II solutions.}
\label{fig:DPF2T}
\end{figure}

\subsection{Test Problem DPF5}
In the fifth case, we construct a minimisation problem whose correlations between the objectives differ in different PF segments as
\begin{equation}
\label{eq5.5}
\begin{split}
&f_1(\mathbf{x})  =  \left\{\begin{array}{lr}
\beta_1(\mathbf{x}), & x_1 < \frac{1}{3},\\
\gamma_1(\mathbf{x}), & \textrm{otherwise};
\end{array}
\right.\\
&\hspace{16mm} \vdots\\
&f_{m-d+1}(\mathbf{x})  =  \left\{\begin{array}{lr}
\beta_{m-d+1}(\mathbf{x}), & x_1 < \frac{1}{3},\\
\gamma_1(\mathbf{x}), & \textrm{otherwise};
\end{array}
\right.\\
&f_{m-d+2}(\mathbf{x}) = \gamma_{2}(\mathbf{x});\\
&\hspace{16mm} \vdots\\
&f_m(\mathbf{x}) = \gamma_d(\mathbf{x});\\
&\text{s.t.} \hspace{4mm} 0 \leq x_i \leq 1, \hspace{4mm}\text{for}\hspace{4mm} i = 1, 2, \dots, n
\end{split}
\end{equation}
with
\begin{equation}
\label{eq5.5a}
\begin{split}
&\beta_1(\mathbf{x}) = \cos(\theta_1)\dots \cos(\theta_{m-2})\cos(\theta_{m-1})(1+g(\mathbf{x}^r));\\
&\beta_2(\mathbf{x}) = \cos(\theta_1)\dots \cos(\theta_{m-2})\sin(\theta_{m-1})(1+g(\mathbf{x}^r));\\
&\hspace{26mm} \vdots\\
&\beta_{m-d+1}(\mathbf{x}) = \cos(\theta_1)\dots \cos(\theta_{{d-1}})\sin(\theta_{{d}})(1+g(\mathbf{x}^r));\\
&\gamma_1(\mathbf{x}) =  \sqrt{\frac{1}{m-d+1}}\cos(\theta_1)\dots \cos(\theta_{d-1})(1+g(\mathbf{x}^r));\\
&\gamma_2(\mathbf{x}) = \cos(\theta_1)\dots \cos(\theta_{d-2})\sin(\theta_{d-1})(1+g(\mathbf{x}^r));\\
&\hspace{26mm} \vdots\\
&\gamma_{d-1}(\mathbf{x}) = \cos(\theta_1)\sin(\theta_2)(1+g(\mathbf{x}^r));\\
&\gamma_d(\mathbf{x}) = \sin(\theta_1)(1+g(\mathbf{x}^r));\\
&g(\mathbf{x}^r) = (x_m - x_1)^2 + \sum_{i = m + 1}^n(x_i - 0.5)^2 ;\\
&\theta_j = \frac{\pi}{2} x_j \hspace{4mm}\text{for}\hspace{4mm}  j\in \{1, \dots, m-1\},
\end{split}
\end{equation}
where $n = d + k - 1$ is the number of decision variables, $k$ (typically set as $10$) denotes the number of elements in $\mathbf{x}^r = (x_m, \dots, x_n)$. From the definition of DPF5, we can see that the objective vectors of the PF satisfy that $f_1^2 + f_2^2 +\dots + f_m^2 = 1$, and the decision variables are partially separable. The PF of this problem contains a degenerate Pareto-optimal segment of $(d-1)$ dimensions and a non-degenerate Pareto-optimal segment of $(m-1)$ dimensions.

\subsection{Setting of Parameters and Mapping Functions in DPF}

\begin{figure*}[bpt]
  \centering
\subfigure[DPF1]{
\centering
\includegraphics[width=0.18\textwidth]{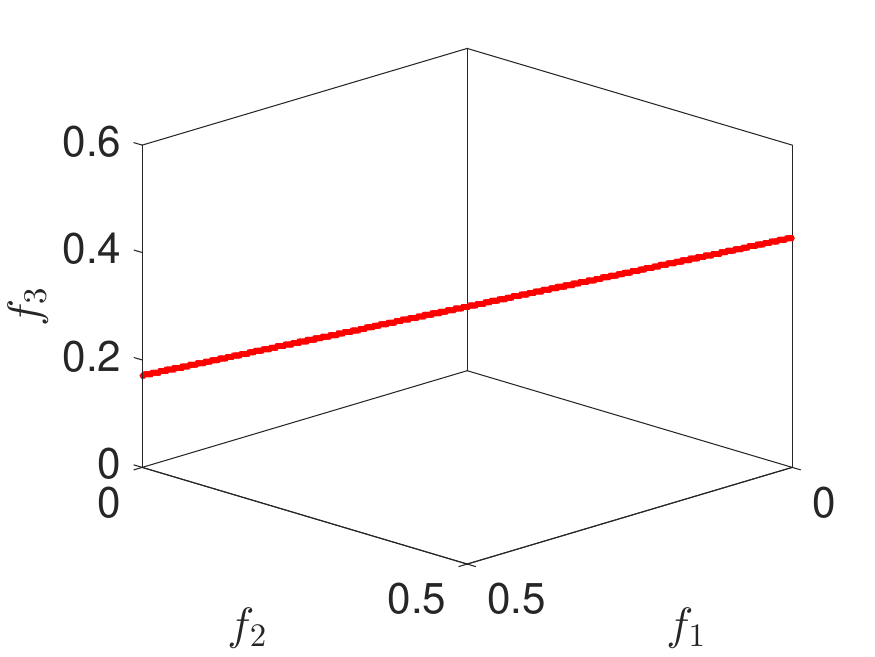}}
\subfigure[DPF2]{
\centering
\includegraphics[width=0.18\textwidth]{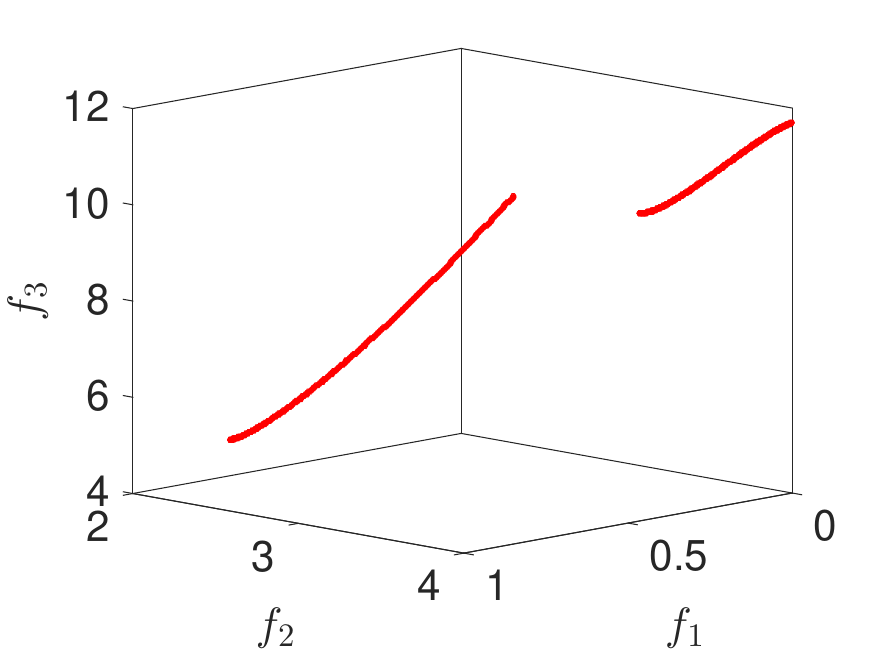}}
\subfigure[DPF3]{
\centering
\includegraphics[width=0.18\textwidth]{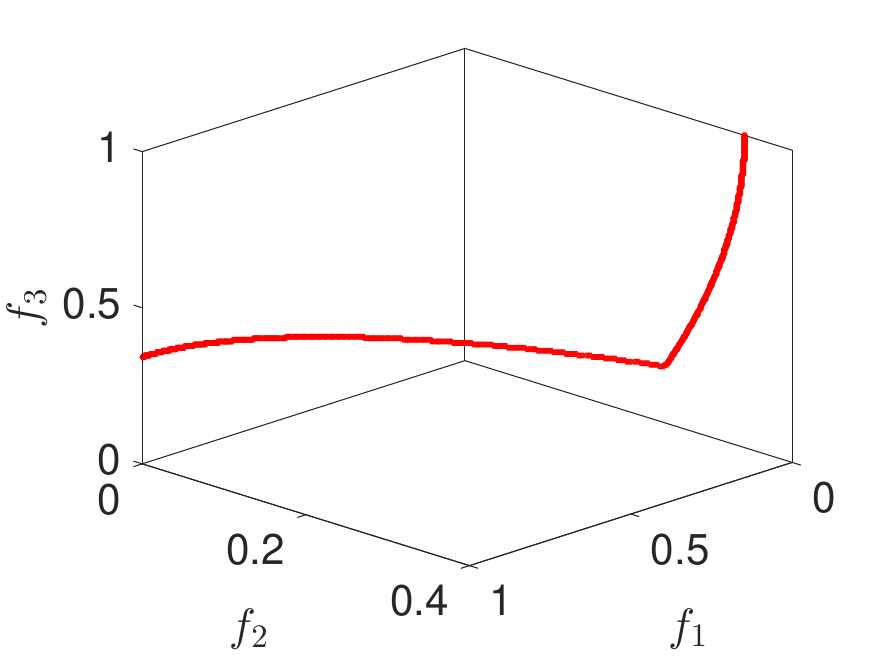}}
\subfigure[DPF4]{
\centering
\includegraphics[width=0.18\textwidth]{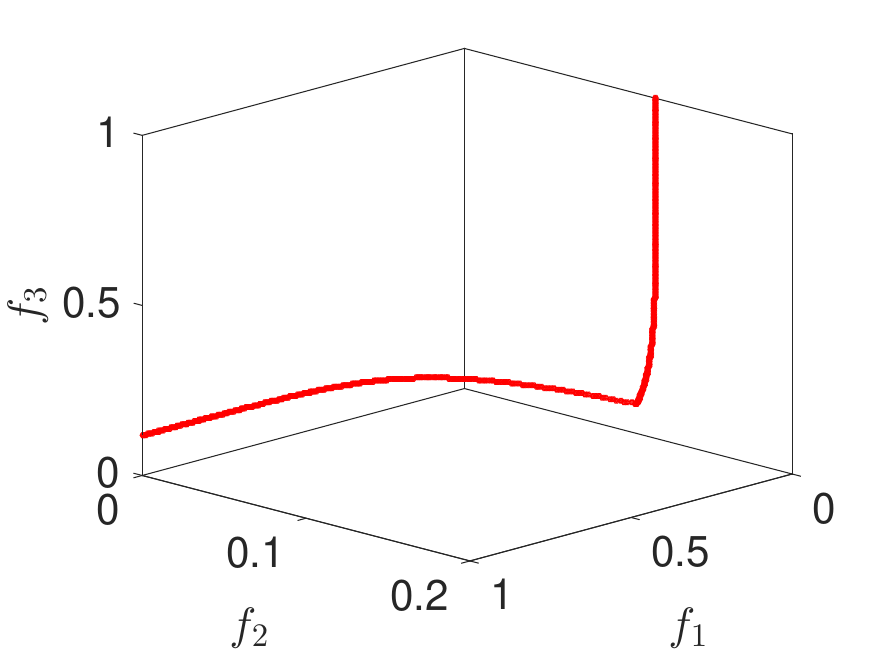}}
\subfigure[DPF5]{
\centering
\includegraphics[width=0.18\textwidth]{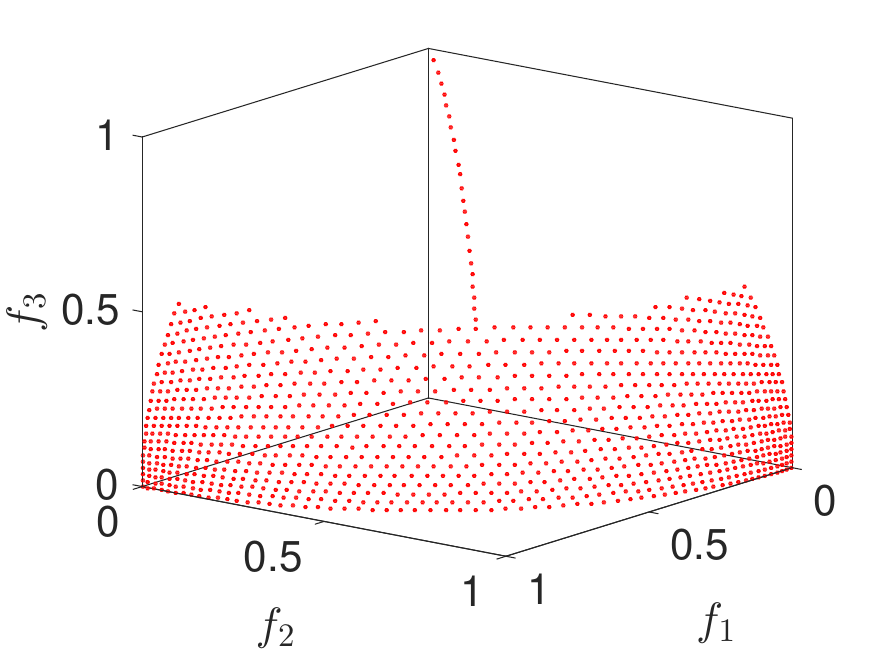}}
\caption{The reference points sampled from the PFs of $3$-objective test problems DPF1 to PDF5 that are with $2$ essential objectives.}
\label{fig:DPFPF}
\end{figure*}

We assign randomly generated numbers to the elements of $\mathbf{u}_1, \dots, \mathbf{u}_{m-d}$ in DPF1-DPF2 and the parameters $\eta_1, \dots, \eta_{m-d}$ in DPF3-DPF4. Despite that the test instances are easy to implement, they may differ in different runs due to the randomly generated parameters, which makes the comparisons of the statistical results hard. To guarantee that these parameters are unchanged in different independent runs, we use a chaos-based pseudo random number generator by following \cite{Cheng2016}. The generated numbers are
\begin{equation}
\label{eq5.6a}
c_i  =  \alpha \times c_{i-1} \times (1-c_{i-1}) \hspace{4mm}\text{for}\hspace{4mm}  i \in \mathbb{Z}^+,
\end{equation}
where $\alpha$ and $c_0$ are parameters for the logistic map in (\ref{eq5.6a}), and typically set as $3.8$ and $0.1$, respectively. The generated numbers are sequentially assigned to the elements of $\mathbf{u}_1, \dots, \mathbf{u}_{m-d}$ in DPF1-DPF2, and the parameters $\eta_1, \dots, \eta_{m-d}$ in DPF3-DPF4 are set as the increasingly sorted results of the generated numbers.

To preserve the dominance relation between the decision vectors, the nonlinearly mapping functions in DPF2 and DPF4 have to be non-decreasing functions and increasing functions, respectively. The choice of the mapping functions is critical for the test problems. Many widely-used increasing functions can be selected to construct the instances of our proposed test problems. However, it is notable that different mapping functions induce different levels of difficulties in the test problems.

Fig.~\ref{fig:DPF2T} shows two solution sets obtained by NSGA-II on DPF2 with the mappings of the quadratic function

\begin{equation}
\label{eq5.2c}
\phi(\tau) = {\tau}^2,
\end{equation}
and the sigmoid function
\begin{equation}
\label{eq5.2b}
\phi(\tau) = \frac{e^{\tau}}{1+e^{\tau}}.
\end{equation}

From Fig.~\ref{fig:DPF2T}(a), we can see that NSGA-II can obtain a solution set that has a good convergence and diversity to the PF of the problem. In Fig.~\ref{fig:DPF2T}(b), however, it is clear that most of the solutions are far from the PF. It illustrates that DPF2 with the sigmoid function is more difficult to be optimised than DPF2 with the quadratic function. The potential reason may be that the input values of the sigmoid function are mapped to values that are approximately equal to one, which decreases the distinction between two different inputs. In this paper, we adopt the quadratic function as the nonlinearly mapping function for DPF2 and DPF4.

Fig.~\ref{fig:DPFPF} shows the scatter plots of the PFs of DPF1 to DPF5 with $m=3$ and $d=2$. From the results, we can see that the PFs of DP1-DPF4 lie in a $1$D manifold, and the PF of DPF5 contains a curve and a part of a $2$D spherical surface. Furthermore, this test suite has a variety of features, \textit{i.e.}, the Pareto optimal geometry, modality, PF shape, etc, and a set of recommendations, \textit{i.e.}, scalable number of objectives and variables, Pareto optima known, dissimilar trade-off ranges, etc.
At last, the parameters of $m$ (the number of objectives) and $d$ (the number of essential objectives) should be positive integers, and satisfy the condition: $m > d > 1$.

\section{Computational Studies}
\label{Sec:5}
This section is devoted to the experimental investigation of the proposed test problems, with a focus on their difference from existing degenerate problems. To do so, we first examine the performance of ten state-of-the-art MOEAs, most of which have been found to be promising on the existing degenerate problems. Then, we look into the impact of the proposed three characteristics and compare the proposed problems with a widely-used degenerate problem by demonstrating the performance difference of five algorithms on them.

\subsection{Tested MOEAs}
In the experiments, ten MOEAs are tested on the proposed test problems, including classical MOEAs, the algorithms specially designed for many-objective optimisation problems (MaOPs), and MOEAs that are based on objective reduction. These ten MOEAs are the nondominated sorting genetic algorithm II (NSGA-II)~\cite{Deb2002}, the strength Pareto evolutionary algorithm 2~\cite{Zitzler2001} with the shift-based density estimation (SDE) strategy~\cite{Li2014a} (SPEA2+SDE), the nondominated sorting genetic algorithm III (NSGA-III)~\cite{Deb2014}, the multiobjective evolutionary algorithm based on decomposition (MOEA/D)~\cite{Zhang2007}, the improved decomposition-based evolutionary algorithm (IDBEA)~\cite{Asafuddoula2015}, the reference vector guided evolutionary algorithm (RVEA)~\cite{Cheng2016b}, the indicator-based evolutionary algorithm (IBEA)~\cite{zitzler2004indicator}, the algorithm for $\delta$ minimum objective subset problem ($\delta$-MOSS)~\cite{Brockhoff2007}, the objective space participation based evolutionary algorithm (OSP)~\cite{jaimes2014objective} and the objective reduction algorithm based on nonlinear correlation information entropy (NCIE)~\cite{Wang2016}. Please note that $\delta$-MOSS, OSP and NCIE are objective reduction methods and they are incorporated into NSGA-II to obtain the PS of MaOPs in our experiments.


\subsection{Parameter Settings for Tested MOEAs}
A simulated binary crossover (SBX) with the probability $p_c = 1.0$ and a polynomial mutation with the probability $p_m = \frac{1}{n}$ (where $n$ denotes the number of decision variables) are used for all MOEAs, and their distribution indexes are both set as $20$ as recommended in \cite{Deb2014}. The parameters in the MOEAs are set by following the suggestion in their original papers. MOEA/D has two commonly-used achievement scalarising functions, Tchebycheff and penalty-based boundary intersection (PBI). In this study, we use the later one and set the neighbourhood size as $\frac{n}{10}$ and the penalty parameter as $5.0$. For IBEA, we set the parameter $\kappa$ as $0.05$. For RVEA, the adaptive frequency $f_r$ and the parameter $\alpha$ are set as $0.1$ and $2.0$, respectively. The parameter $\delta$ is set to $0.2$ for $\delta$-MOSS. Since the algorithms $\delta$-MOSS, OSP and NCIE are incorporated into NSGA-II, we execute the objective reduction for every $10$ generations in NSGA-II.

\subsection{Performance Metrics}
To compare the performance of the MOEAs on the proposed test problems, the hypervolume (HV)~\cite{bader2011hype} is adopted in the experiments. HV (also known as the Lebesgue measure or S metric) computes the volume of the objective space between the obtained solution set and a given reference point. It can provide combined information about convergence and diversity of a solution set~\cite{bader2011hype}. The selection of the reference point is crucial for the calculation of HV. Choosing a reference point whose elements are slightly larger than the upper bound of the corresponding objectives on the PF is an acceptable strategy since the effects of convergence and diversity of the set can be well-balanced~\cite{auger2009theory}. In our experiments, we set the values of the elements of the reference point as the results of $1.1$ times the upper bound of the corresponding objectives on the PF. Since the computational complexity is so high that only a solution set with a maximum of four or five objectives is tractable within reasonable time limits, we approximately estimate the HV result of a solution set by the Monte Carlo method used in ~\cite{bader2011hype} with $10^7$ sampling points. To better understand the performance of the MOEAs, we also calculate a representative set, which is constructed by $1000$ reference vectors sampled from the PF. Then, we report the ratio of the HV values between the solution set obtained by the MOEA and the representative set, denoted as HVR. From the investigation on quality evaluation of solution sets in multiobjective optimisation~\cite{li2019quality}, it recommends evaluating the quality by viewing the results visually if it is available.

\subsection{Results of MOEAs}
In the proposed test problems, three parameters should be provided, \textit{i.e.}, the number of objectives $m$, the number of essential objectives $d$, and the number of decision variables $n$. In the first experiment, we test the instances with $m = 3, 6, 10$ and $d = 2, 3, 5$, respectively. The number of decision variables is set as the recommended value as stated in the corresponding problems. The maximum number of generations is taken as the termination condition, which is set to $1000$ for all tested algorithms. For MOEA/D and RVEA, the population size is determined by the simplex-lattice design factor and the number of objectives. We follow the setting in \cite{Cheng2016b} where the population size is specified to $105, 132$ and $275$ for the problems with $3, 6$, and $10$ objectives, respectively. In this experiment, $31$ repetitions of Monte Carlo simulation are conducted on each instance for each algorithm, and the statistical results of the HVR values of the ten MOEAs are reported in Table~\ref{tab2}. The statistical results of the ten MOEAs in terms of the inverted generational distance (IGD)~\cite{Bosman2003, coello2004study} and generational distance (GD)~\cite{van1998multiobjective} metrics can be found in Supplementary Material\footnote{The Supplementary Material document of this paper is available at https://liangli-zhen.github.io/papers/DPF-S.pdf.} (Section II). From the results in Table~\ref{tab2}, we have the following observations.

For three-objective test instances, all of the tested algorithms perform well in most cases. The Pareto-dominance-based algorithms NSGA-II and SPEA2+SDE achieve slightly higher HVR values than other algorithms on DPF1-DPF4. The objective reduction-based algorithms OSP and $\delta$-MOSS perform as well as NSGA-II on DPF1 and DPF2, while they are inferior to NSGA-II on DPF4. This is due to the fact that these two objective reduction-based algorithms, which select a subset of the original objective set as the criteria to optimise the problem, may not be able to work well for implicitly redundant objectives which DPF3 and DPF4 are designed with. The decomposition-based methods, MOEA/D, RVEA and IDBEA perform worse than Pareto dominance-based algorithms on these degenerate problems as only a small proportion of the weight vectors\footnote{They are used to emphasize the importance of different objective functions in the scalar optimization problem for the decomposition-based methods.}, correspond to the PF.

For six-objective test instances, SPEA2+SDE obtains the highest HVR values on DPF1 and DPF2. IBEA achieves the best results on DPF3 and DPF5, and NSGA-II outperforms the others on DPF4. In addition, the objective reduction-based methods, OSP and NCIE, can obtain competitive performance compared with the best-performing algorithms on DPF1 and DPF2, where the essential objectives are explicitly included in the objective set of the problem. However, they are much inferior to the best-performing algorithms on DPF3 and DPF4, where the essential objectives are not explicitly included. A good objective reduction for such problems needs to be able to extract the essential objectives instead of selecting them from the objective set of the problem in DPF3 and DPF4. The decomposition-based methods MOEA/D, RVEA and IDBEA obtain lower HVR values than the Pareto dominance-based algorithms on DPF2 and DPF4, where the problem objectives are nonlinearly dependent on the essential objectives. The decomposition-based algorithms have a smaller proportion of the weight vectors corresponding to the PF in this six-dimensional objective space than in the three-dimensional objective space.

In terms of the ten-objective test instances, SPEA2+SDE achieves the highest HVR values on DPF1 and DPF2, IBEA on DPF3 and DPF5, and NSGA-III on DPF4. The methods of $\delta$-MOSS, NSGA-II and NCIE perform poorly on DPF5. In the objective reduction-based algorithms, OSP performs better than NSGA-II on DPF1, DPF2 and DPF5, but it is inferior to its integrated method on DPF3 and DPF4. $\delta$-MOSS outperforms NSGA-II, but it fails to converge onto the PF on DPF5. NCIE obtains lower HVR values than NSGA-II on DPF1-DPF5. IDBEA obtains much higher HVR values than MOEA/D on DPF1, DPF3 and DPF4, but lower values on DPF2 and DPF5. NSGA-III, which combines the ideas of the decomposition-based approach and the Pareto dominance-based approach, achieves the highest HVR value on DPF4.

\begin{figure*}[hbpt]
\centering
\subfigure[SPEA2+SDE on DPF1]{
\centering
\includegraphics[width=0.18\textwidth]{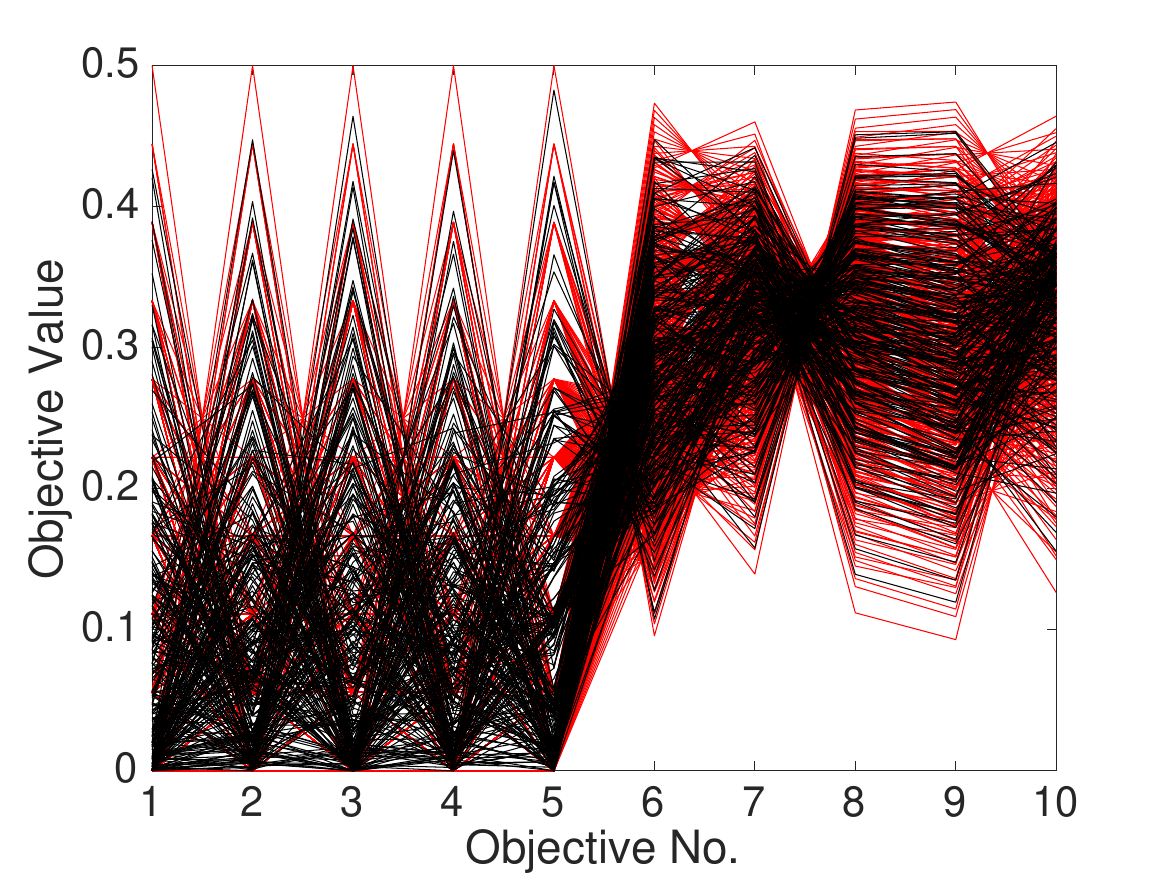}}
\subfigure[SPEA2+SDE on DPF2]{
\centering
\includegraphics[width=0.18\textwidth]{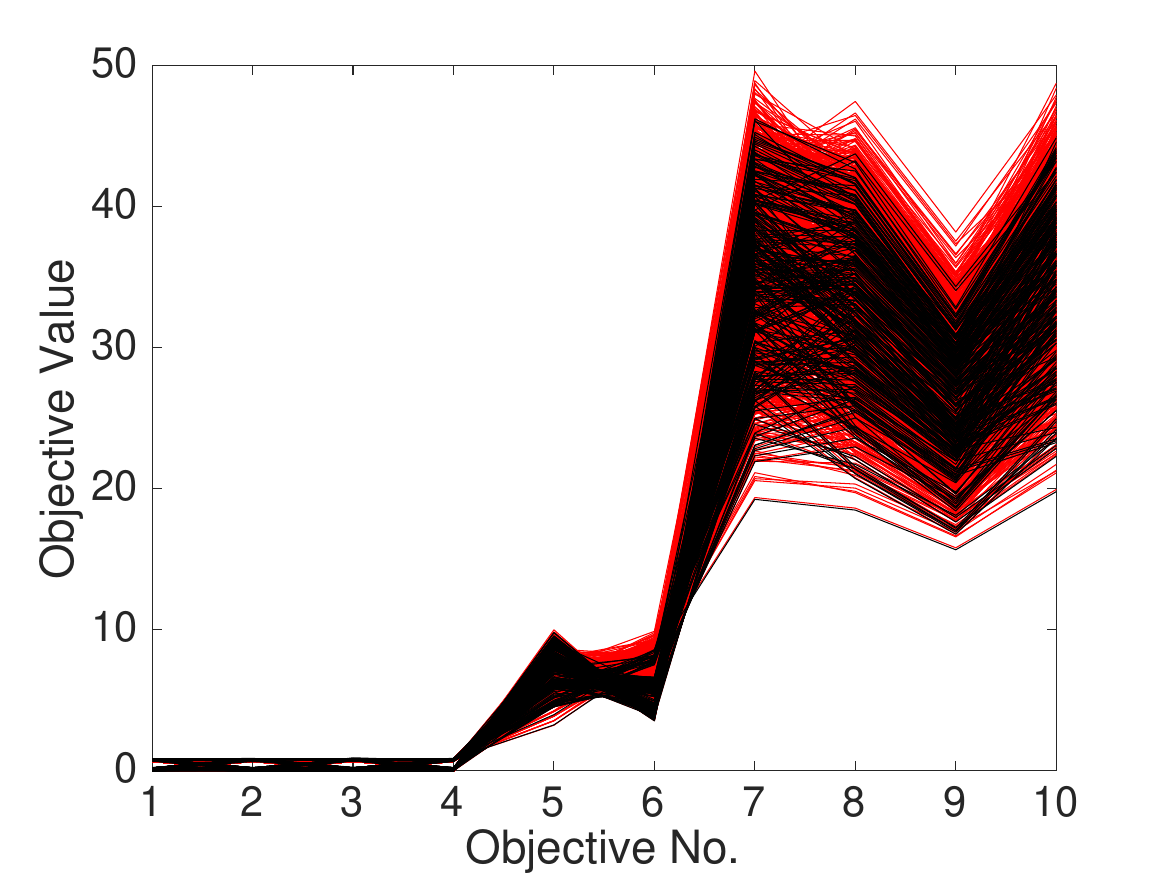}}
\subfigure[IBEA on DPF3]{
\centering
\includegraphics[width=0.18\textwidth]{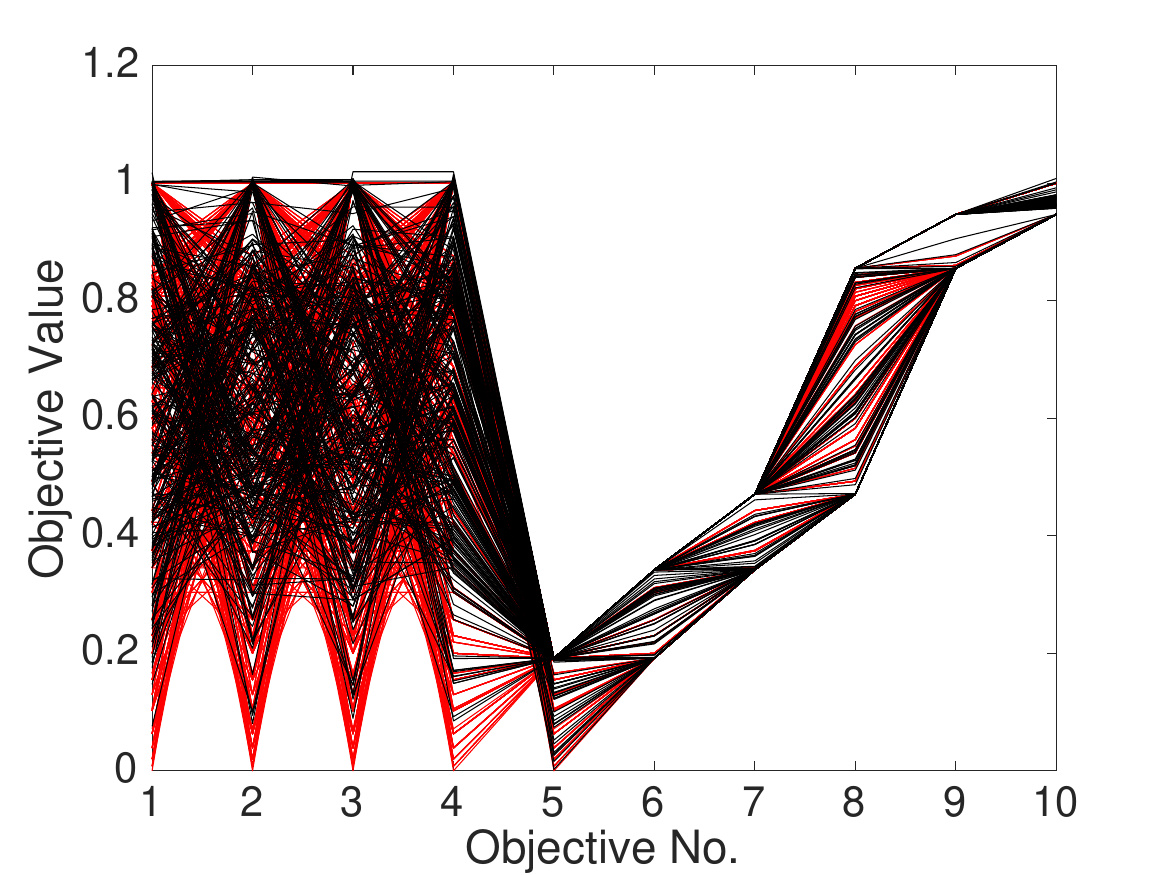}}
\subfigure[NSGA-III on DPF4]{
\centering
\includegraphics[width=0.18\textwidth]{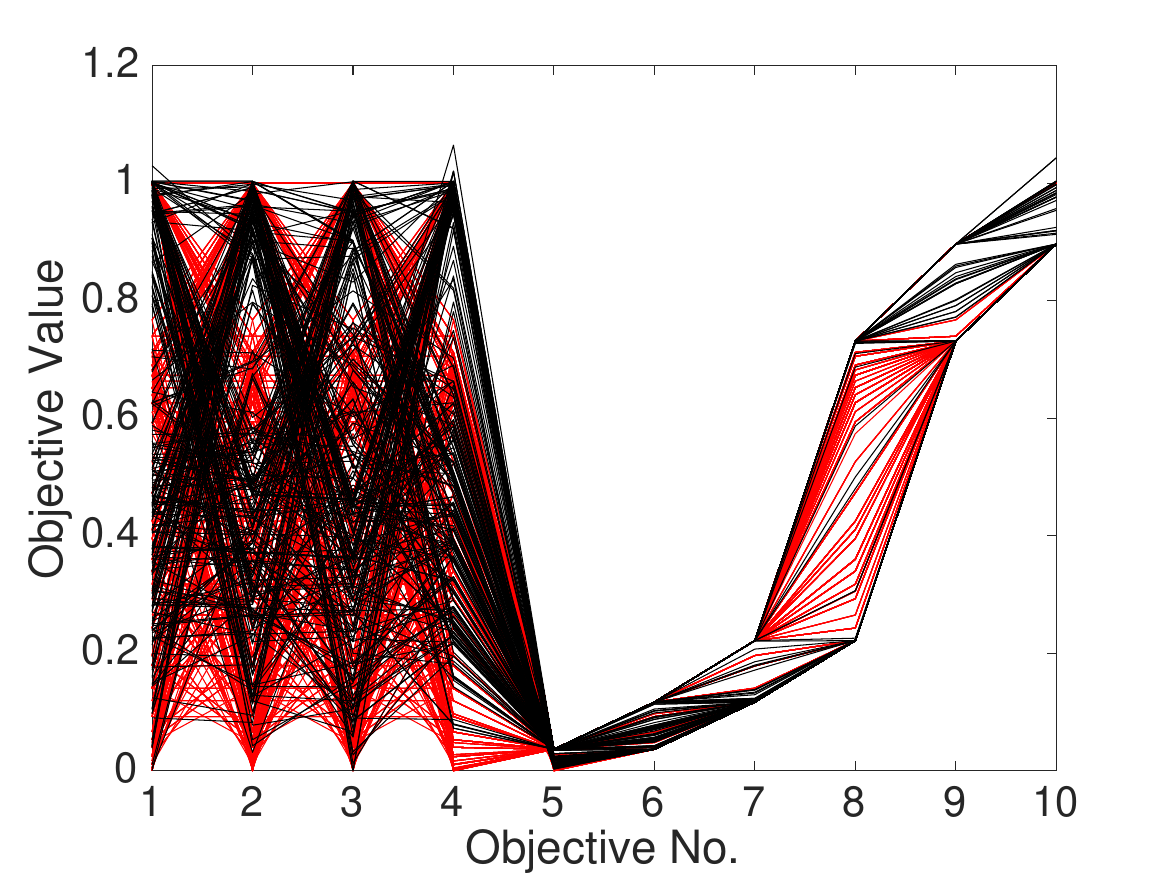}}
\subfigure[IBEA on DPF5]{
\centering
\includegraphics[width=0.18\textwidth]{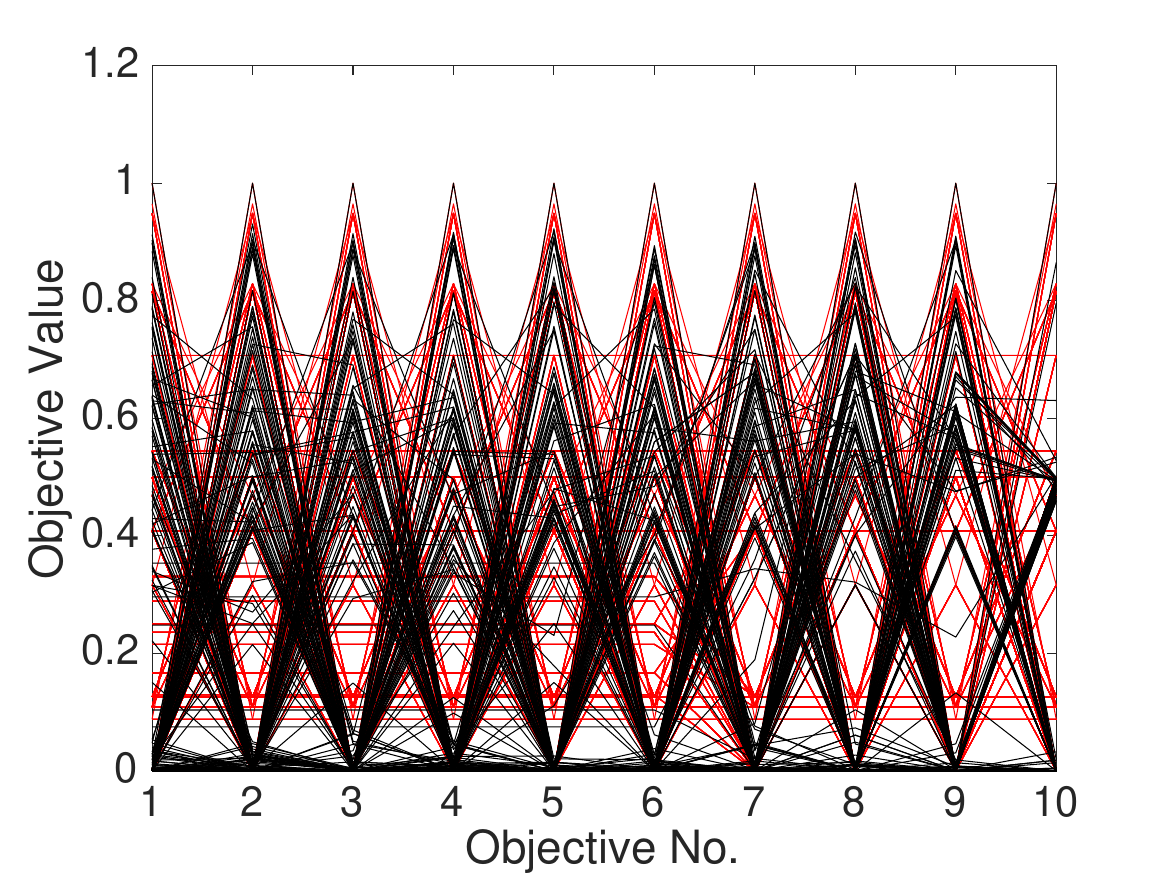}}
\caption{Nondominated solutions obtained by the algorithms that achieved the best mean HV values on ten-objective DPF1 to PDF5 with $d=5$ in the run associated with its best HV value. The red dot lines represent the reference points sampled from the PFs and the black lines denote the solutions obtained by the MOEAs after $2,000$ generations optimisation.}
\label{fig:DPFM10}
\end{figure*}

To visually understand the performance of the tested algorithms on the proposed method, we also show the parallel coordinates plot\footnote{The approach of sampling reference points~\cite{tian2018} on the true PFs can be seen in Supplementary Material (Section I).} of the results obtained by the algorithms that achieved the best HV values in the $31$ runs in Fig.~\ref{fig:DPFM10}. From the results, we can see that even though SPEA2+SDE obtains the best HV value on DPF1 and DPF2, it fails to find the solutions on the boundary of the PFs (see Fig.~\ref{fig:DPFM10}(a)-(b)). From Fig.~\ref{fig:DPFM10}(c), we can see that a large portion of the PF of DPF3 is not overlapping with the results obtained by IBEA. Even though NSGA-III achieves the best HV value on DPF4, its solution set has a poor diversity over the PF, which can be seen from the objective values of its results on the $8$-th objective in Fig.~\ref{fig:DPFM10}(d). It can also be seen that only a small number of solutions obtained by IBEA cover the degenerate part of the PF on DPF5 since the solutions on the degenerate PF should have the same value on the first five objectives in Fig.~\ref{fig:DPFM10}(e)

\begin{table*}[bpt]
\caption{The statistical results (mean and standard deviation) of the HVR values on the proposed test problems. The best result regarding the mean for each problem instance is highlighted in boldface.}
\label{tab2}
\center
\begin{center}
\begin{tabular}{l|llllll}
\hline
\# of Objs & Method & DPF1 & DPF2 & DPF3 & DPF4 & DPF5\\
\hline
\multirow{10}{*}{m = 3, d = 2}
&NSGA-II & 9.85E-01 (2.92E-03)  & 9.92E-01 (5.74E-04)  & 9.90E-01 (7.29E-04)  & \textbf{9.96E-01} (7.71E-04)  & 9.21E-01 (6.15E-03) \\
&SPEA2+SDE & \textbf{9.88E-01} (9.76E-04)  & 9.92E-01 (4.85E-04)  & 9.90E-01 (1.03E-03)  & 9.90E-01 (4.81E-03)  & 9.55E-01 (4.16E-03) \\
&NSGA-III & 9.70E-01 (3.97E-03)  & 9.82E-01 (3.46E-03)  & 9.84E-01 (2.03E-03)  & 9.90E-01 (2.87E-03)  & 9.61E-01 (1.03E-03) \\
&MOEA/D & 9.76E-01 (1.29E-03)  & 6.35E-01 (6.31E-02)  & 9.53E-01 (1.08E-05)  & 8.70E-01 (5.35E-03)  & 9.61E-01 (1.24E-05) \\
&IDBEA & 9.54E-01 (3.04E-02)  & 9.69E-01 (1.17E-02)  & 9.83E-01 (9.82E-04)  & 1.18E-01 (2.77E-01)  & 9.61E-01 (4.19E-05) \\
&RVEA & 8.10E-01 (5.86E-02)  & 6.98E-01 (7.97E-02)  & 9.49E-01 (2.36E-04)  & 8.34E-01 (1.56E-01)  & 9.57E-01 (2.91E-05) \\
&IBEA & 5.70E-01 (5.71E-02)  & 9.56E-01 (1.41E-01)  & \textbf{9.92E-01} (3.80E-04)  & 1.89E-01 (2.87E-02)  & \textbf{9.67E-01} (2.97E-03) \\
&$\delta$-MOSS & 7.45E-01 (4.15E-01)  & 4.80E-01 (5.04E-01)  & 9.51E-01 (1.73E-01)  & 4.49E-01 (4.41E-01)  & 9.22E-01 (6.90E-03) \\
&OSP & 8.93E-01 (1.54E-01)  & 9.51E-01 (1.04E-01)  & 7.96E-01 (9.07E-02)  & 4.39E-01 (2.43E-01)  & 6.35E-01 (9.21E-02) \\
&NCIE & 9.85E-01 (2.34E-03)  & \textbf{9.93E-01} (3.36E-04)  & 9.75E-01 (2.08E-04)  & 9.51E-01 (1.68E-01)  & 8.79E-01 (1.38E-01) \\
\hline
\multirow{10}{*}{m = 6, d = 3}
&NSGA-II & 9.05E-01 (1.15E-02)  & 8.30E-01 (4.02E-02)  & 8.16E-01 (1.30E-02)  & \textbf{8.82E-01} (2.11E-02)  & 6.57E-01 (3.34E-02) \\
&SPEA2+SDE & \textbf{9.27E-01} (7.42E-03)  & \textbf{9.25E-01} (7.89E-03)  & 8.88E-01 (6.59E-03)  & 8.63E-01 (3.12E-02)  & 9.53E-01 (3.18E-03) \\
&NSGA-III & 8.47E-01 (2.45E-02)  & 8.30E-01 (3.41E-02)  & 7.80E-01 (2.15E-02)  & 8.24E-01 (6.34E-02)  & 9.51E-01 (1.93E-03) \\
&MOEA/D & 8.52E-01 (4.10E-03)  & 2.03E-01 (7.27E-02)  & 3.68E-01 (1.30E-01)  & 6.79E-01 (3.57E-03)  & 9.54E-01 (1.20E-03) \\
&IDBEA & 8.50E-01 (7.28E-02)  & 6.90E-01 (1.13E-01)  & 8.04E-01 (4.71E-03)  & 1.38E-06 (7.67E-06)  & 9.53E-01 (1.69E-03) \\
&RVEA & 6.05E-01 (6.47E-02)  & 2.33E-01 (6.20E-02)  & 5.04E-01 (1.16E-01)  & 4.93E-01 (1.26E-01)  & 9.50E-01 (1.07E-03) \\
&IBEA & 3.06E-01 (6.63E-02)  & 9.12E-01 (9.74E-03)  & \textbf{9.04E-01} (5.80E-03)  & 3.09E-02 (5.65E-03)  & \textbf{9.68E-01} (2.10E-03) \\
&$\delta$-MOSS & 7.82E-01 (2.06E-01)  & 7.64E-01 (1.38E-01)  & 8.47E-01 (1.31E-02)  & 7.29E-01 (3.45E-01)  & 6.53E-01 (2.48E-02) \\
&OSP & 8.74E-01 (1.18E-02)  & 7.05E-01 (1.18E-01)  & 4.37E-01 (3.13E-01)  & 5.26E-01 (3.24E-01)  & 4.60E-01 (3.80E-02) \\
&NCIE & 8.94E-01 (1.33E-02)  & 5.70E-01 (3.22E-01)  & 3.81E-01 (3.44E-01)  & 1.86E-01 (3.08E-01)  & 4.07E-01 (1.72E-01) \\
\hline
\multirow{11}{*}{m = 10, d = 5}
&NSGA-II & 7.13E-01 (2.67E-02)  & 3.32E-01 (2.64E-02)  & 5.37E-01 (6.00E-02)  & 8.83E-01 (8.54E-02)  & 3.74E-04 (2.08E-03) \\
&SPEA2+SDE & \textbf{8.62E-01} (9.97E-03)  & \textbf{8.69E-01} (1.56E-02)  & 7.39E-01 (5.16E-02)  & 2.58E-01 (2.86E-02)  & 9.85E-01 (1.08E-03) \\
&NSGA-III & 6.07E-01 (9.38E-02)  & 4.67E-01 (5.96E-02)  & 8.00E-01 (6.20E-02)  & \textbf{9.85E-01} (9.04E-02)  & 9.86E-01 (6.87E-04) \\
&MOEA/D & 6.22E-01 (1.14E-02)  & 2.88E-02 (4.13E-02)  & 3.72E-02 (3.21E-02)  & 2.03E-02 (5.44E-03)  & 9.87E-01 (5.38E-04) \\
&IDBEA & 8.20E-01 (5.39E-03)  & 2.17E-01 (1.48E-01)  & 5.81E-02 (5.81E-02)  & 4.11E-06 (2.29E-05)  & 9.53E-01 (1.47E-01) \\
&RVEA & 3.58E-01 (4.85E-02)  & 1.06E-02 (1.27E-02)  & 1.97E-02 (2.82E-02)  & 3.52E-02 (4.00E-02)  & 9.87E-01 (5.58E-04) \\
&IBEA & 2.50E-01 (5.28E-02)  & 8.43E-01 (2.39E-01)  & \textbf{9.77E-01} (3.01E-02)  & 3.19E-03 (3.69E-04)  & \textbf{9.90E-01} (9.12E-04) \\
&$\delta$-MOSS & 6.67E-01 (5.92E-02)  & 3.20E-01 (2.92E-02)  & 7.24E-01 (5.22E-02)  & 8.49E-01 (4.35E-01)  & 0.00E+00 (0.00E+00) \\
&OSP & 8.13E-01 (1.43E-02)  & 5.69E-01 (4.75E-02)  & 3.92E-01 (9.63E-02)  & 6.42E-01 (2.23E-01)  & 7.01E-01 (4.00E-02) \\
&NCIE & 2.28E-01 (2.73E-01)  & 1.73E-01 (9.83E-02)  & 2.49E-01 (1.95E-01)  & 1.31E-01 (2.57E-01)  & 3.65E-01 (1.73E-01) \\
\hline
\hline
\end{tabular}
\end{center}
\end{table*}

\subsection{Impact of the Proposed Three Characteristics}
In the previous subsection, we have presented the performance of the current state-of-the-art algorithms on the proposed test problems and have found that they provide big challenges to the algorithms. However, we may not be able to conclude that this under-performance of the considered algorithms is caused by the proposed three characteristics since the tested problems have different features with respect to their essential objectives (\textit{e.g.}, multi-modal, bias and disconnected). To investigate the impact of the proposed three characteristics, in this section we modify the original DPFs by making them have the same essential objectives as DTLZ5($I$,$M$), called DPF1A--DPF5A (see Appendix \ref{DefinitionT3}). Therefore, the performance     difference of algorithms between these problems can fully boil down to the proposed characteristics.

In this experiment, we compare the performance of five algorithms, which include three objective reduction-based MOEAs ($\delta$-MOSS, OSP and NCIE), an improved decomposition-based algorithm (IDBEA) and an indicator-based algorithm (IBEA) on DPF1A-DPF5A and DTLZ5($I$, $M$)~\cite{Deb2005on}. We test the instances with $I = 3, M=10$ for DTLZ5($I$, $M$), and $d=3, m = 10$ for the proposed problems. The number of decision variables of DTLZ5($I$, $M$) is set as $19$ by following the recommendation in \cite{Deb2005on}, and the same setting is adopted for DPF1A--DPF5A. We run the tested MOEAs for $31$ times with the population size of $100$ and the maximum number of executing generations $1000$.

\begin{figure*}[bpt]
\centering
\subfigure[$\delta$-MOSS]{
\centering
\includegraphics[width=0.18\textwidth]{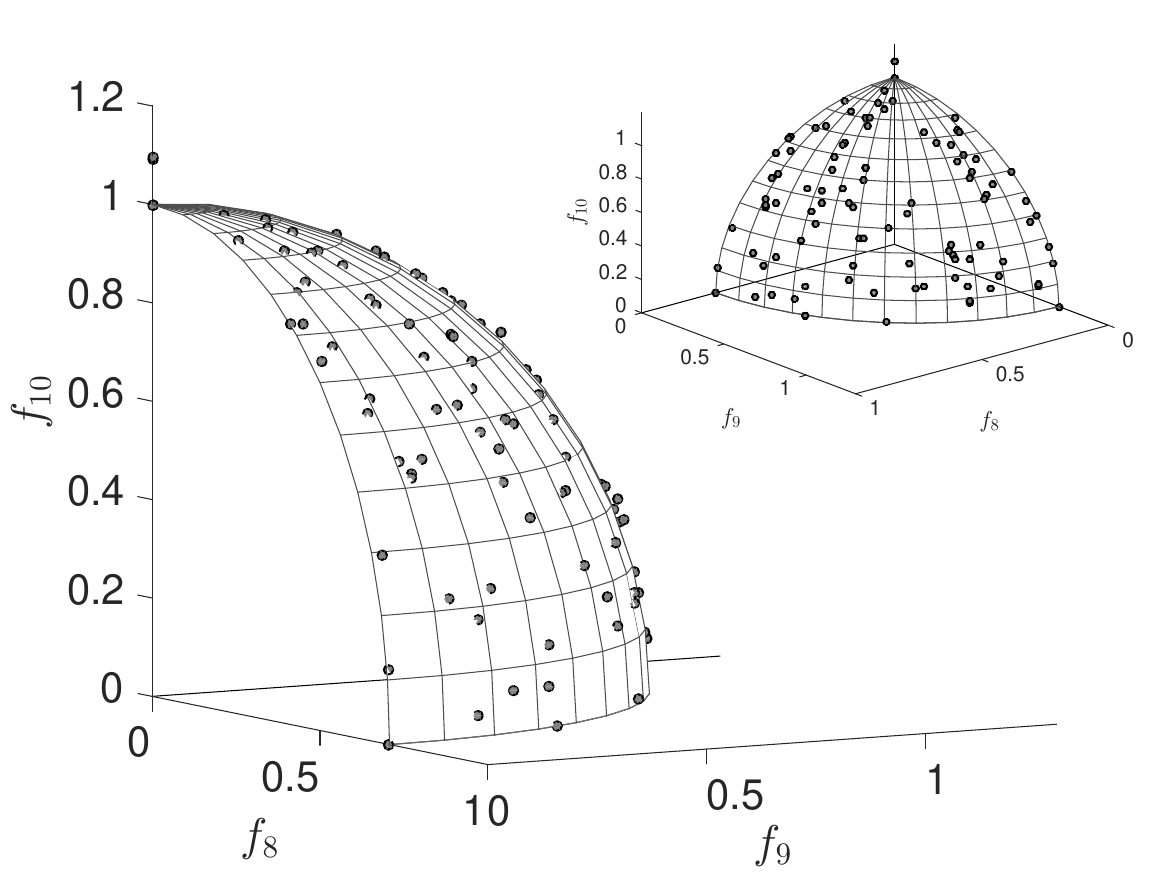}}
\subfigure[NCIE]{
\centering
\includegraphics[width=0.18\textwidth]{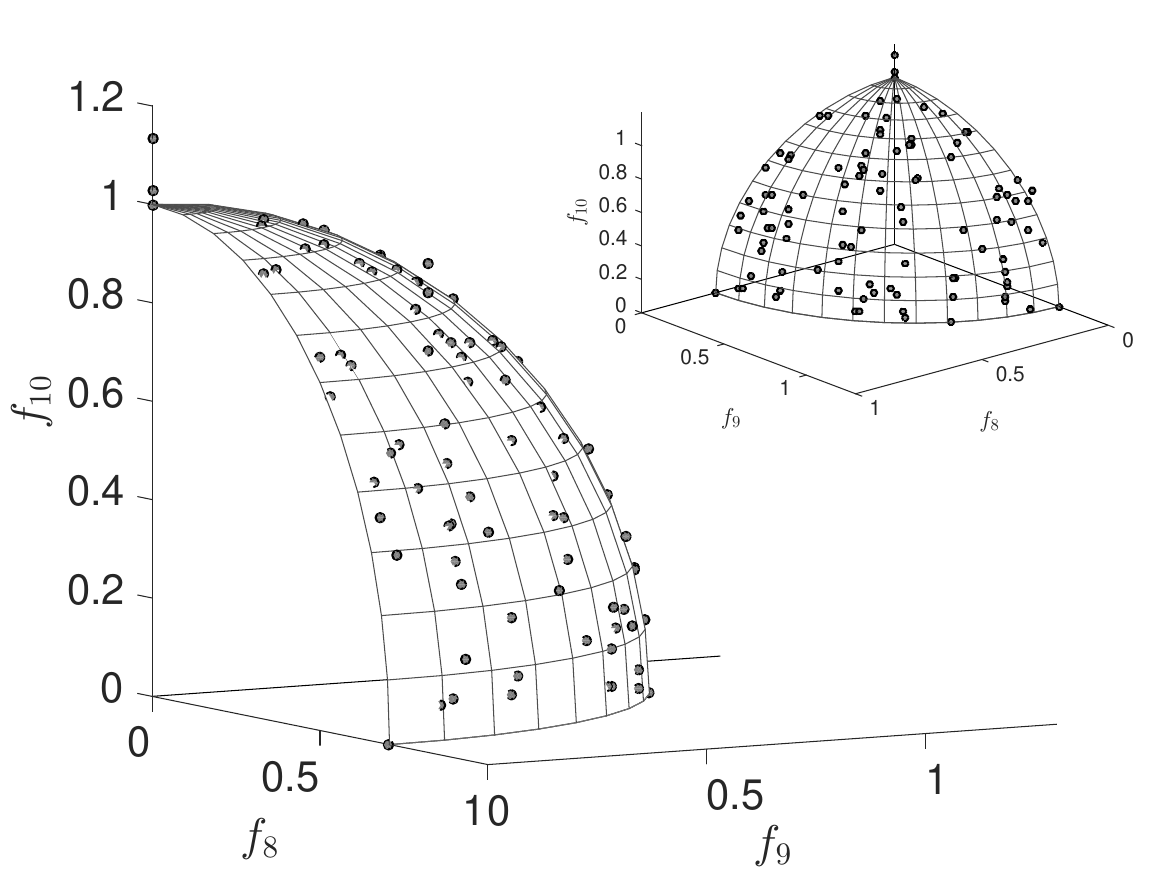}}
\subfigure[OSP]{
\centering
\includegraphics[width=0.18\textwidth]{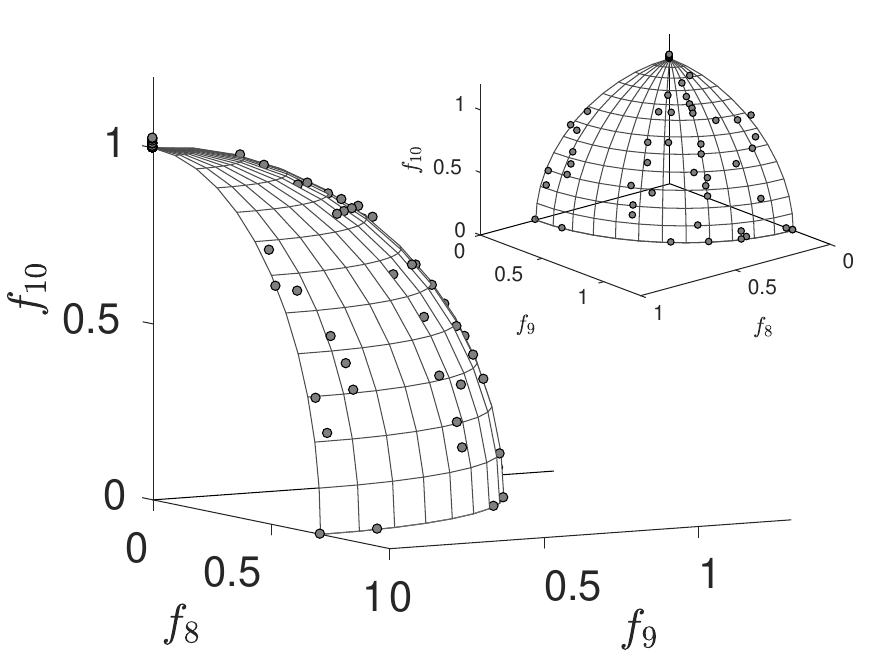}}
\subfigure[IBEA]{
\centering
\includegraphics[width=0.18\textwidth]{DTLZ5IMCIBEA.pdf}}
\subfigure[IDBEA]{
\centering
\includegraphics[width=0.18\textwidth]{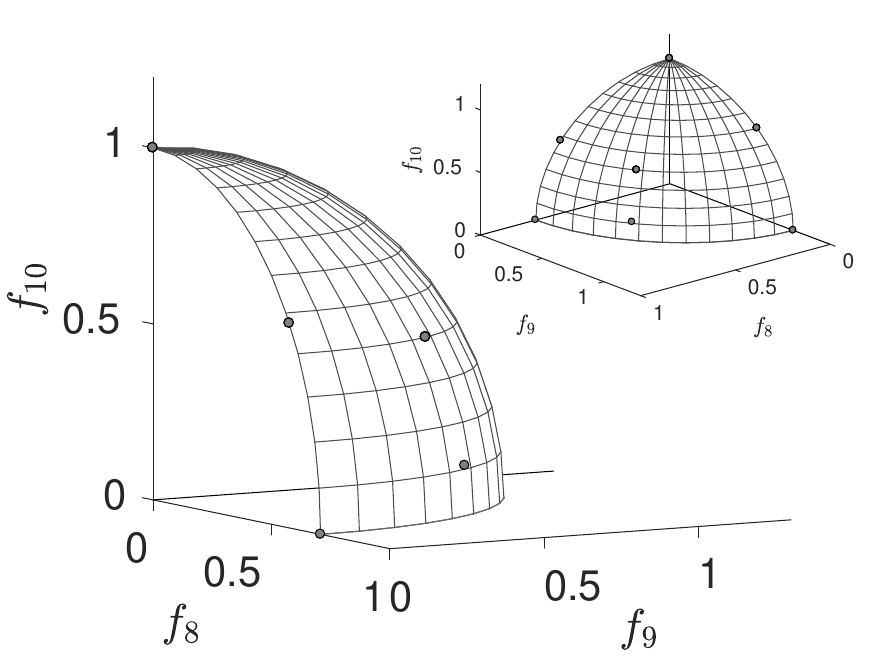}}
\caption{The solution set of the five objective reduction-based algorithms on DTLZ5($3$, $10$) in the run associated with its best HV value, where the grid mesh denotes the PF of the problem.}
\label{fig:comparisonV}
\end{figure*}

\begin{figure*}[hbpt]
\centering
\subfigure[$\delta$-MOSS on DPF1A]{
\centering
\includegraphics[width=0.18\textwidth]{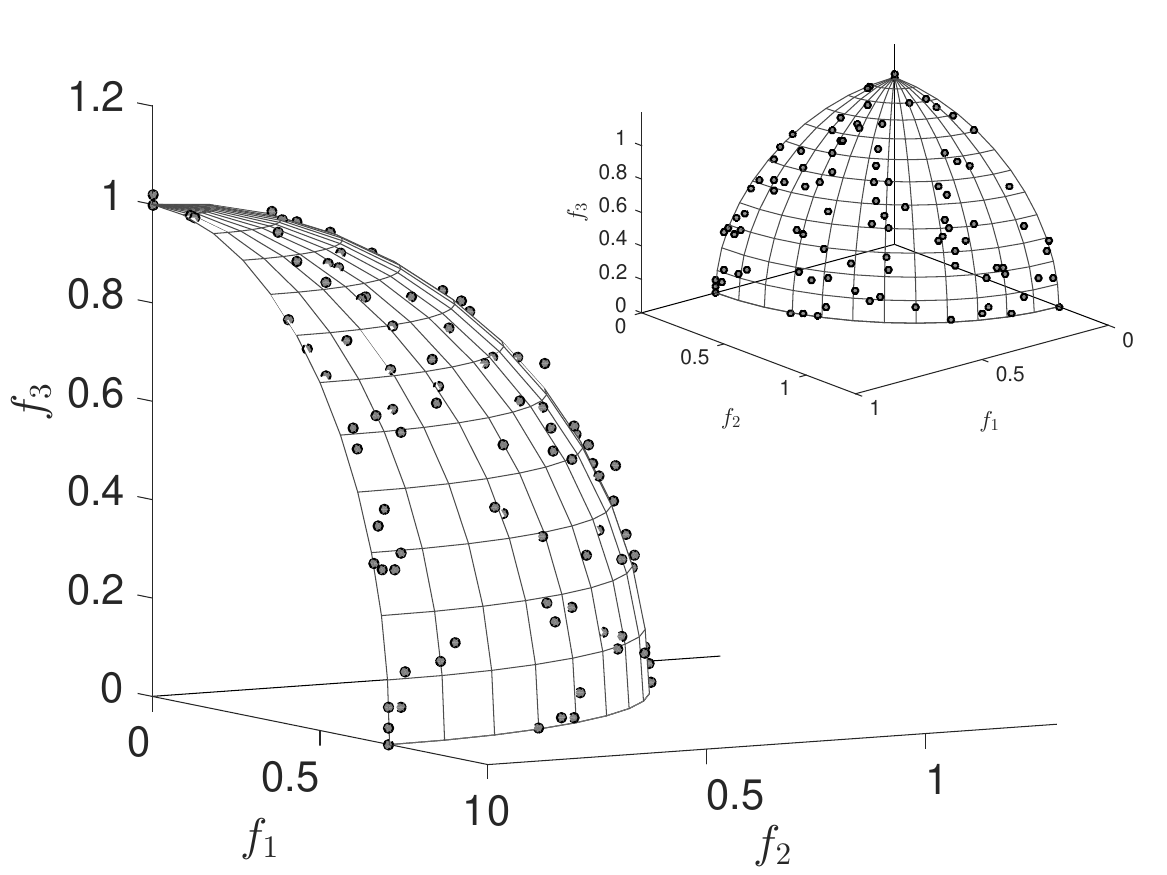}}
\subfigure[NCIE on DPF1A]{
\centering
\includegraphics[width=0.18\textwidth]{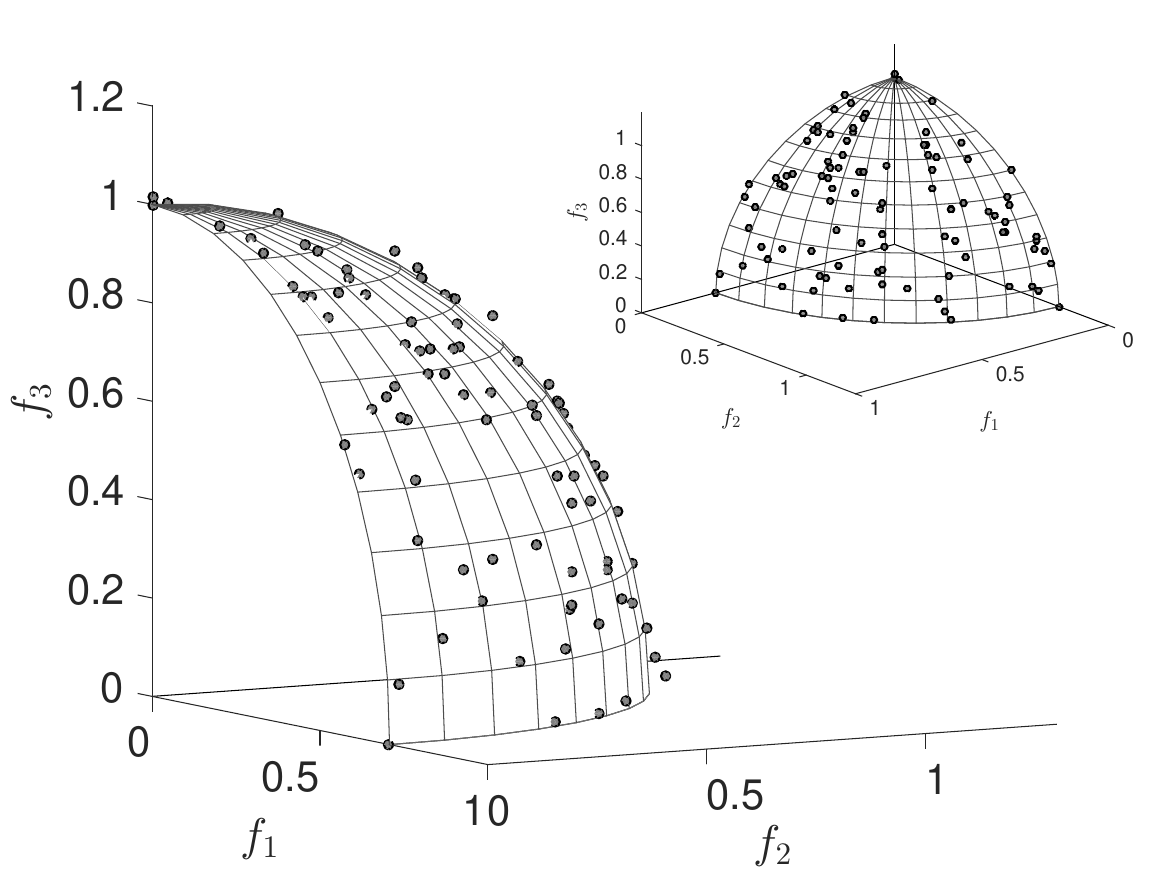}}
\subfigure[OSP on DPF1A]{
\centering
\includegraphics[width=0.18\textwidth]{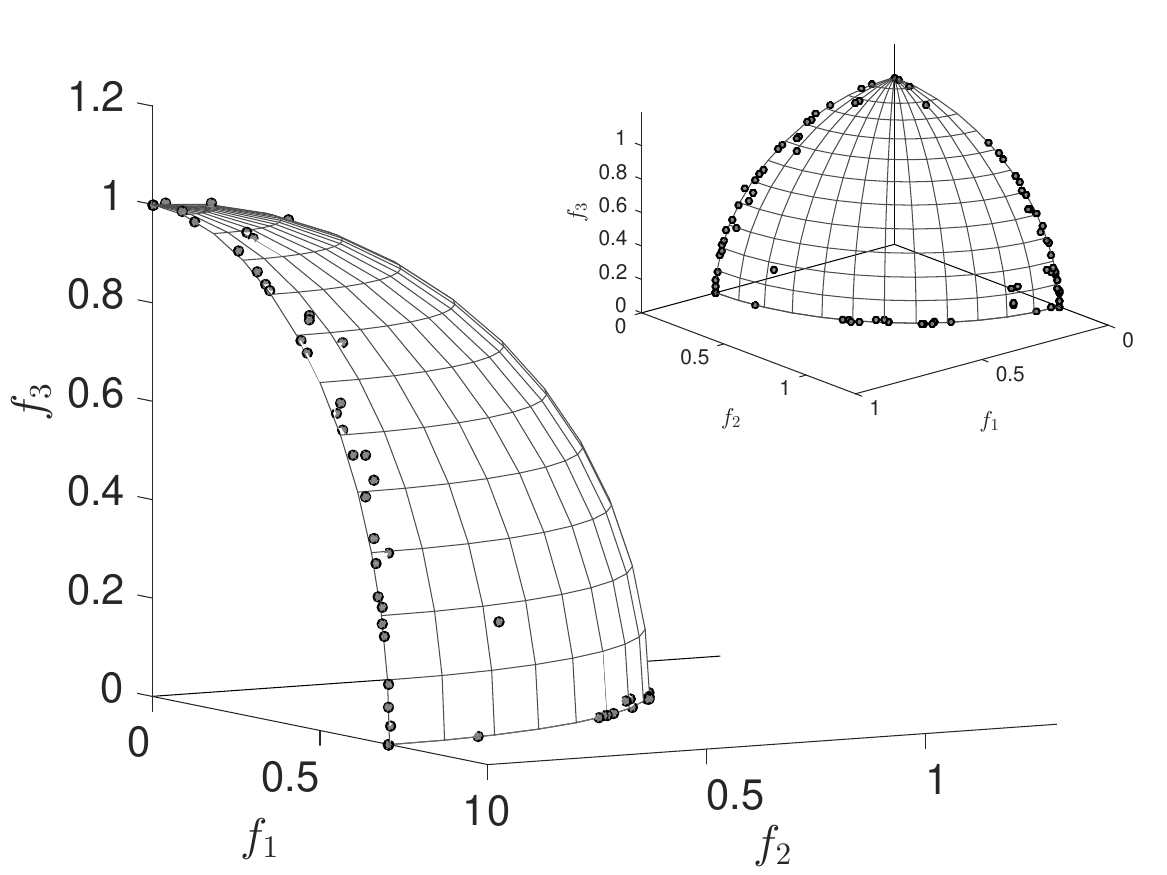}}
\subfigure[IBEA on DPF1A]{
\centering
\includegraphics[width=0.18\textwidth]{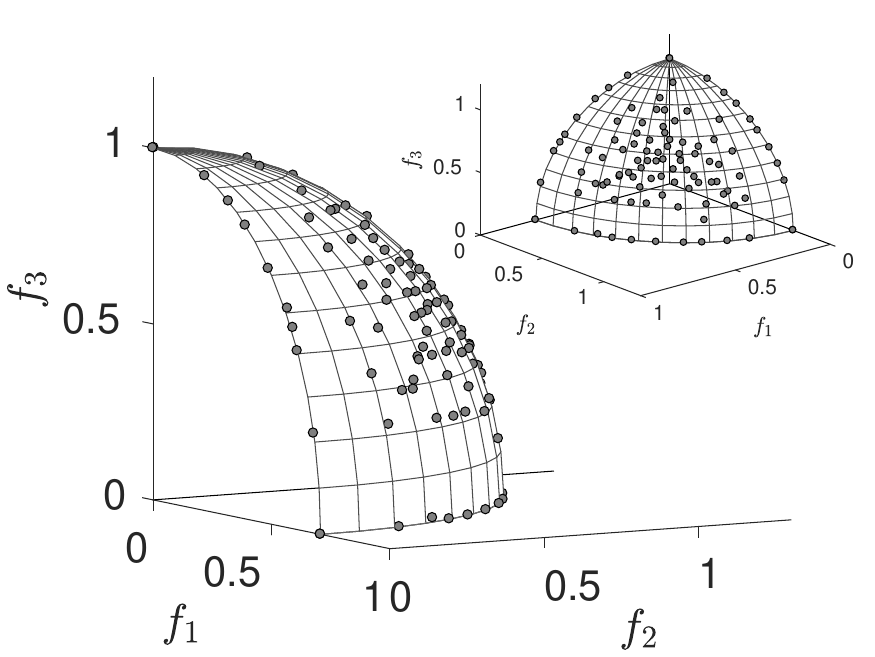}}
\subfigure[IDBEA on DPF1A]{
\centering
\includegraphics[width=0.18\textwidth]{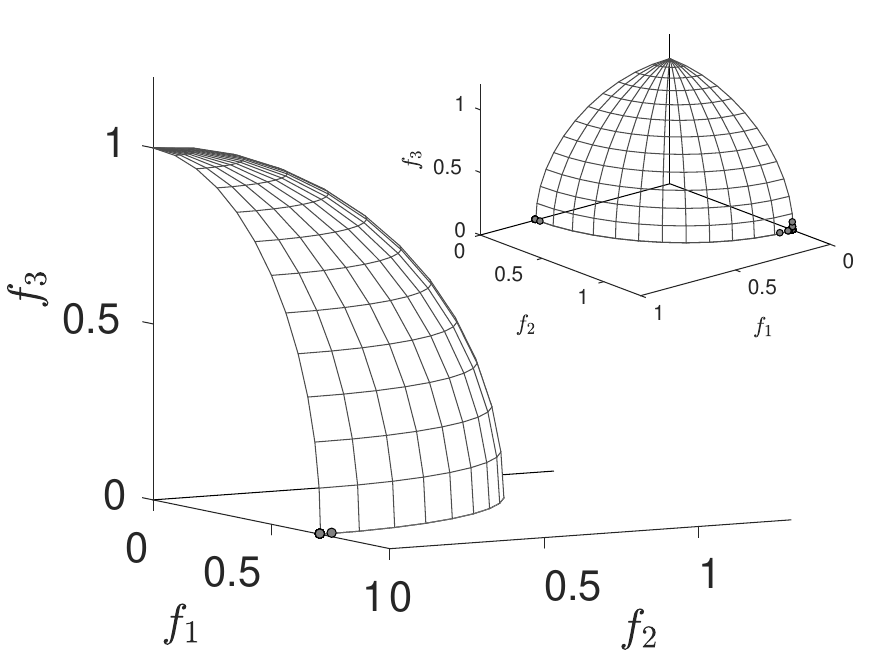}}
\subfigure[$\delta$-MOSS on DPF2A]{
\centering
\includegraphics[width=0.18\textwidth]{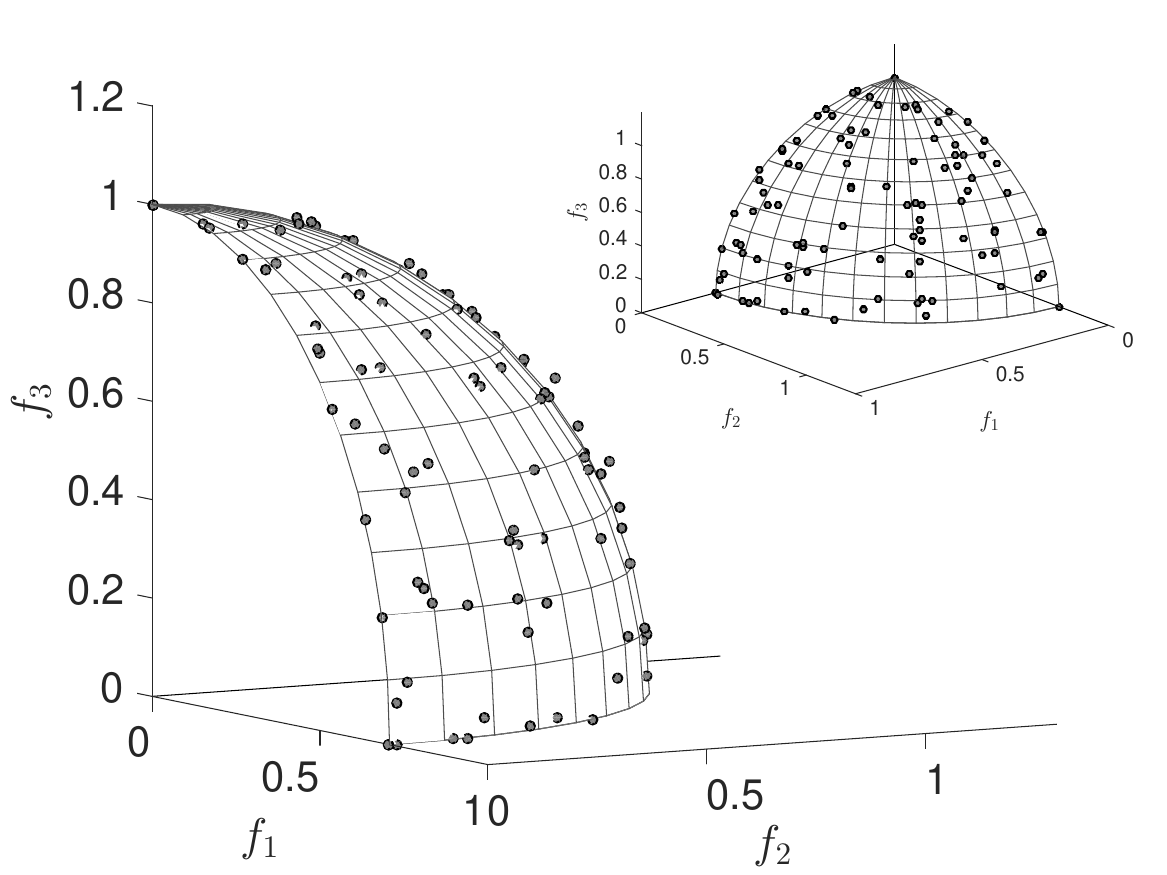}}
\subfigure[NCIE on DPF2A]{
\centering
\includegraphics[width=0.18\textwidth]{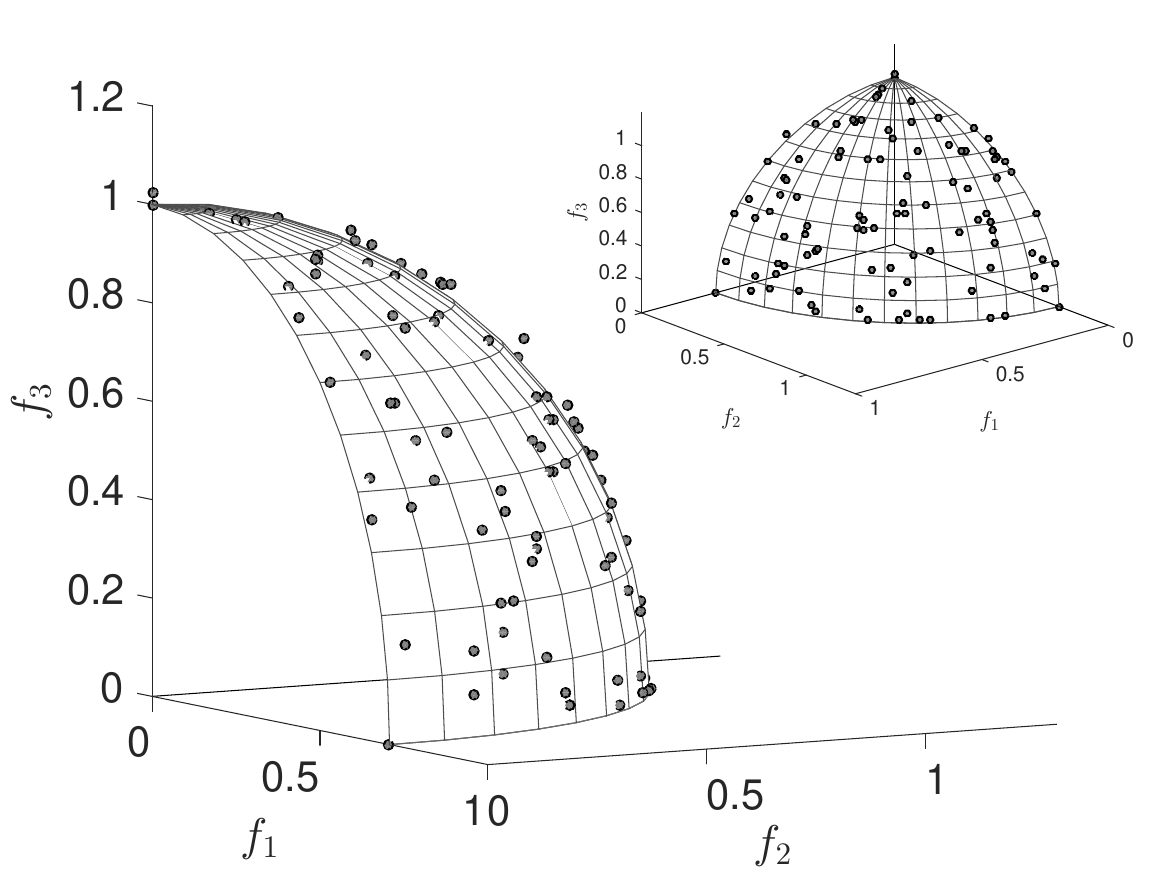}}
\subfigure[OSP on DPF2A]{
\centering
\includegraphics[width=0.18\textwidth]{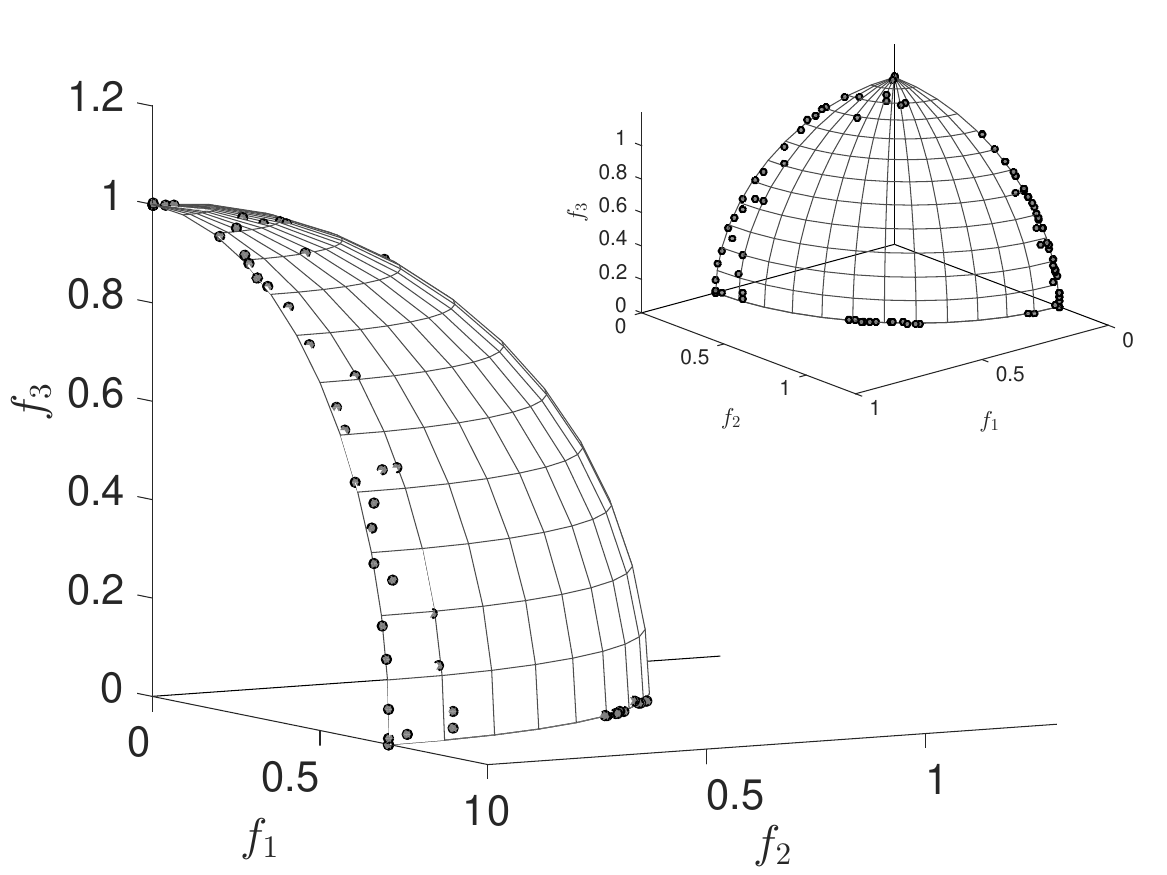}}
\subfigure[IBEA on DPF2A]{
\centering
\includegraphics[width=0.18\textwidth]{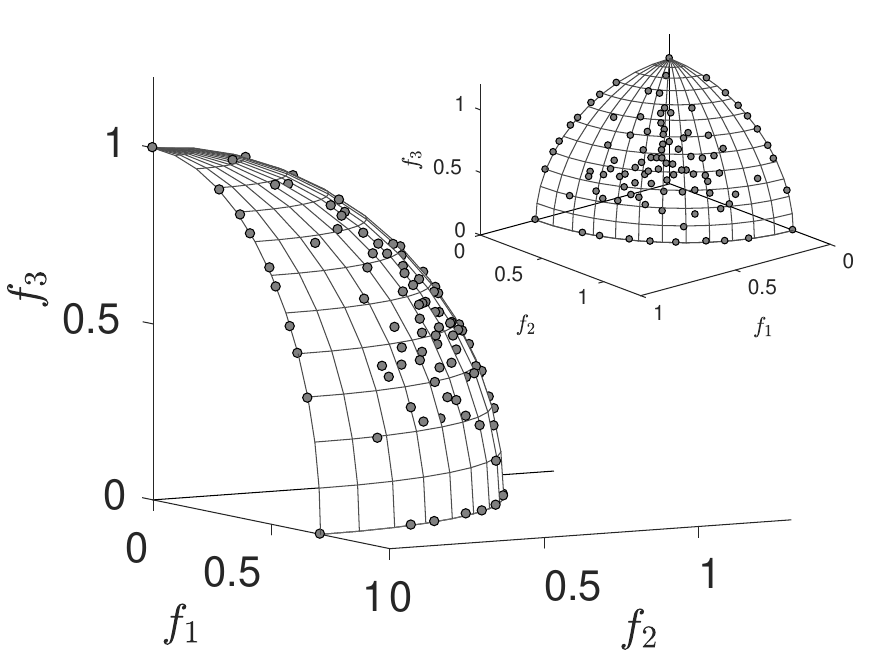}}
\subfigure[IDBEA on DPF2A]{
\centering
\includegraphics[width=0.18\textwidth]{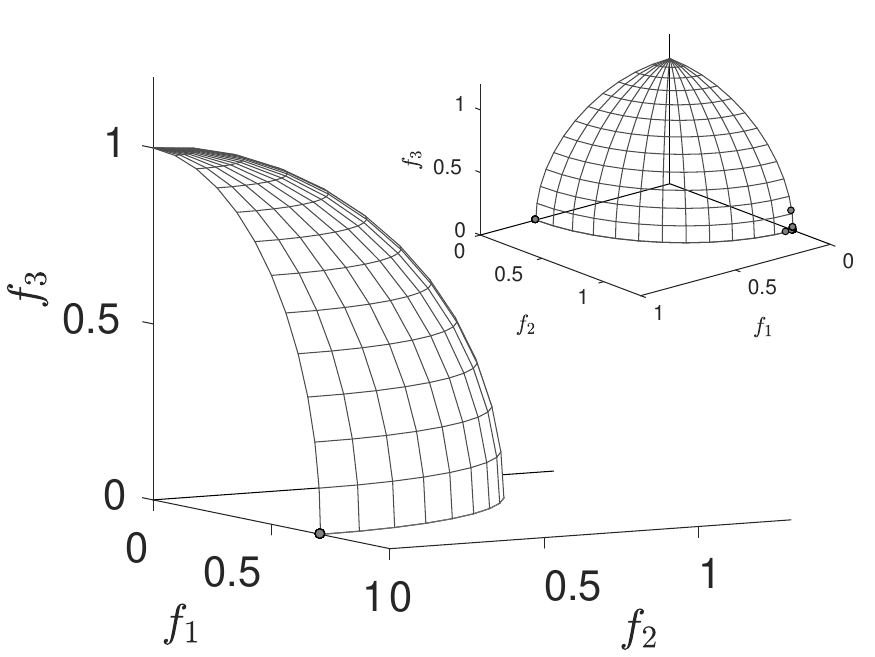}}
\subfigure[$\delta$-MOSS on DPF3A]{
\centering
\includegraphics[width=0.18\textwidth]{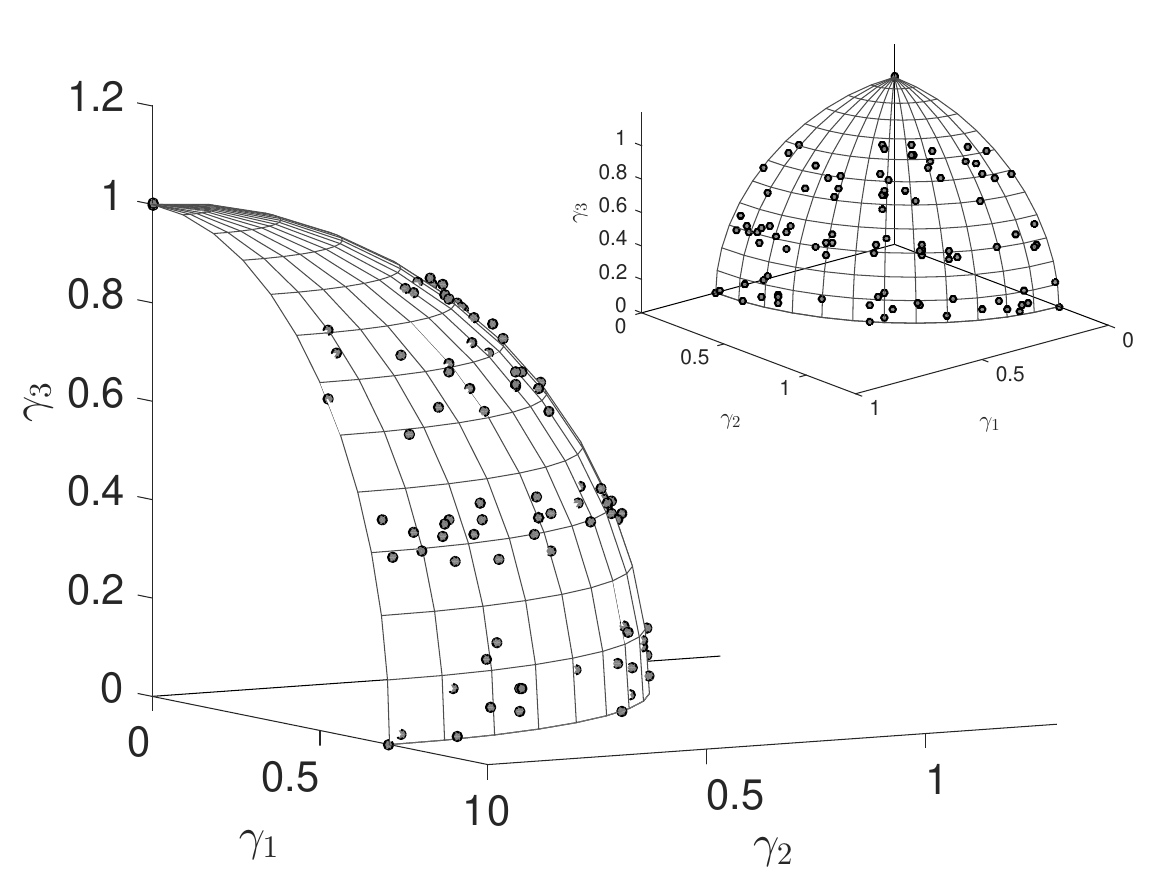}}
\subfigure[NCIE on DPF3A]{
\centering
\includegraphics[width=0.18\textwidth]{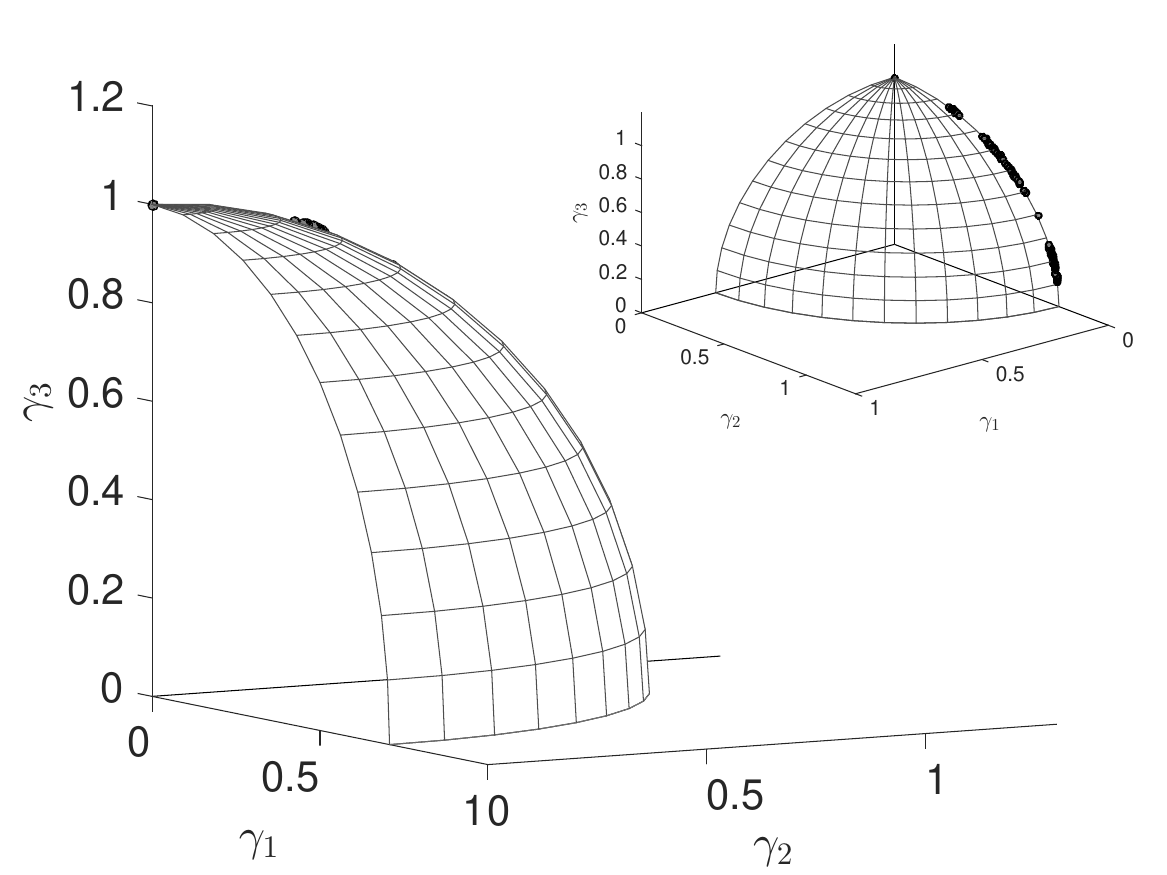}}
\subfigure[OSP on DPF3A]{
\centering
\includegraphics[width=0.18\textwidth]{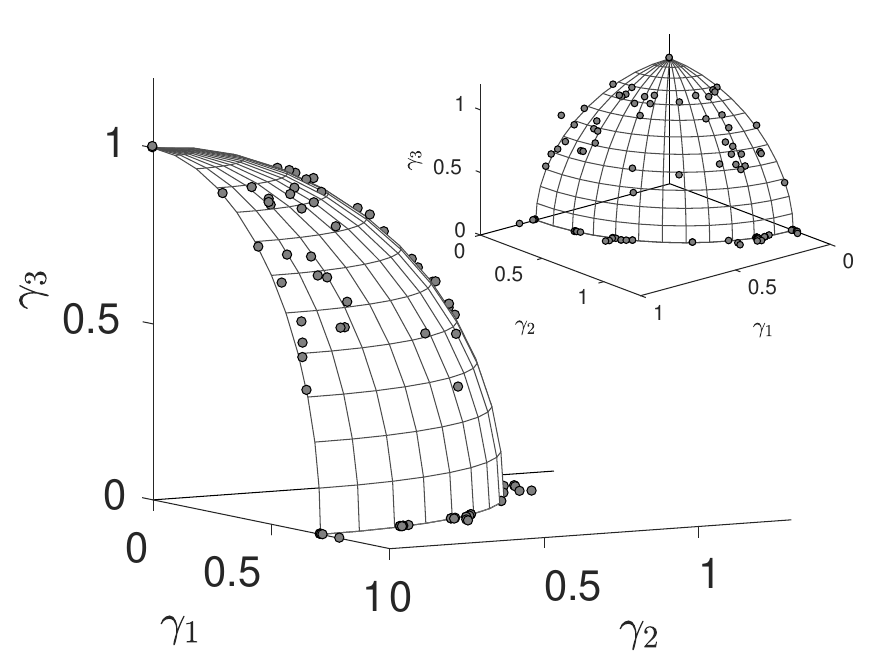}}
\subfigure[IBEA on DPF3A]{
\centering
\includegraphics[width=0.18\textwidth]{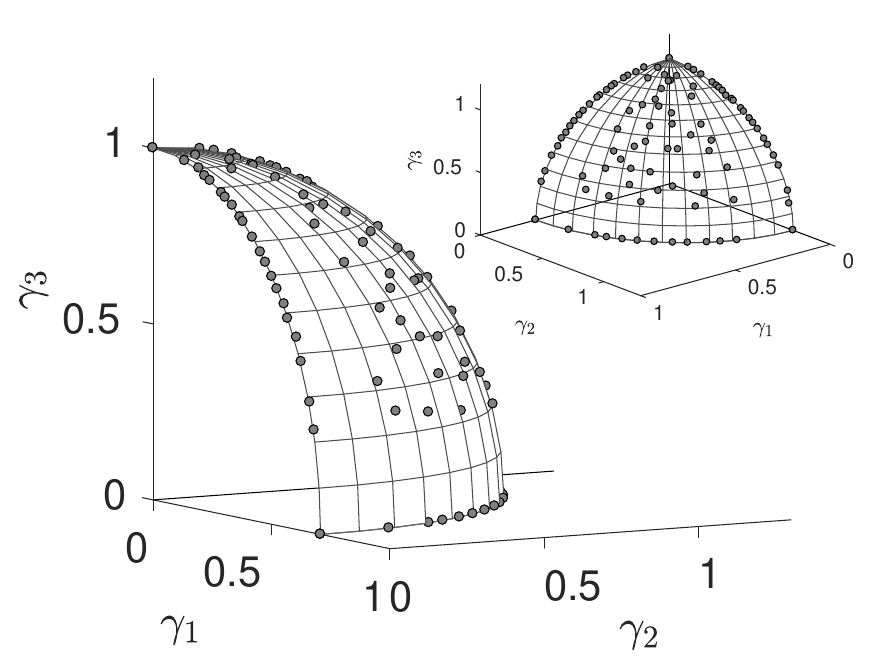}}
\subfigure[IDBEA on DPF3A]{
\centering
\includegraphics[width=0.18\textwidth]{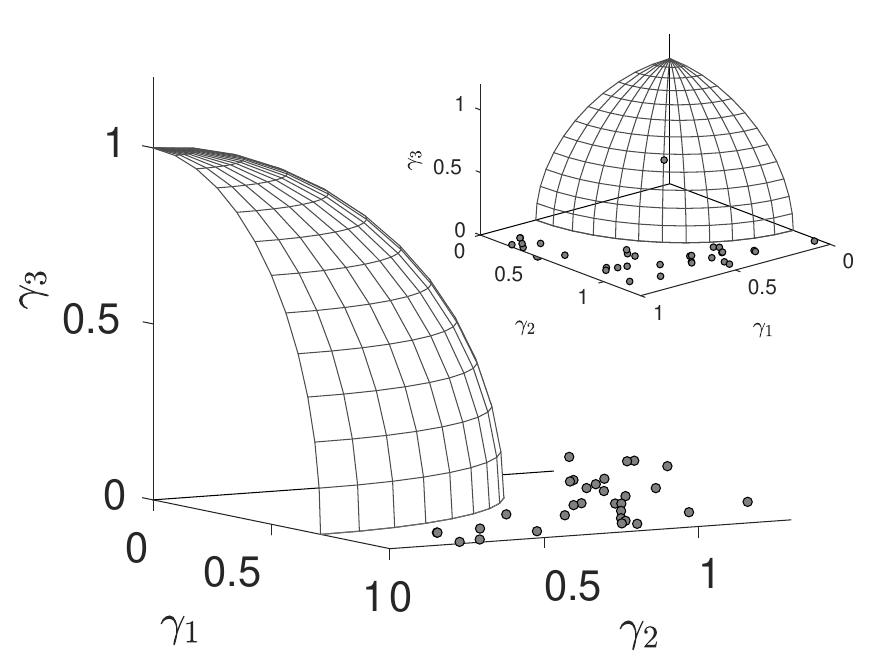}}
\subfigure[$\delta$-MOSS on DPF4A]{
\centering
\includegraphics[width=0.18\textwidth]{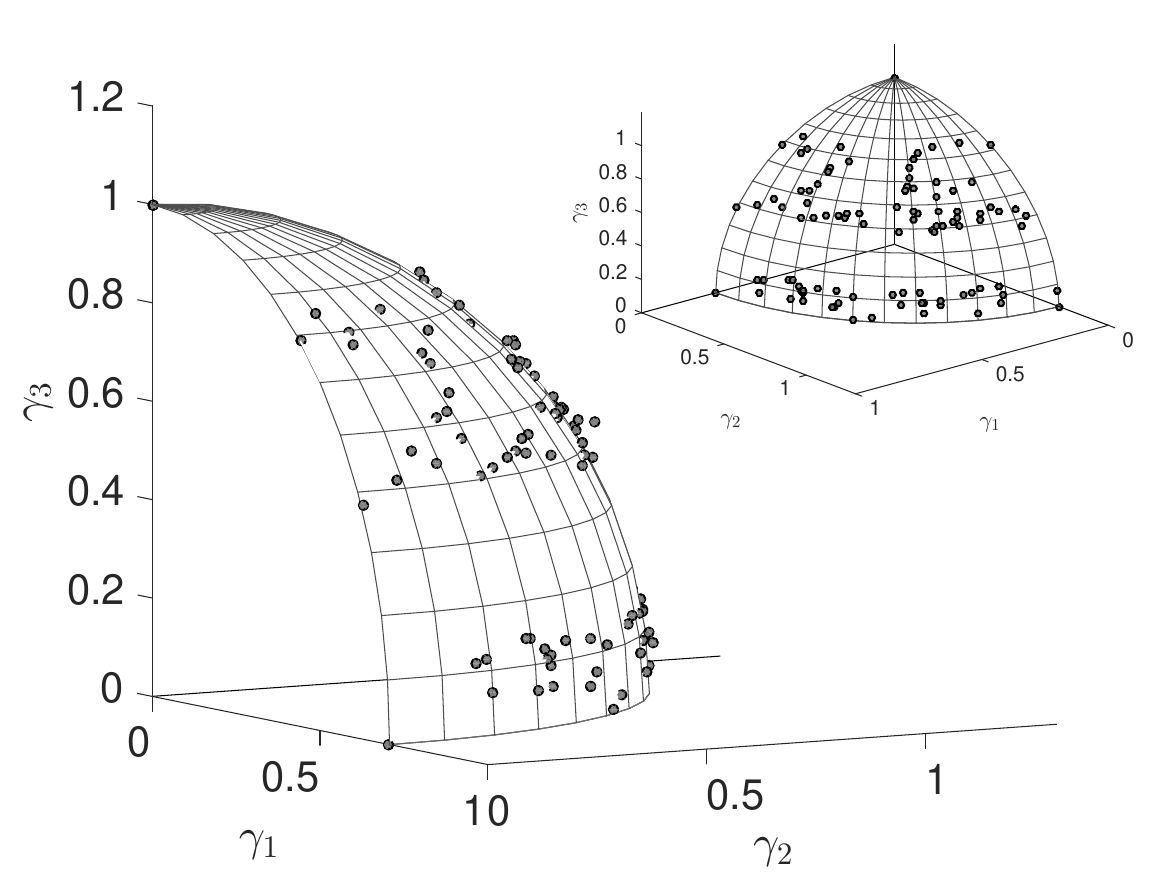}}
\subfigure[NCIE on DPF4A]{
\centering
\includegraphics[width=0.18\textwidth]{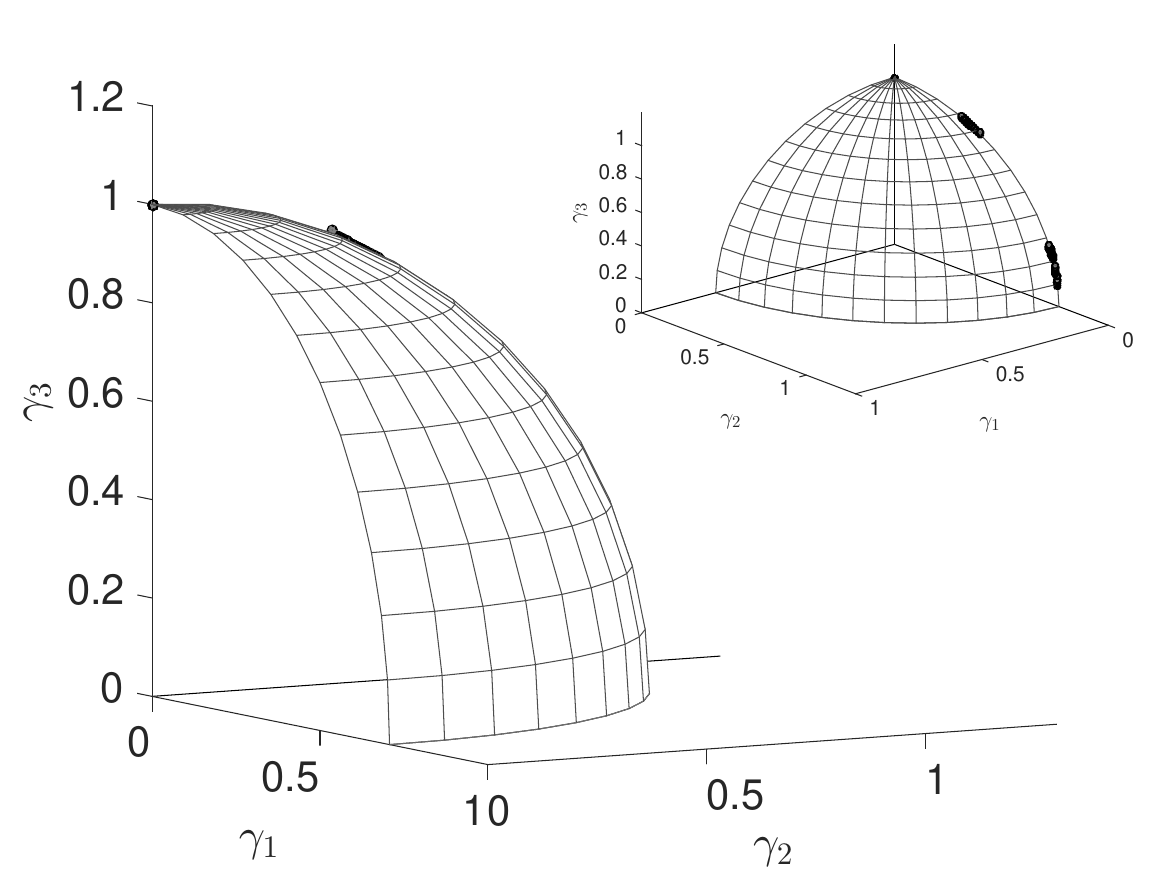}}
\subfigure[OSP on DPF4A]{
\centering
\includegraphics[width=0.18\textwidth]{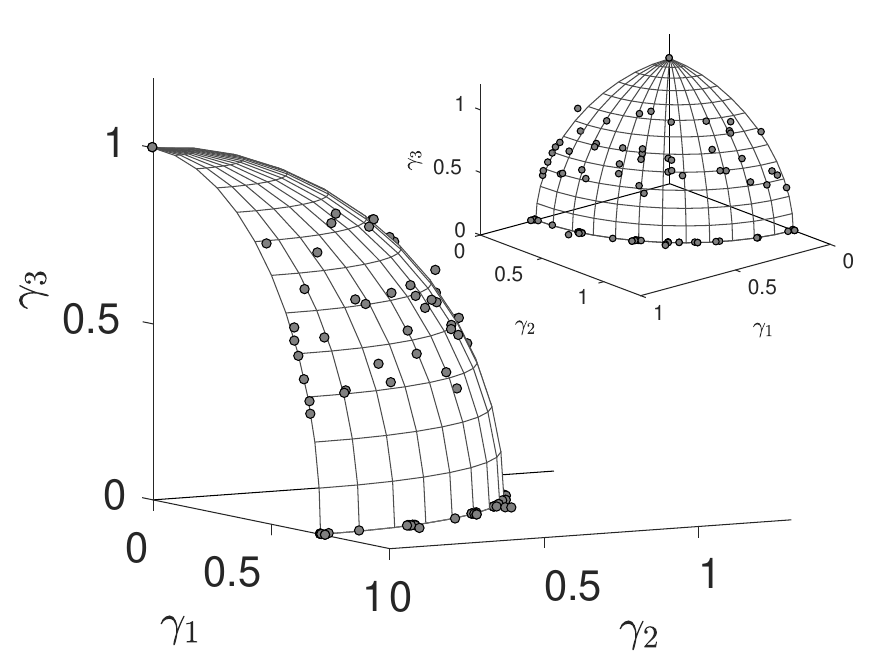}}
\subfigure[IBEA on DPF4A]{
\centering
\includegraphics[width=0.18\textwidth]{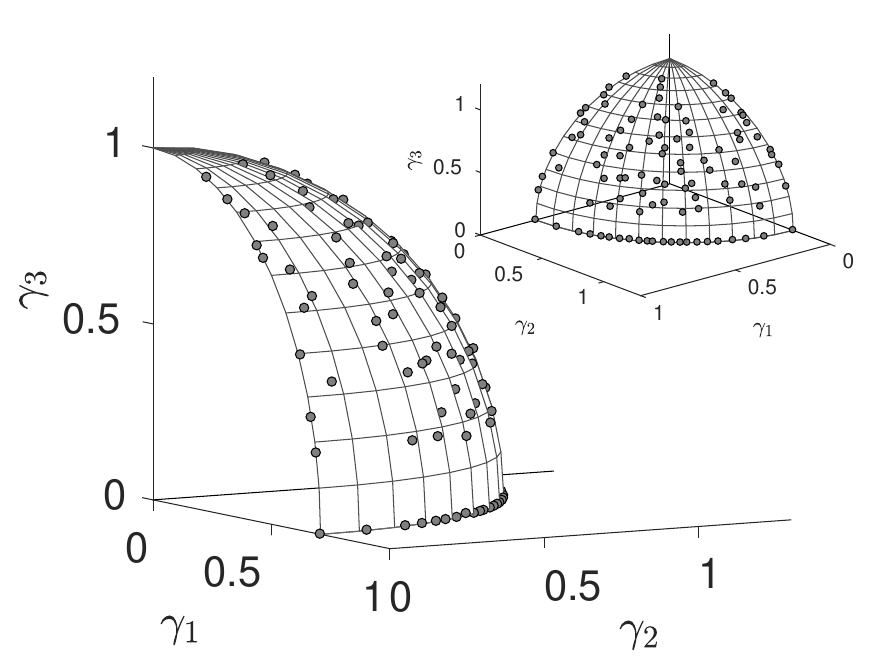}}
\subfigure[IDBEA on DPF4A]{
\centering
\includegraphics[width=0.18\textwidth]{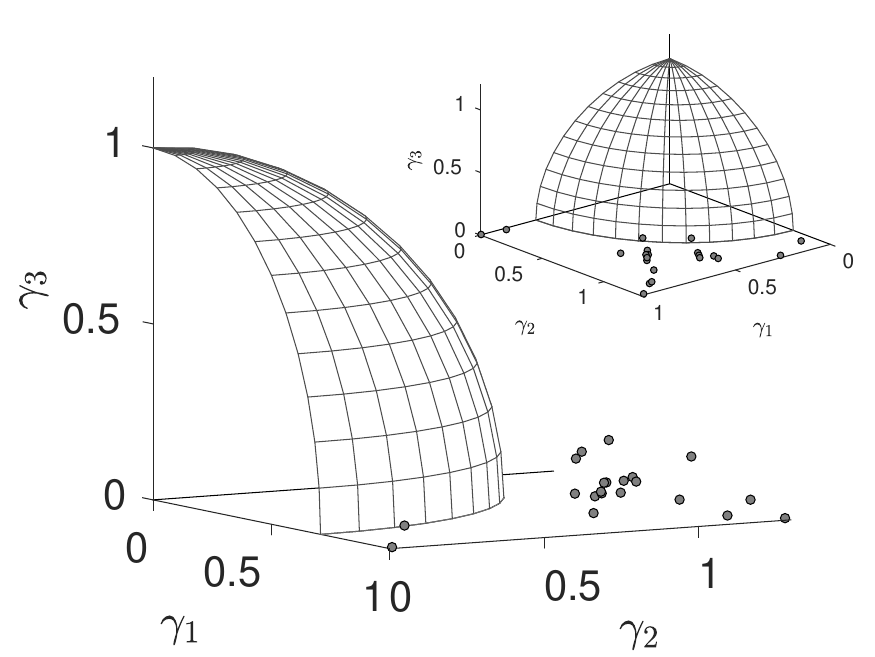}}
\subfigure[$\delta$-MOSS on DPF5A]{
\centering
\includegraphics[width=0.18\textwidth]{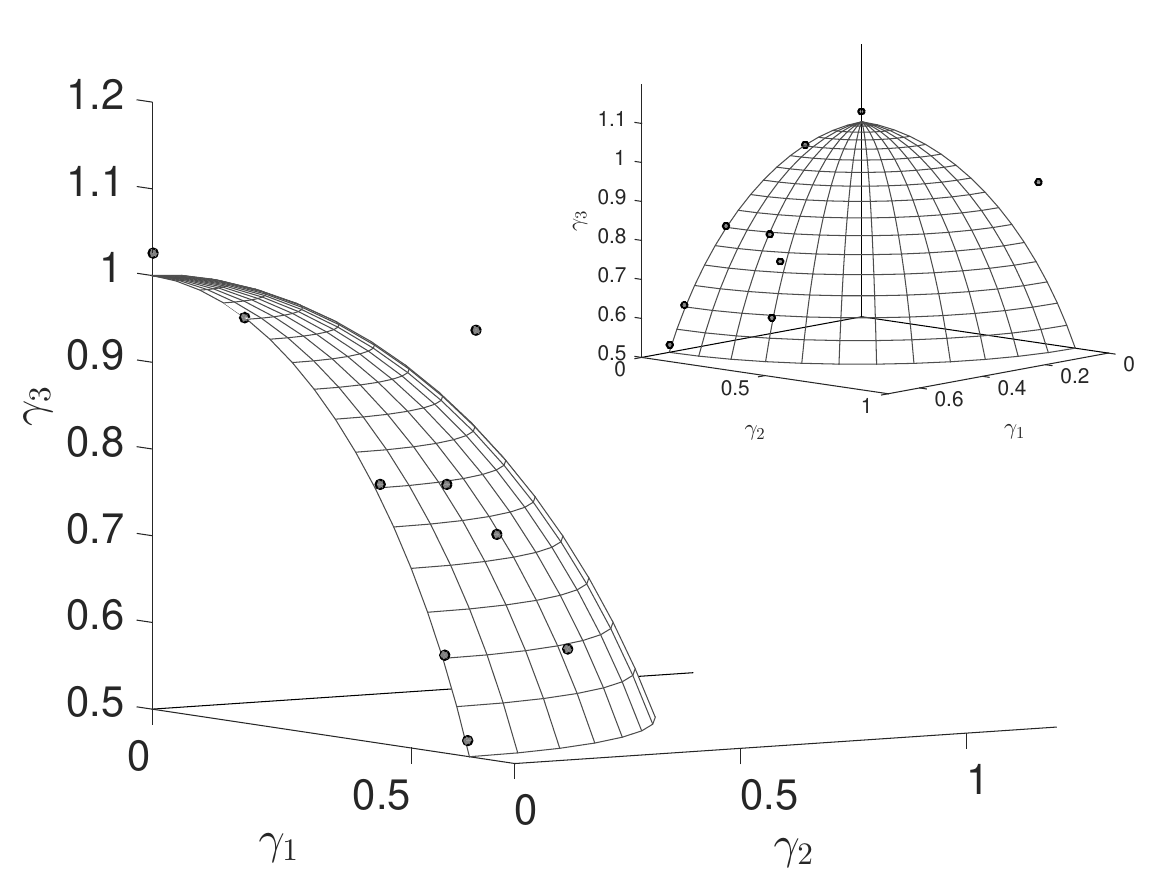}}
\subfigure[NCIE on DPF5A]{
\centering
\includegraphics[width=0.18\textwidth]{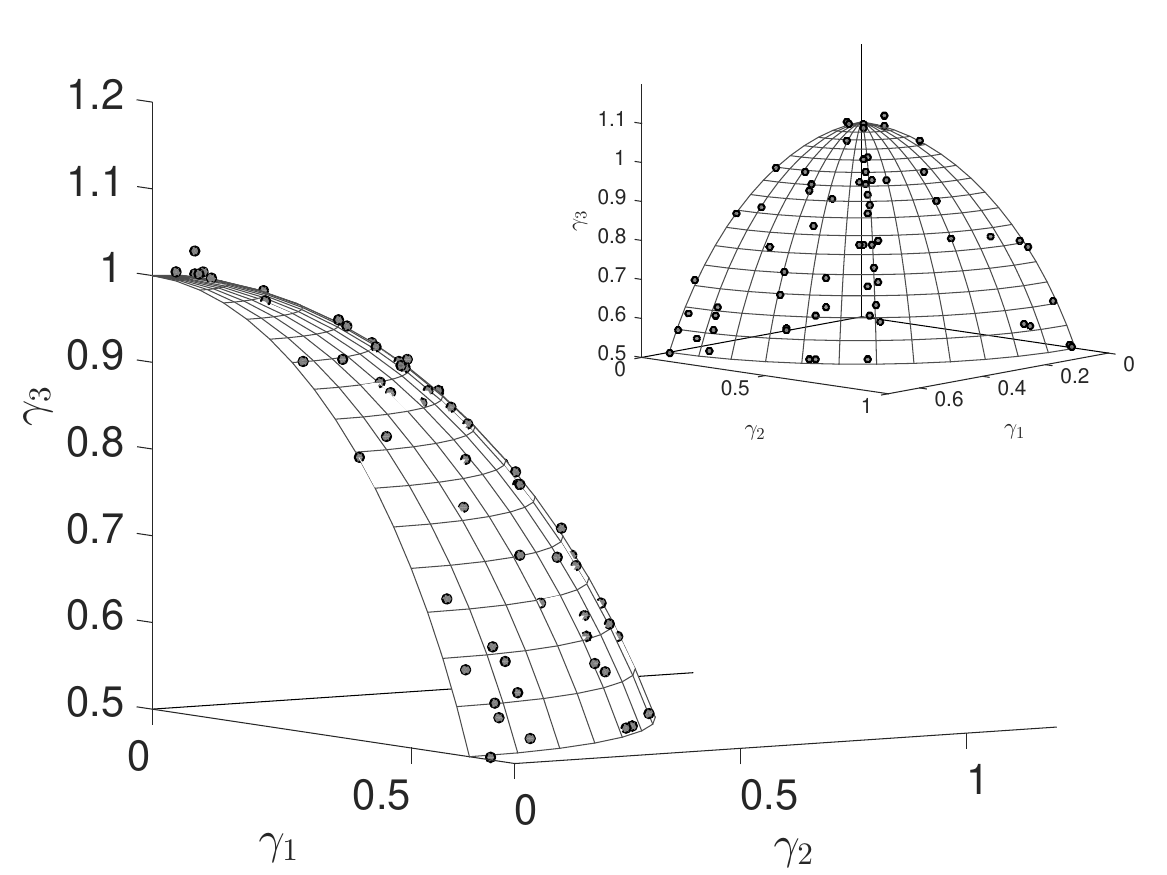}}
\subfigure[OSP on DPF5A]{
\centering
\includegraphics[width=0.18\textwidth]{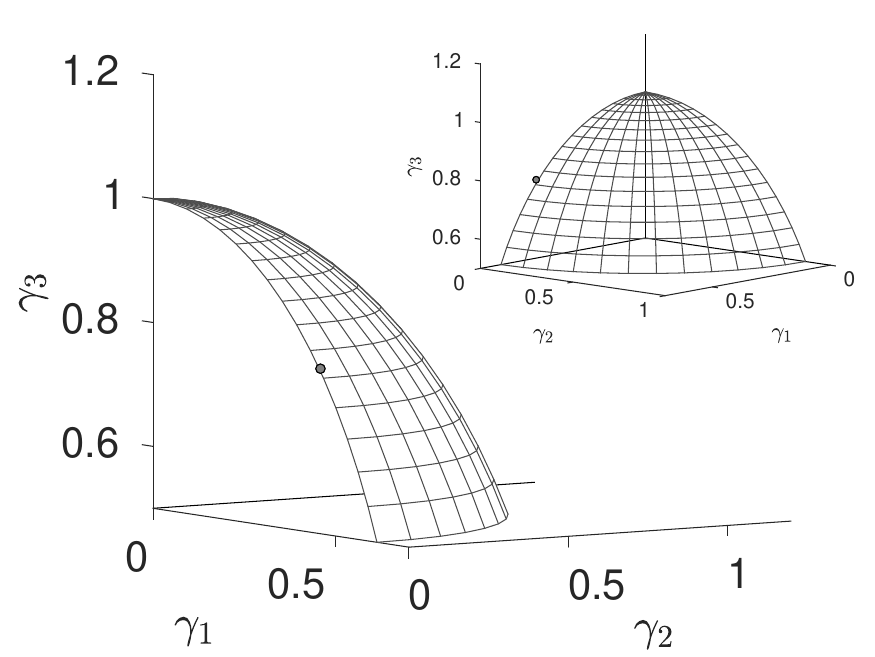}}
\subfigure[IBEA on DPF5A]{
\centering
\includegraphics[width=0.18\textwidth]{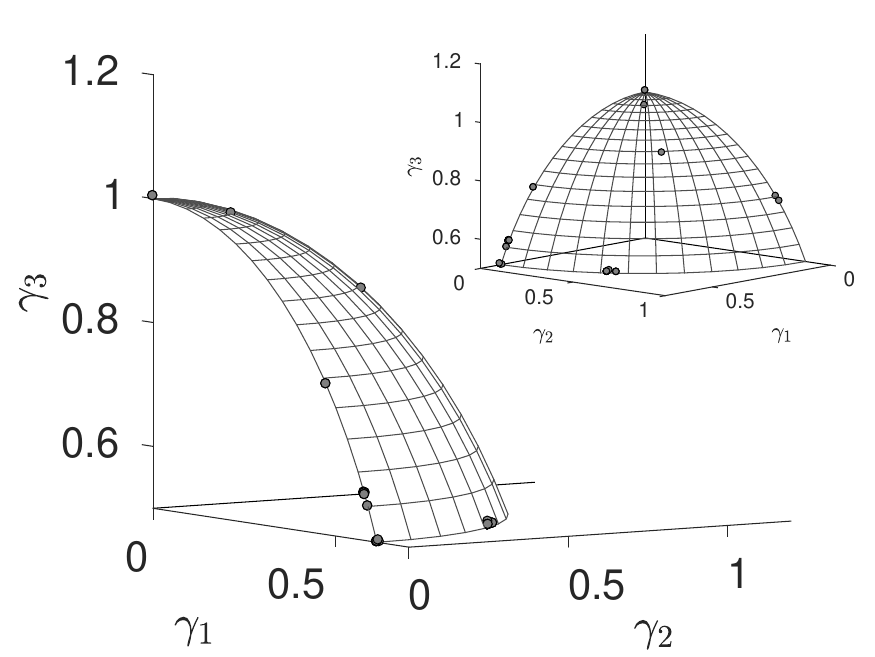}}
\subfigure[IDBEA on DPF5A]{
\centering
\includegraphics[width=0.18\textwidth]{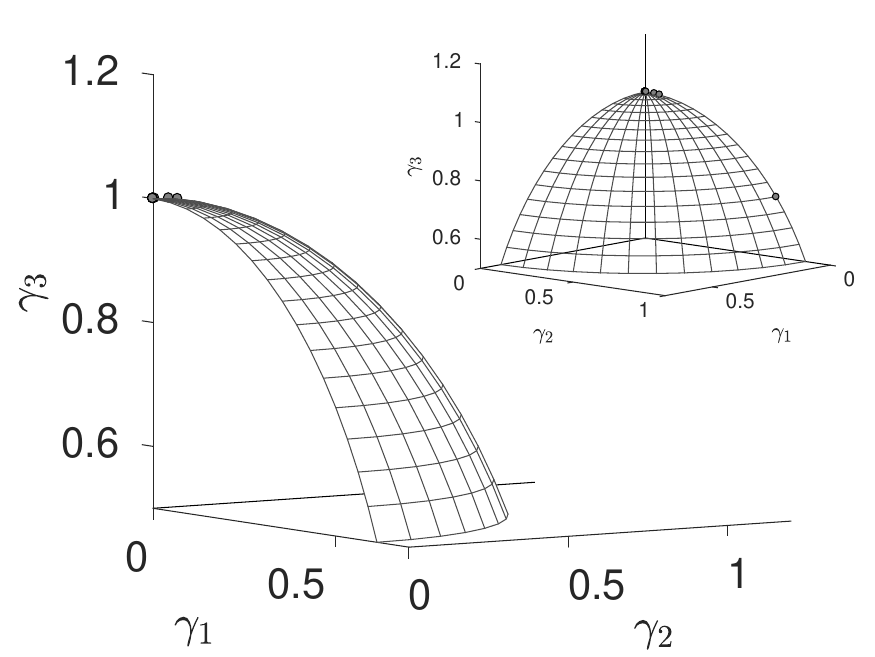}}
\caption{The solution set of the five tested algorithms on DPF1A--PDF5A with $m =10$, $d = 3$ in the run associated with its best HV value (measured in essential objective space), where the grid mesh denotes the PF of the problem in the essential objective space. From top to the bottom are the results on DPF1A to PDF4A, and the degenerate part of DPF5A. For the test instance of DPF5A, the scatter plot only shows the degenerate part of the PF and the solutions whose last objective values are not less than $0.5$ (since the last objective value of the degenerate part of the PF is not less than $0.5$.)}
\label{fig:comparison}
\end{figure*}

\begin{figure*}[hbpt]
\centering
\subfigure[$\delta$-MOSS on DPF5A]{
\centering
\includegraphics[width=0.18\textwidth]{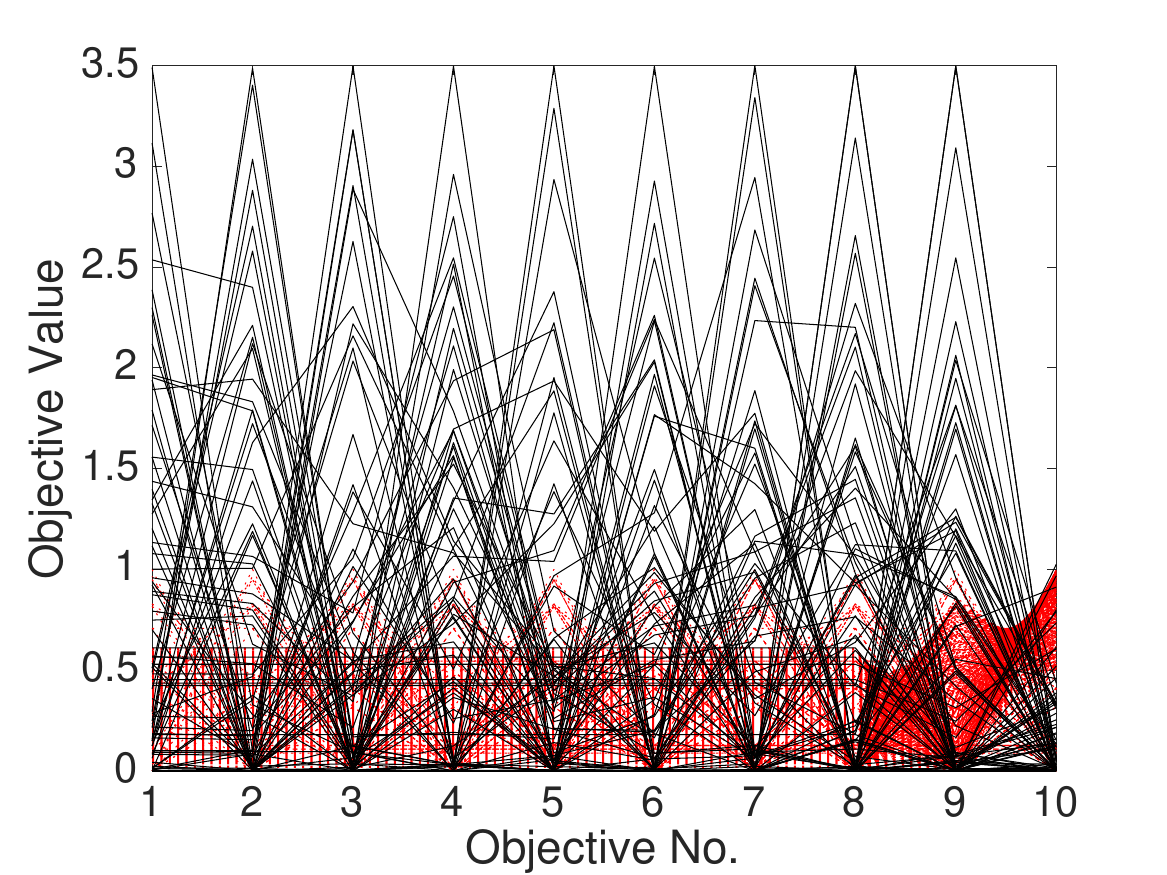}}
\subfigure[NCIE on DPF5A]{
\centering
\includegraphics[width=0.18\textwidth]{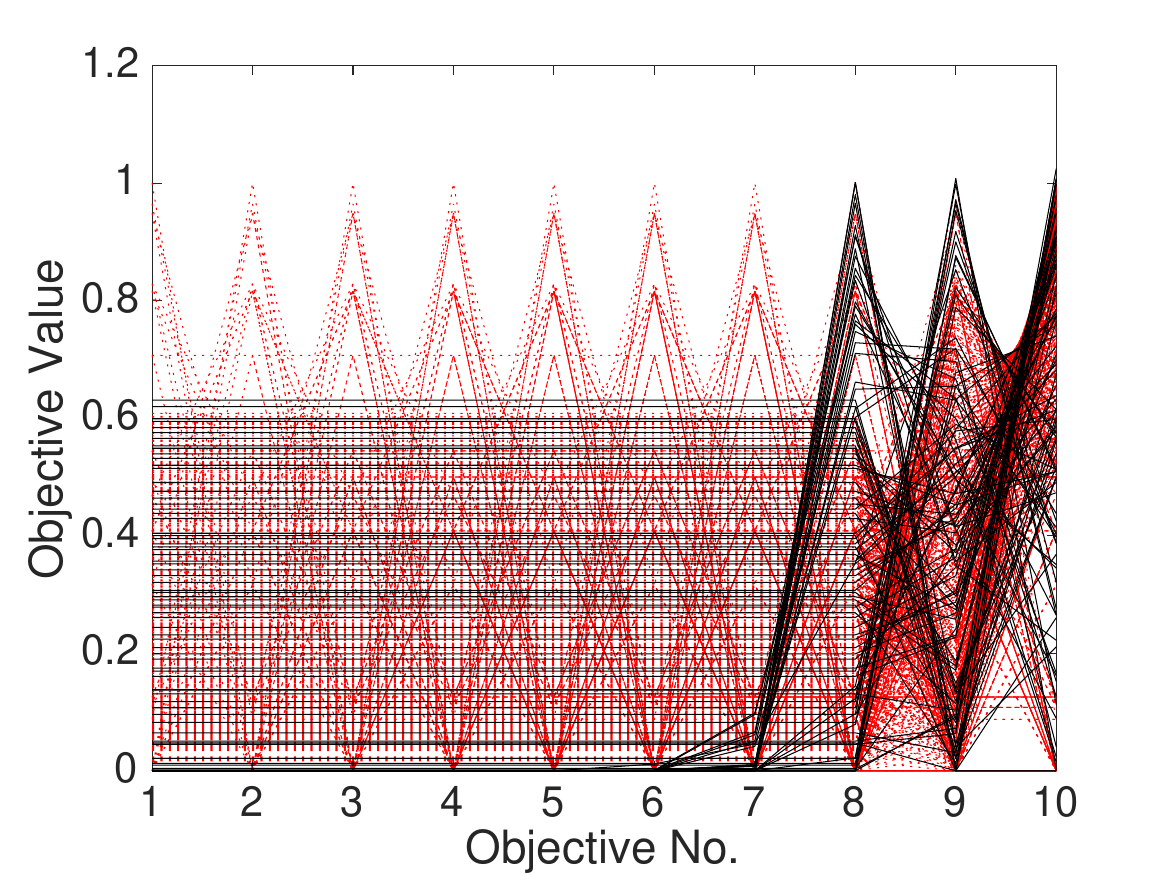}}
\subfigure[OSP on DPF5A]{
\centering
\includegraphics[width=0.18\textwidth]{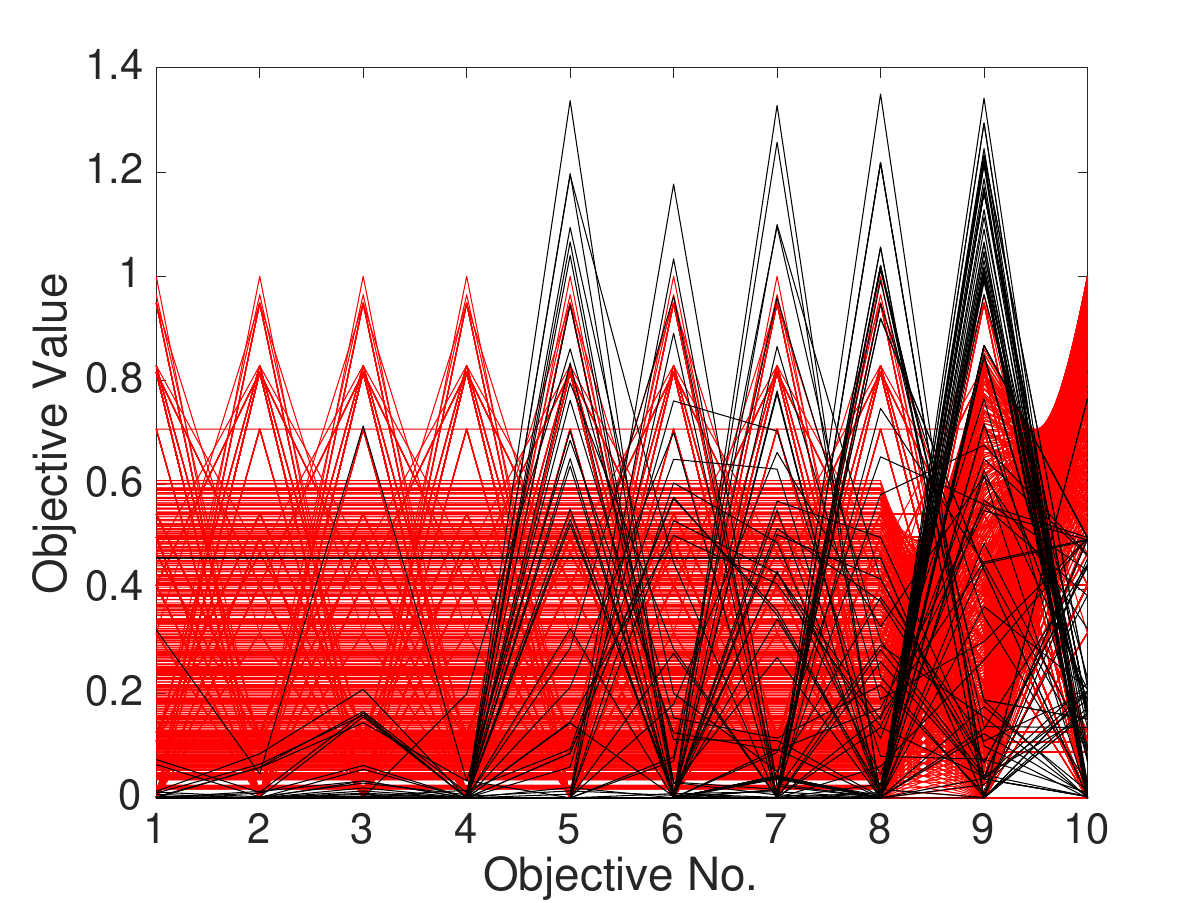}}
\subfigure[IBEA on DPF5A]{
\centering
\includegraphics[width=0.18\textwidth]{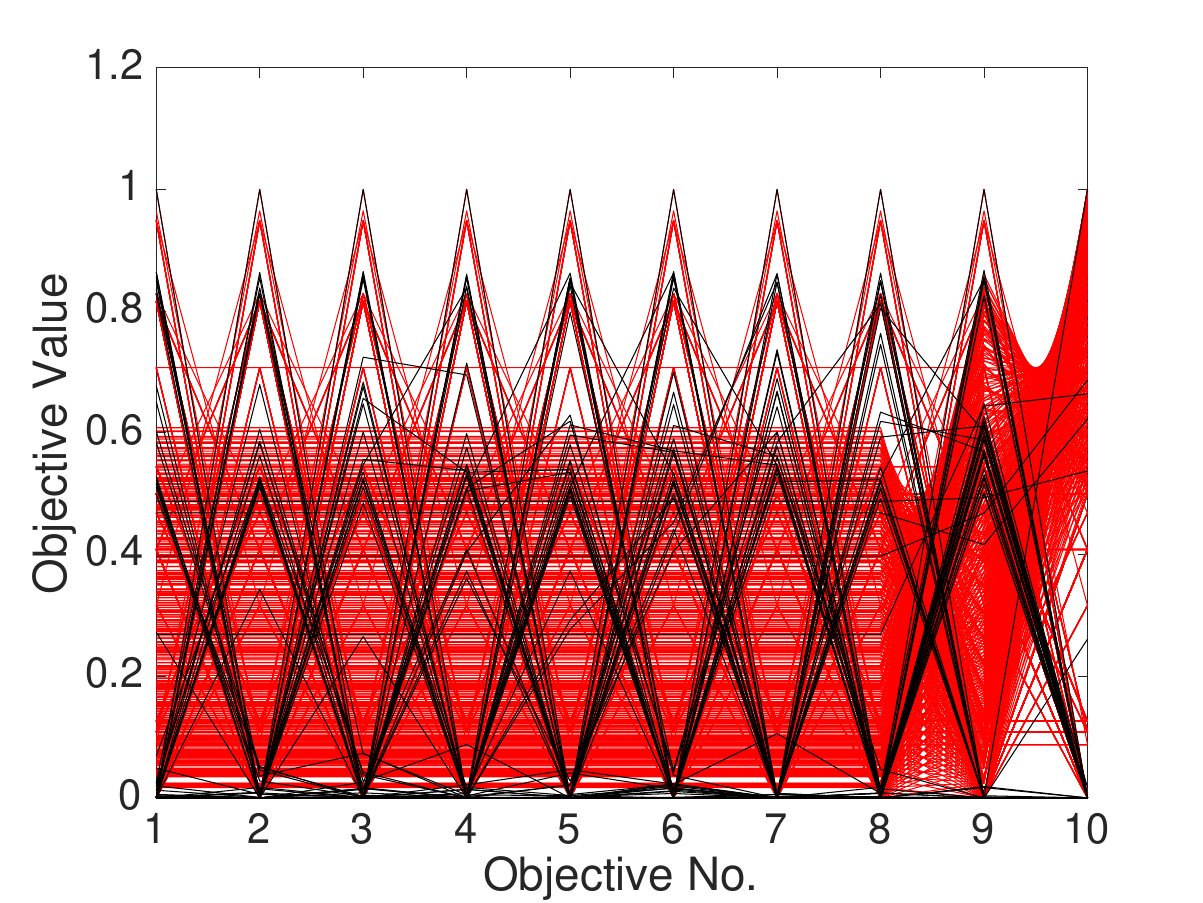}}
\subfigure[IDBEA on DPF5A]{
\centering
\includegraphics[width=0.18\textwidth]{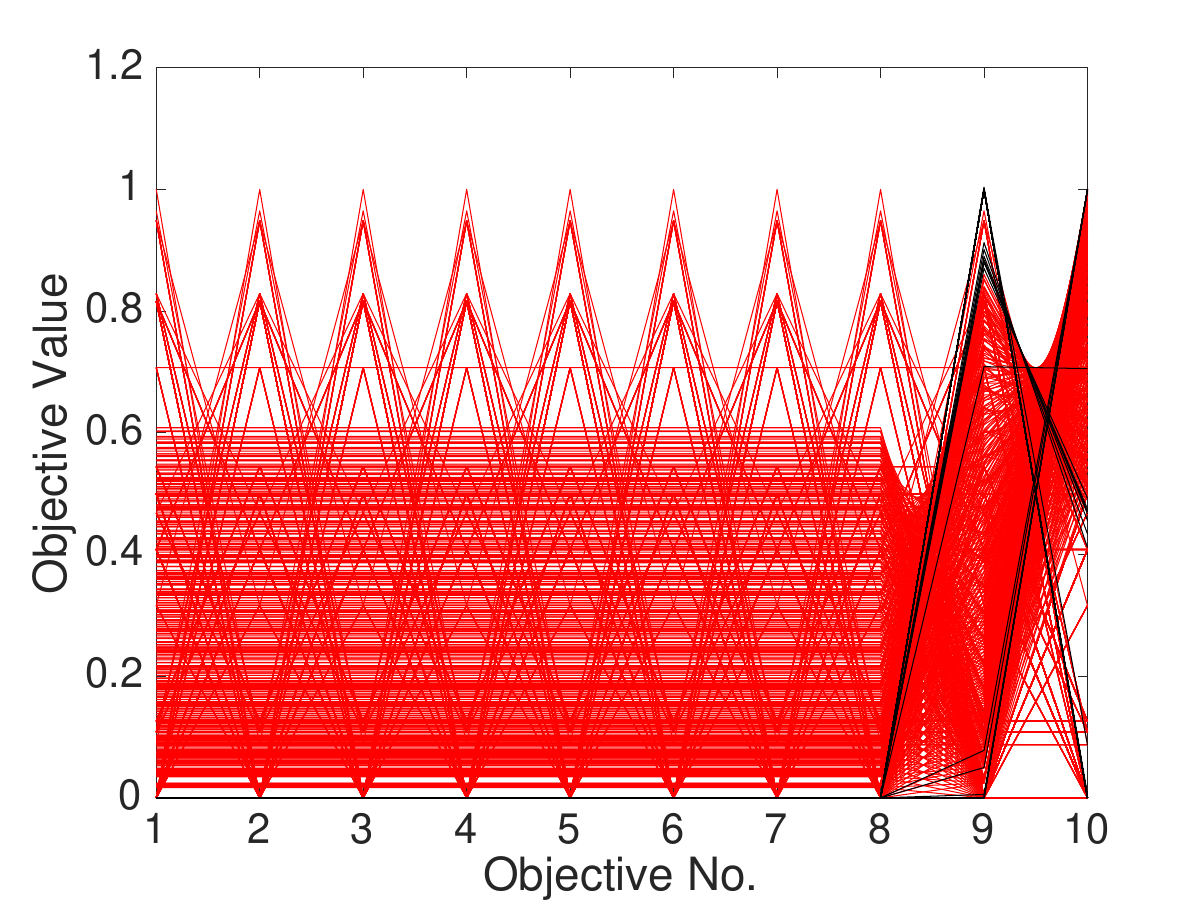}}
\caption{The solution set of the five algorithms on $10$-objective PDF5A with $d=3$ in the run associated with its best HV value, where the red dot lines represent the reference points sampled from the PF of the problem and the black lines denote the solutions.}
\label{fig:DPF5C}
\end{figure*}

The results of the run with the best HV value\footnote{The results of the run with the median HV value of the tested algorithms can be found in Supplementary Material (Section III).} (measured in the essential objective space) on DTLZ5($I$, $M$) and on the proposed problems are shown in Fig.~\ref{fig:comparisonV} and Fig.~\ref{fig:comparison}, respectively. From the results, we have the following observations:

1) The first four algorithms, $\delta$-MOSS, OSP, NCIE and IBEA, can obtain a solution set that has a good convergence and diversity to the PF of DTLZ5($I$, $M$). Since the first $(M-I+1)$ objectives of DTLZ5($I$, $M$) are linearly dependent on each other, it is easier for the objective reduction algorithms to discover the essential objectives of the problem. However, the decomposition-based algorithm still has difficulties to obtain a diverse solution set.

2) The solution sets of OSP and IDBEA have a poor diversity on DPF1A compared with the results on DTLZ5($I$, $M$), and the other three algorithms can obtain fairly good results. On DPF2A, the performance of $\delta$-MOSS and NCIE decreases slightly compared with the results on DPF1A, and OSP and IDBEA still cannot obtain diverse solution sets. In addition, the results obtained by these algorithms on DPF1A and DPF2A are not as diverse as the results on DTLZ5($I$, $M$) since the relation between the redundant objectives and the essential objectives is mutually linearly correlated in DTLZ5($I$, $M$), linearly correlated (but not mutually linearly correlated) in DPF1A, and nonlinearly correlated in DPF2A.

3) IDBEA fails to converge to the PF on DPF3A and DPF4A. The other four methods can converge to the PF, but they fail to maintain the diversity of the solution set. These two test problems are harder than DPF1A and DPF2A since their redundant objectives exist implicitly whereas the redundant objectives of DPF1A and DPF2A exist explicitly.

4) None of these algorithms can obtain good results on the degenerate part of the PF on DPF5A. We also show the parallel coordinate plot of the whole solution sets of these five algorithms in Fig.~\ref{fig:DPF5C}, from which we can see that a large number of the solutions obtained by $\delta$-MOSS do not converge to the PF. The objective value range of $\delta$-MOSS's solutions is approximately from $0$ to $3.5$, which is far from that of the PF (from $0$ to $1$). The value of the last objective for the degenerate part of DPF5A's PF is larger than  $0.5$. OSP has only one individual lying on the degenerate part of the PF as shown in Fig.~\ref{fig:comparison}(w). Even though NCIE and IDBEA can obtain some solutions that lie on the degenerate part of the PF on DPF5A, they fail to obtain diverse solutions in the non-degenerate part of the PF. It can be seen from Fig.~\ref{fig:DPF5C}(b) and  Fig.~\ref{fig:DPF5C}(e) that a large part of the PF is not covered by the solutions of NCIE and IDBEA.

\section{Discussion}
\label{Sec:6}
Based on the analysis of the experimental results, we can see that the proposed three characteristics have a significant impact on the performance of the existing algorithms. They bring different types of difficulties for objective reduction techniques. Specifically, the explicit redundancy of objectives may make the algorithms hard to obtain diverse solution set; the implicit redundancy among objectives posts a big challenge to the objective reduction methods based on objective selection; the partial redundancy makes all the methods struggle to find/maintain well-distributed solutions on both degenerate and non-degenerate segments of the PF. Overall, none of the tested algorithms can obtain good performance on the high-dimensional degenerate instances, which has shown the difficulty of the proposed problem suite.

We have the following observations regarding why certain algorithms would struggle in optimising these problems and provide some potentially useful suggestions to degenerate multiobjective optimisation problems.

\subsection{Regarding the Pareto-dominance-based algorithms}
Pareto-dominance-based algorithms (and their variants), such as NSGA-II and SPEA2+SDE, are good options for low-dimensional degenerate problems; With more objectives being involved, the performance of this category of approaches degrades significantly. Existing objective reduction techniques have shown their effectiveness in improving the performance of Pareto-dominance-based algorithms on the problems with explicit redundancy. However, it is still struggling for them to solve the problems with implicit redundancy, which demonstrates the need for more effective objective reduction techniques. A potentially useful way to extract/estimate the essential objectives is to analyse the population information during the evolution process. More specifically, the objectives can be reflected by the objective values of the individuals in the population. Each essential objective could be expressed as (a function of) a combination of the problem objectives. The expression aims to maximise the conflict between the exacted essential objectives and satisfy the constraint: the dominance relation between the individuals and the distribution of the individuals should be preserved as much as possible in the essential objective space by comparing with the dominance relation and the distribution in the original problem objective space. We have already devoted our effort on objective reduction for visualising many-objective solution sets by following this guideline~\cite{zhen2018obj}.


\subsection{Regarding the decomposition-based algorithms}
Decomposition-based algorithms fail to obtain good results on degenerate problems since a large proportion of the weight vectors may be far from the PF~\cite{Li2017}. The effectiveness of using the adaptation mechanism~\cite{Asafuddoula2015, asafuddowla2018} in IDBEA has been verified in our experiments. However, the adaption mechanism in IDBEA is not strong enough to handle the proposed degenerate test problems. We recommend generating weight vectors by using the information of the whole evolution process. Specifically, we can employ machine learning approaches to discover the subspace where the Pareto-front lies on and to generate weight vectors in this subspace to explore and to exploit. Furthermore, the reference information~\cite{Bechikh2015} may be helpful to improve the performance of the decomposition-based algorithms in solving a degenerate problem.

\subsection{Regarding the indicator-based algorithms}
The indicator-based algorithms adopt an indicator to select better candidates during the evolution process. The indicator-based algorithm IBEA performs better than the peer algorithms on DPF3, where the problem objectives are linearly dependent on the essential objectives. This is due to that, in this case, the solution set with a better value in terms of the adopted indicator on the problem objectives would have a better value on the essential objectives as well. However, it performs poorly on DPF4, where the problem objectives are nonlinearly dependent on the essential objectives. In addition,  indicators which treat each problem equally, such as HV metric, may not good alternatives for solving degenerate problems. A potentially useful strategy for this kind of algorithms to solve the degenerated multiobjective problems is to consider an indicator which can obtain better value on the problem objectives when the solution set consists of the candidates measured on the essential objectives.

\subsection{Regarding the partial redundancy}
The partial redundancy brings a big challenge to all the tested algorithms. The existing MOEAs are hard to find/maintain well-distributed solutions on both degenerate and non-degenerate segments of the PF. A potential way to deal with this is to maintain two subpopulations to search the degenerate segment and the non-degenerate segment of the PF, respectively. The subpopulation used to search for the degenerate segment should consider the redundancy of the objectives and is expected to obtain solutions that lie on a low-dimensional subspace. The subpopulation used to search for the non-degenerate segment should emphasise the diversity of the non-dominated solution set among the whole problem objective space.

\section{Conclusion}
\label{Sec:7}
This paper discusses three characteristics that lead to degenerate multiobjective optimisation problems (MOPs), \textit{i.e.}, explicitly redundant objectives, implicitly redundant objectives, and partially redundant objectives. The first two characteristics make the problem with a complete degenerate PF, while the third one results in a partially degenerate PF for the problem.

Five test problems are designed based on these three characteristics with a uniform formulation. Among them, DPF1 and DPF2 have explicitly redundant objectives, DFP3 and DPF4 have implicitly redundant objectives, and DPF5 has partially redundant objectives. DPF1 and DPF2 are designed to test an algorithm's ability of the objective selection, DPF3 and DPF4 to test an algorithm's ability of objective extraction, and DPF5 to test an algorithm's ability to maintain different sub-populations on the degenerate and the non-degenerate segments of the PF. The proposed test problems, with flexible attributes, are much more generic than existing degenerate problems, making it accessible to conducting a comprehensive study of various optimisation methods on degenerate problems.

Ten representative MOEAs have been tested on the proposed problems. In contrast to existing degenerate problems, our problems have introduced new features (with varying difficulty) that can challenge various objective reduction methods. This has been evidenced in our experimental studies where none of the tested MOEAs is able to well solve all the proposed problems. This, therefore, suggests a need of developing new methods to solve MOPs with degenerate PFs.

In future work, we will attempt to develop effective (linear and nonlinear) objective reduction approaches based on the information obtained from the whole evolution process. Then, we can integrate these objective reduction approaches into existing evolutionary algorithms to solve degenerate MaOPs.

\begin{appendices}
\section{Proof of Theorem 1}
\label{ProofT1}
{\it Proof of Sufficiency}: For any $\mathbf{x}_i, \mathbf{x}_j \in \Omega$,  if $\mathbf{x}_i \prec \mathbf{x}_j$ in the original objective space, we have that
\begin{equation}
\label{eq.a11}
\begin{array}{c}
f_{1}(\mathbf{x}_i) \leq f_{1}(\mathbf{x}_j),\\
\vdots\\
f_{d}(\mathbf{x}_i) \leq f_{d}(\mathbf{x}_j),\\
\end{array}
\end{equation}
and $\exists \mu \in \{1, 2, \dots, d\}$ that satisfies
\begin{equation}
\label{eq.a11b}
f_{\mu}(\mathbf{x}_i) < f_{\mu}(\mathbf{x}_j).
\end{equation}

Considering a new objective $f_{\nu}, \nu \in \{d+1, d+2, \dots, m\}$, we obtain that
\begin{equation}
\label{eq.a12}
\begin{array}{c}
f_{\nu}(\mathbf{x}_i) = h_{\nu}(f_1(\mathbf{x}_i), \dots, f_m(\mathbf{x}_i)),\\
f_{\nu}(\mathbf{x}_j) = h_{\nu}(f_1(\mathbf{x}_j), \dots, f_m(\mathbf{x}_j)).\\
\end{array}
\end{equation}
Combining (\ref{eq.a12}) and (\ref{eq.a11}), and based on the fact that $h_{\nu}$ is a non-decreasing function corresponding to $f_1, f_2, \dots, f_d$, it flows that

\begin{equation}
\label{eq.a13}
f_{\nu}(\mathbf{x}_i) \leq f_{\nu}(\mathbf{x}_j),
\end{equation}
such that we can draw the conclusion that the $\mathbf{x}_i \prec \mathbf{x}_j$ in the new objective space.

{\it Proof of Necessity}: For any $\mathbf{x}_i, \mathbf{x}_j \in \Omega$,  if $\mathbf{x}_i \prec \mathbf{x}_j$ in the new objective space. Since the set of the original objectives is a subset of the set of the new objectives, there are two cases:

a) $\mathbf{x}_i \prec \mathbf{x}_j$ in the space spanned by the first $d$ objectives, which directly completes the proof.

b) the objective values of $\mathbf{x}_i$ and $\mathbf{x}_j$ are equal on the original objectives, \textit{i.e.},
\begin{equation}
\label{eq.a14}
\begin{array}{c}
f_{1}(\mathbf{x}_i) = f_{1}(\mathbf{x}_j),\\
\vdots\\
f_{d}(\mathbf{x}_i) = f_{d}(\mathbf{x}_j),\\
\end{array}
\end{equation}
and $\exists \psi \in \{d+1, d+2, \dots, m\}$ that satisfies
\begin{equation}
\label{eq.a14b}
f_{\psi}(\mathbf{x}_i) < f_{\psi}(\mathbf{x}_j).
\end{equation}

From the definition of the objective $f_\psi$, we have
\begin{equation}
\label{eq.a15}
\begin{array}{c}
f_{\psi}(\mathbf{x}_i) = h_{\psi}(f_1(\mathbf{x}_i), \dots, f_m(\mathbf{x}_i)),\\
f_{\psi}(\mathbf{x}_j) = h_{\psi}(f_1(\mathbf{x}_j), \dots, f_m(\mathbf{x}_j)).\\
\end{array}
\end{equation}

Combining the results in (\ref{eq.a14}) and (\ref{eq.a15}), we have $f_\psi(\mathbf{x}_i) = f_\psi(\mathbf{x}_j)$, which contradicts with the result in (\ref{eq.a14b}). This means the case b) does not exist, \textit{i.e.}, $\mathbf{x}_i \prec \mathbf{x}_j$ holds in the original objective space.
This completes the proof.

\section{Proof of Theorem 2}
\label{ProofT2}
Let us first consider the situation of the problem with three objectives, and there are only two essential objectives.

{\it Proof of Sufficiency}: For any $\mathbf{x}_i, \mathbf{x}_j \in \Omega$,  if $\mathbf{x}_i \prec \mathbf{x}_j$ in the original objective space, we have that

\begin{equation}
\label{eq.a21}
\begin{array}{c}
\gamma_{1}(\mathbf{x}_i) \leq \gamma_{1}(\mathbf{x}_j),\\
\gamma_{2}(\mathbf{x}_i) \leq \gamma_{2}(\mathbf{x}_j),\\
\end{array}
\end{equation}
and $\exists \mu \in \{1, 2\}$ that satisfies
\begin{equation}
\label{eq.a21b}
\gamma_{\mu}(\mathbf{x}_i) < \gamma_{\mu}(\mathbf{x}_j).
\end{equation}

Supposing that
\begin{equation}
\label{eq.a22}
\begin{array}{c}
\gamma_{1}(\mathbf{x}_i) \leq \gamma_{1}(\mathbf{x}_j);\\
\gamma_{2}(\mathbf{x}_i) < \gamma_{2}(\mathbf{x}_j),\\
\end{array}
\end{equation}
we obtain that
\begin{equation}
\label{eq.a23}
\begin{array}{c}
f_{1}(\mathbf{x}_i) \leq f_{1}(\mathbf{x}_j)\\
\end{array}
\end{equation}
and the following three possibilities:

Case 1): $\gamma_{2}(\mathbf{x}_i)> \eta_1$
\begin{equation}
\label{eq.a24}
\begin{array}{c}
f_{2}(\mathbf{x}_i)= f_{1}(\mathbf{x}_j);\\
f_{3}(\mathbf{x}_i)< f_{3}(\mathbf{x}_j).\\
\end{array}
\end{equation}

Case 2): $\gamma_{2}(\mathbf{x}_i) < \eta_1 < \gamma_{2}(\mathbf{x}_j)$
\begin{equation}
\label{eq.a25}
\begin{array}{c}
f_{2}(\mathbf{x}_i) < f_{1}(\mathbf{x}_j);\\
f_{3}(\mathbf{x}_i) < f_{3}(\mathbf{x}_j).\\
\end{array}
\end{equation}

Case 3): $\eta_1 > \gamma_{2}(\mathbf{x}_j)$
\begin{equation}
\label{eq.a26}
\begin{array}{c}
f_{2}(\mathbf{x}_i) < f_{1}(\mathbf{x}_j);\\
f_{3}(\mathbf{x}_i) = f_{3}(\mathbf{x}_j).\\
\end{array}
\end{equation}

Based on (\ref{eq.a23}) and (\ref{eq.a24})-(\ref{eq.a26}), we can draw the conclusion that the $\mathbf{x}_i \prec \mathbf{x}_j$ in the new objective space.

{\it Proof of Necessity}: For any $\mathbf{x}_i, \mathbf{x}_j \in \Omega$,  if $\mathbf{x}_i \prec \mathbf{x}_j$ in the new objective space, we have that
\begin{equation}
\label{eq.a27}
\begin{array}{c}
f_{1}(\mathbf{x}_i) \leq f_{1}(\mathbf{x}_j),\\
f_{2}(\mathbf{x}_i) \leq f_{2}(\mathbf{x}_j),\\
f_{3}(\mathbf{x}_i) \leq f_{3}(\mathbf{x}_j),\\
\end{array}
\end{equation}
and $\exists \nu \in \{1, 2, 3\}$ that satisfies
\begin{equation}
\label{eq.a28}
f_{\nu}(\mathbf{x}_i) < f_{\nu}(\mathbf{x}_j).
\end{equation}

If $f_{1}(\mathbf{x}_i) < f_{1}(\mathbf{x}_j)$, it is clear that
\begin{equation}
\label{eq.a210}
\begin{array}{c}
\gamma_{1}(\mathbf{x}_i) < \gamma_{1}(\mathbf{x}_j);\\
\gamma_{2}(\mathbf{x}_i) \leq \gamma_{2}(\mathbf{x}_j);\\
\end{array}
\end{equation}

If $f_{1}(\mathbf{x}_i) = f_{1}(\mathbf{x}_j)$, we have that
\begin{equation}
\label{eq.a211}
\gamma_{1}(\mathbf{x}_i) = \gamma_{1}(\mathbf{x}_j),
\end{equation}
and the following two possibilities:

Case 1): $f_{2}(\mathbf{x}_i) < f_{2}(\mathbf{x}_j)$
\begin{equation}
\label{eq.a212}
\gamma_{2}(\mathbf{x}_i) < \gamma_{2}(\mathbf{x}_j).
\end{equation}

Case 2): $f_{3}(\mathbf{x}_i) < f_{3}(\mathbf{x}_j)$
\begin{equation}
\label{eq.a213}
\gamma_{2}(\mathbf{x}_i) < \gamma_{2}(\mathbf{x}_j).
\end{equation}

Combining the results in (\ref{eq.a210}) and (\ref{eq.a211})-(\ref{eq.a213}), we find that $\mathbf{x}_i \prec \mathbf{x}_j$ holds in the original objective space as well.

It is clear that the above analysis is also true to the situation with any number of objectives.
This completes the proof.

\section{Essential Objectives of DPF1A-DPF5A}
\label{DefinitionT3}
The definition of the essential objectives of DPF1A-DPF4A and the degenerate part of DPF5A is the same as that of DTLZ5($I$, $M$)~\cite{Deb2005on}:
\begin{equation}
\label{eq6.1a}
\begin{split}
&\gamma_1(\mathbf{x}) = \cos(\theta_1)\dots \cos(\theta_{d-2})\cos(\theta_{d-1})\sin(\frac{\pi}{4})(1+g(\mathbf{x}^r));\\
&\gamma_2(\mathbf{x}) = \cos(\theta_1)\dots \cos(\theta_{d-2})\sin(\theta_{d-1})(1+g(\mathbf{x}^r));\\
&\hspace{26mm} \vdots\\
&\gamma_{d-1}(\mathbf{x}) = \cos(\theta_1)\sin(\theta_2)(1+g(\mathbf{x}^r));\\
&\gamma_d(\mathbf{x}) = \sin(\theta_1)(1+g(\mathbf{x}^r));\\
&g(\mathbf{x}^r) = \sum_{i = d}^n(x_i - 0.5)^2 ;\\
&\theta_j = \frac{\pi}{2} x_j \hspace{4mm}\text{for}\hspace{4mm}  j\in \{1, \dots, d-1\},
\end{split}
\end{equation}
where $n$ is the number of decision variables, and $\mathbf{x}^r = (x_d, \dots, x_n)$.

\end{appendices}

\section*{Acknowledgments}
This work was supported by the Ministry of Science and Technology of China~(Grant No. 2017YFC0804003), the Science and Technology Innovation Committee Foundation of Shenzhen~(Grant No. ZDSYS201703031748284), Shenzhen Peacock Plan~(Grant No. KQTD2016112514355531), the Program for Guangdong Introducing Innovative and Entrepreneurial Teams(Grant No. 2017ZT07X386), the Program for University Key Laboratory of Guangdong Province(Grant No. 2017KSYS008), and the Engineering and Physical Sciences Research Council (EPSRC) of U.K.~(Grant Nos. EP/J017515/1 and EP/P005578/1).

\bibliographystyle{IEEEtran}
\begin{small}
\bibliography{mybib}

\begin{thebibliography}{10}
\providecommand{\url}[1]{#1}
\csname url@samestyle\endcsname
\providecommand{\newblock}{\relax}
\providecommand{\bibinfo}[2]{#2}
\providecommand{\BIBentrySTDinterwordspacing}{\spaceskip=0pt\relax}
\providecommand{\BIBentryALTinterwordstretchfactor}{4}
\providecommand{\BIBentryALTinterwordspacing}{\spaceskip=\fontdimen2\font plus
\BIBentryALTinterwordstretchfactor\fontdimen3\font minus
  \fontdimen4\font\relax}
\providecommand{\BIBforeignlanguage}[2]{{%
\expandafter\ifx\csname l@#1\endcsname\relax
\typeout{** WARNING: IEEEtran.bst: No hyphenation pattern has been}%
\typeout{** loaded for the language `#1'. Using the pattern for}%
\typeout{** the default language instead.}%
\else
\language=\csname l@#1\endcsname
\fi
#2}}
\providecommand{\BIBdecl}{\relax}
\BIBdecl

\bibitem{Ishibuchi2016de}
H.~Ishibuchi, H.~Masuda, and Y.~Nojima, ``Pareto fronts of many-objective
  degenerate test problems,'' \emph{IEEE Transactions on Evolutionary
  Computation}, vol.~20, no.~5, pp. 807--813, Oct 2016.

\bibitem{jain1990theory}
P.~Jain and A.~M. Agogino, ``Theory of design: {An} optimization perspective,''
  \emph{Mechanism and Machine Theory}, vol.~25, no.~3, pp. 287--303, 1990.

\bibitem{musselman1980tradeoff}
K.~Musselman and J.~Talavage, ``A tradeoff cut approach to multiple objective
  optimization,'' \emph{Operations Research}, vol.~28, no.~6, pp. 1424--1435,
  1980.

\bibitem{gu2001optimisation}
L.~Gu, R.~Yang, C.-H. Tho, M.~Makowskit, O.~Faruquet, and Y.~L. Y.~Li,
  ``Optimisation and robustness for crashworthiness of side impact,''
  \emph{International Journal of Vehicle Design}, vol.~26, no.~4, pp. 348--360,
  2001.

\bibitem{sinha2013using}
A.~Sinha, D.~K. Saxena, K.~Deb, and A.~Tiwari, ``Using objective reduction and
  interactive procedure to handle many-objective optimization problems,''
  \emph{Applied Soft Computing}, vol.~13, no.~1, pp. 415--427, 2013.

\bibitem{Hierons2016}
R.~M. Hierons, M.~Li, X.~Liu, S.~Segura, and W.~Zheng, ``{SIP: Optimal} product
  selection from feature models using many-objective evolutionary
  optimisation,'' \emph{ACM Transactions on Software Engineering and
  Methodology}, vol.~25, no.~3, 2016.

\bibitem{Deb2005a}
K.~Deb, L.~Thiele, M.~Laumanns, and E.~Zitzler, ``Scalable test problems for
  evolutionary multiobjective optimization,'' in \emph{Evolutionary
  Multiobjective Optimization: Theoretical Advances and Applications},
  A.~Abraham, L.~Jain, and R.~Goldberg, Eds.\hskip 1em plus 0.5em minus
  0.4em\relax London: Springer London, 2005, pp. 105--145.

\bibitem{huband2005scalable}
S.~Huband, L.~Barone, L.~While, and P.~Hingston, ``A scalable multi-objective
  test problem toolkit,'' in \emph{International Conference on Evolutionary
  Multi-Criterion Optimization}.\hskip 1em plus 0.5em minus 0.4em\relax
  Springer, Berlin, Heidelberg, 2005, pp. 280--295.

\bibitem{Huband2006}
S.~Huband, P.~Hingston, L.~Barone, and L.~While, ``A review of multiobjective
  test problems and a scalable test problem toolkit,'' \emph{IEEE Trans. on
  Evolutionary Computation}, vol.~10, no.~5, pp. 477--506, October 2006.

\bibitem{Deb2005on}
K.~Deb and D.~K. Saxena, ``On finding {Pareto-optimal} solutions through
  dimensionality reduction for certain large-dimensional multi-objective
  optimization problems,'' Indian Institute of Technology Kanpur, Tech. Rep.
  2005011, 2005.

\bibitem{Saxena2013objective}
D.~K. Saxena, J.~A. Duro, A.~Tiwari, K.~Deb, and Q.~Zhang, ``Objective
  reduction in many-objective optimization: {Linear} and nonlinear
  algorithms,'' \emph{IEEE Transactions on Evolutionary Computation}, vol.~17,
  no.~1, pp. 77--99, 2013.

\bibitem{karlsson1988scanline}
R.~G. Karlsson and M.~H. Overmars, ``Scanline algorithms on a grid,'' \emph{BIT
  Numerical Mathematics}, vol.~28, no.~2, pp. 227--241, 1988.

\bibitem{Ishibuchi2013many}
H.~Ishibuchi, M.~Yamane, N.~Akedo, and Y.~Nojima, ``Many-objective and
  many-variable test problems for visual examination of multiobjective
  search,'' in \emph{2013 IEEE Congress on Evolutionary Computation}, June
  2013, pp. 1491--1498.

\bibitem{Cheung2016}
Y.~Cheung, F.~Gu, and H.~L. Liu, ``Objective extraction for many-objective
  optimization problems: Algorithm and test problems,'' \emph{IEEE Transactions
  on Evolutionary Computation}, vol.~20, no.~5, pp. 755--772, Oct 2016.

\bibitem{Rectangle}
M.~Li, S.~Yang, and X.~Liu, ``A test problem for visual investigation of
  high-dimensional multi-objective search,'' in \emph{Proceedings of the IEEE
  Congress on Evolutionary Computation (CEC)}, 2014, pp. 2140--2147.

\bibitem{Li2017}
M.~Li, C.~Grosan, S.~Yang, X.~Liu, and X.~Yao, ``Multiline distance
  minimization: {A} visualized many-objective test problem suite,'' \emph{IEEE
  Transactions on Evolutionary Computation}, vol.~22, no.~1, pp. 61--78, Feb
  2018.

\bibitem{saxena2011framework}
D.~K. Saxena, Q.~Zhang, J.~A. Duro, and A.~Tiwari, ``Framework for
  many-objective test problems with both simple and complicated pareto-set
  shapes,'' in \emph{International Conference on Evolutionary Multi-Criterion
  Optimization}.\hskip 1em plus 0.5em minus 0.4em\relax Springer, 2011, pp.
  197--211.

\bibitem{liu2017adaptively}
H.-L. Liu, L.~Chen, Q.~Zhang, and K.~Deb, ``Adaptively allocating search effort
  in challenging many-objective optimization problems,'' \emph{IEEE
  Transactions on Evolutionary Computation}, 2017.

\bibitem{Brockhoff2009}
D.~Brockhoff and E.~Zitzler, ``Objective reduction in evolutionary
  multiobjective optimization: Theory and applications,'' \emph{Evolutionary
  Computation}, vol.~17, no.~2, pp. 135--166, 2009.

\bibitem{Singh2011}
H.~K. Singh, A.~Isaacs, and T.~Ray, ``A {Pareto} corner search evolutionary
  algorithm and dimensionality reduction in many-objective optimization
  problems,'' \emph{IEEE Trans. on Evolutionary Computation}, vol.~15, no.~4,
  pp. 539--556, August 2011.

\bibitem{jaimes2014objective}
A.~L. Jaimes, C.~A.~C. Coello, H.~Aguirre, and K.~Tanaka, ``Objective space
  partitioning using conflict information for solving many-objective
  problems,'' \emph{Information Sciences}, vol. 268, pp. 305--327, 2014.

\bibitem{Wang2016}
H.~Wang and X.~Yao, ``Objective reduction based on nonlinear correlation
  information entropy,'' \emph{Soft Computing}, vol.~20, no.~6, pp. 2393--2407,
  2016.

\bibitem{yuan2017objective}
Y.~Yuan, Y.~S. Ong, A.~Gupta, and H.~Xu, ``Objective reduction in
  many-objective optimization: Evolutionary multiobjective approaches and
  comprehensive analysis,'' \emph{IEEE Transactions on Evolutionary
  Computation}, vol.~22, no.~2, pp. 189--210, 2018.

\bibitem{Cheng2016}
R.~Cheng, Y.~Jin, M.~Olhofer, and B.~sendhoff, ``Test problems for large-scale
  multiobjective and many-objective optimization,'' \emph{IEEE Transactions on
  Cybernetics}, vol.~47, no.~12, pp. 4108--4121, Dec 2017.

\bibitem{Deb2002}
K.~Deb, A.~Pratap, S.~Agarwal, and T.~Meyarivan, ``A fast and elitist
  multiobjective genetic algorithm: {NSGA-II},'' \emph{IEEE Trans. on
  Evolutionary Computation}, vol.~6, no.~2, pp. 182--197, 2002.

\bibitem{Zitzler2001}
E.~Zitzler, M.~Laumanns, and L.~Thiele, ``{SPEA2: Improving} the strength
  pareto evolutionary algorithm for multiobjective optimization,'' in
  \emph{Evolutionary Methods for Design, Optimisation and Control}.\hskip 1em
  plus 0.5em minus 0.4em\relax International Center for Numerical Methods in
  Engineering, 2002, pp. 95--100.

\bibitem{Li2014a}
M.~Li, S.~Yang, and X.~Liu, ``Shift-based density estimation for {Pareto}-based
  algorithms in many-objective optimization,'' \emph{IEEE Trans. on
  Evolutionary Computation}, vol.~18, no.~3, pp. 348--365, June 2014.

\bibitem{Deb2014}
K.~Deb and H.~Jain, ``An evolutionary many-objective optimization algorithm
  using reference-point-based nondominated sorting approach, part {I}: Solving
  problems with box constraints,'' \emph{IEEE Transactions on Evolutionary
  Computation}, vol.~18, no.~4, pp. 577--601, Aug 2014.

\bibitem{Zhang2007}
Q.~Zhang and H.~Li, ``{MOEA/D: A} multiobjective evolutionary algorithm based
  on decomposition,'' \emph{IEEE Trans. on Evolutionary Computation}, vol.~11,
  no.~6, pp. 712--731, December 2007.

\bibitem{Asafuddoula2015}
M.~Asafuddoula, T.~Ray, and R.~Sarker, ``A decomposition-based evolutionary
  algorithm for many objective optimization,'' \emph{IEEE Transactions on
  Evolutionary Computation}, vol.~19, no.~3, pp. 445--460, June 2015.

\bibitem{Cheng2016b}
R.~Cheng, Y.~Jin, M.~Olhofer, and B.~Sendhoff, ``A reference vector guided
  evolutionary algorithm for many-objective optimization,'' \emph{IEEE
  Transactions on Evolutionary Computation}, vol.~20, no.~5, pp. 773--791, Oct
  2016.

\bibitem{zitzler2004indicator}
E.~Zitzler and S.~K{\"u}nzli, ``Indicator-based selection in multiobjective
  search,'' in \emph{International Conference on Parallel Problem Solving from
  Nature}.\hskip 1em plus 0.5em minus 0.4em\relax Springer, Berlin, Heidelberg,
  2004, pp. 832--842.

\bibitem{Brockhoff2007}
D.~Brockhoff and E.~Zitzler, ``Improving hypervolume-based multiobjective
  evolutionary algorithms by using objective reduction methods,'' in \emph{2007
  IEEE Congress on Evolutionary Computation}, Sept 2007, pp. 2086--2093.

\bibitem{bader2011hype}
J.~Bader and E.~Zitzler, ``{HypE: An} algorithm for fast hypervolume-based
  many-objective optimization,'' \emph{Evolutionary computation}, vol.~19,
  no.~1, pp. 45--76, 2011.

\bibitem{auger2009theory}
A.~Auger, J.~Bader, D.~Brockhoff, and E.~Zitzler, ``Theory of the hypervolume
  indicator: optimal $\mu$-distributions and the choice of the reference
  point,'' in \emph{Proceedings of the tenth ACM SIGEVO workshop on Foundations
  of genetic algorithms}.\hskip 1em plus 0.5em minus 0.4em\relax ACM, 2009, pp.
  87--102.

\bibitem{li2019quality}
M.~Li and X.~Yao, ``Quality evaluation of solution sets in multiobjective
  optimisation: A survey,'' \emph{ACM Computing Surveys}, vol.~52, no.~2, pp.
  26:1--26:38, 2019.

\bibitem{Bosman2003}
P.~A.~N. Bosman and D.~Thierens, ``The balance between proximity and diversity
  in multiobjective evolutionary algorithms,'' \emph{IEEE Transactions on
  Evolutionary Computation}, vol.~7, no.~2, pp. 174--188, April 2003.

\bibitem{coello2004study}
C.~A.~C. Coello and M.~R. Sierra, ``A study of the parallelization of a
  coevolutionary multi-objective evolutionary algorithm,'' in \emph{Proceedings
  of the 3rd Mexican International Conference on Artificial
  Intelligence}.\hskip 1em plus 0.5em minus 0.4em\relax Springer, 2004, pp.
  688--697.

\bibitem{van1998multiobjective}
D.~A. Van~Veldhuizen and G.~B. Lamont, ``Multiobjective evolutionary algorithm
  research: A history and analysis,'' Department of Electrical and Computer
  Engineering. Graduate School of Engineering, Air Force Institute of
  Technology, Wright Patterson, Technical Report TR-98-03, Tech. Rep., 1998.

\bibitem{tian2018}
Y.~Tian, X.~Xiang, X.~Zhang, R.~Cheng, and Y.~Jin, ``Sampling reference points
  on the pareto fronts of benchmark multi-objective optimization problems,'' in
  \emph{2018 IEEE Congress on Evolutionary Computation (CEC)}, July 2018, pp.
  1--6.

\bibitem{zhen2018obj}
L.~Zhen, M.~Li, D.~Peng, and X.~Yao, ``Objective reduction for visualising
  many-objective solution sets,'' in press, 2019.

\bibitem{asafuddowla2018}
M.~Asafuddoula, H.~K. Singh, and T.~Ray, ``An enhanced decomposition-based
  evolutionary algorithm with adaptive reference vectors,'' \emph{IEEE
  Transactions on Cybernetics}, vol.~48, no.~8, pp. 2321--2334, Aug 2018.

\bibitem{Bechikh2015}
S.~Bechikh, M.~Kessentini, L.~B. Said, and K.~Gh¨¦dira, ``Preference
  incorporation in evolutionary multiobjective optimization: a survey of the
  state-of-the-art,'' in \emph{Advances in Computers}, 2015, vol.~98, pp.
  141--207.

\end{thebibliography}
\end{small}
\end{document}